\documentclass[amsa]{ipart}

\RequirePackage[OT1]{fontenc}
\RequirePackage{amsthm,amsmath,amssymb,graphicx,epsfig,subcaption,multirow}
\RequirePackage[numbers,square]{natbib} %
\RequirePackage[colorlinks]{hyperref}
\newcommand{\argmin}{\operatornamewithlimits{argmin}}
\newcommand{\colwidth}{0.8cm}

\usepackage{array}
\newcolumntype{L}[1]{>{\raggedright\let\newline\\\arraybackslash\hspace{0pt}}m{#1}}
\newcolumntype{C}[1]{>{\centering\let\newline\\\arraybackslash\hspace{0pt}}m{#1}}
\newcolumntype{R}[1]{>{\raggedleft\let\newline\\\arraybackslash\hspace{0pt}}m{#1}}

\pubyear{0000}
\volume{0}
\issue{0}
\firstpage{1}
\lastpage{9}

\received{\smonth{12} \sday{7}, \syear{2012}}

\begin{document}

\begin{frontmatter}
\title{Visual Concepts and Compositional Voting}

\begin{aug}
\author{\fnms{Jianyu} \snm{Wang}, \fnms{Zhishuai} \snm{Zhang}, \fnms{Cihang} \snm{Xie}, \fnms{Yuyin} \snm{Zhou}, \fnms{Vittal} \snm{Premachandran}, \fnms{Jun} \snm{Zhu}, \fnms{Lingxi} \snm{Xie}, \fnms{Alan} \snm{Yuille}}

\affiliation{University of California, Los Angeles and Johns Hopkins University}

\end{aug}

\begin{abstract}

It is very attractive to formulate vision in terms of pattern theory \cite{Mumford2010pattern}, where patterns are defined hierarchically by compositions of elementary building blocks. But applying pattern theory to real world images is currently less successful than discriminative methods such as deep networks. Deep networks, however, are black-boxes which are hard to interpret and can easily be fooled by adding occluding objects. It is natural to wonder whether by better understanding deep networks we can extract building blocks which can be used to develop pattern theoretic models. This motivates us to study the internal representations of a deep network using vehicle images from the PASCAL3D+ dataset. We use clustering algorithms to study the population activities of the features and extract a set of visual concepts which we show are visually tight and correspond to semantic parts of vehicles. To analyze this we annotate these vehicles by their semantic parts to create a new dataset, VehicleSemanticParts, and evaluate visual concepts as unsupervised part detectors. We show that visual concepts perform fairly well but are outperformed by supervised discriminative methods such as Support Vector Machines (SVM). We next give a more detailed analysis of visual concepts and how they relate to semantic parts. Following this, we use the visual concepts as building blocks for a simple pattern theoretical model, which we call compositional voting. In this model several visual concepts combine to detect semantic parts. We show that this approach is significantly better than discriminative methods like SVM and deep networks trained specifically for semantic part detection. Finally, we return to studying occlusion by creating an annotated dataset with occlusion, called VehicleOcclusion, and show that compositional voting outperforms even deep networks when the amount of occlusion becomes large.


\end{abstract}

\end{frontmatter}

\section{Introduction}
\label{Introduction}

It is a pleasure to write an article for a honoring David Mumford's enormous contributions to both artificial and biological vision.  This article addresses one of David's main interests, namely the development of grammatical and pattern theoretic models of vision \cite{zhu2007stochastic}\cite{Mumford2010pattern}. These are generative models which represent objects and images in terms of hierarchical compositions of elementary building blocks. This is arguably the most promising approach to developing models of vision which have the same capabilities as the human visual system and can deal with the exponential complexity of images and visual tasks. Moreover, as boldly conjectured by David \cite{mumford1992computational,lee2003hierarchical}, they suggest plausible models of biological visual systems and, in particular, the role of top-down processing.

\begin{figure}[h]
\centering
\includegraphics[width=0.33\linewidth]{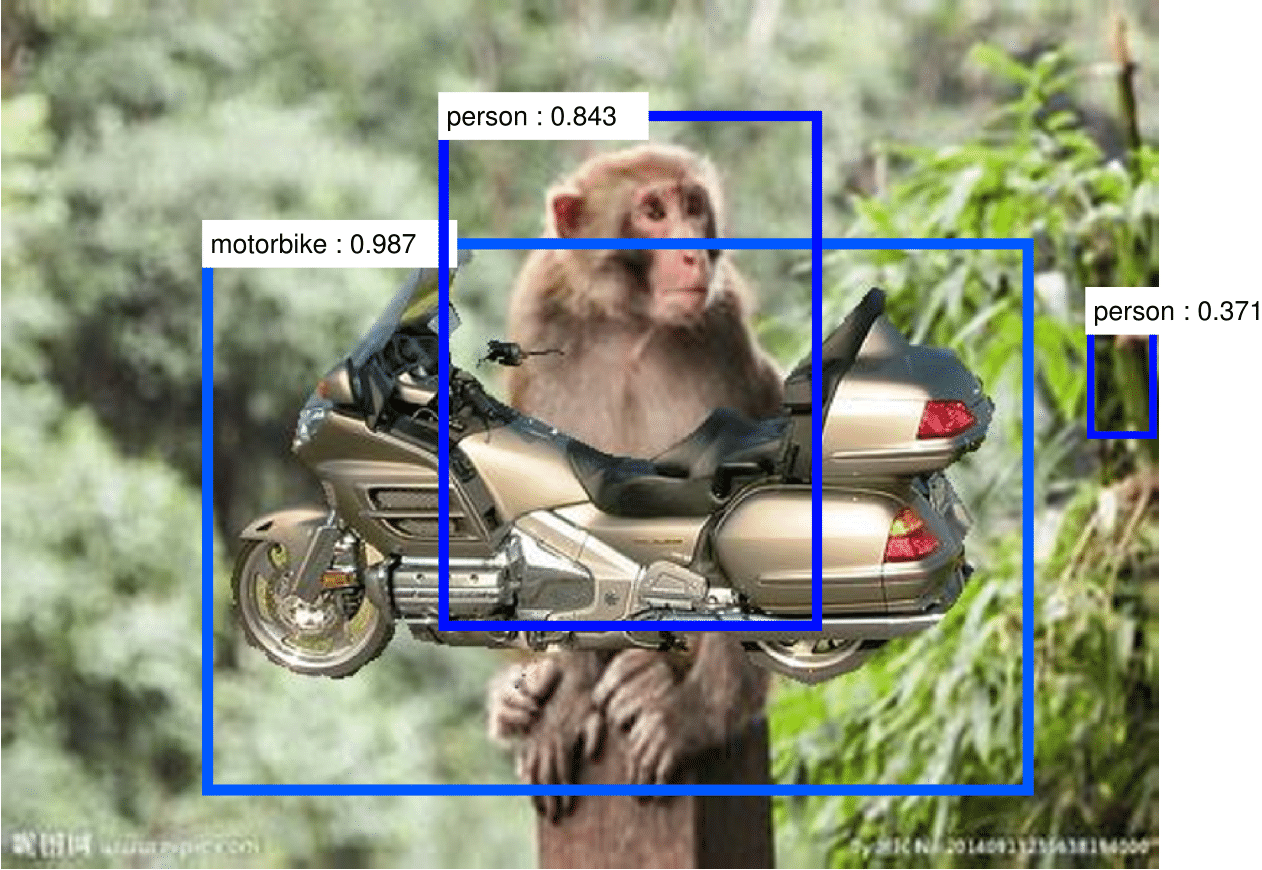}
\includegraphics[width=0.33\linewidth]{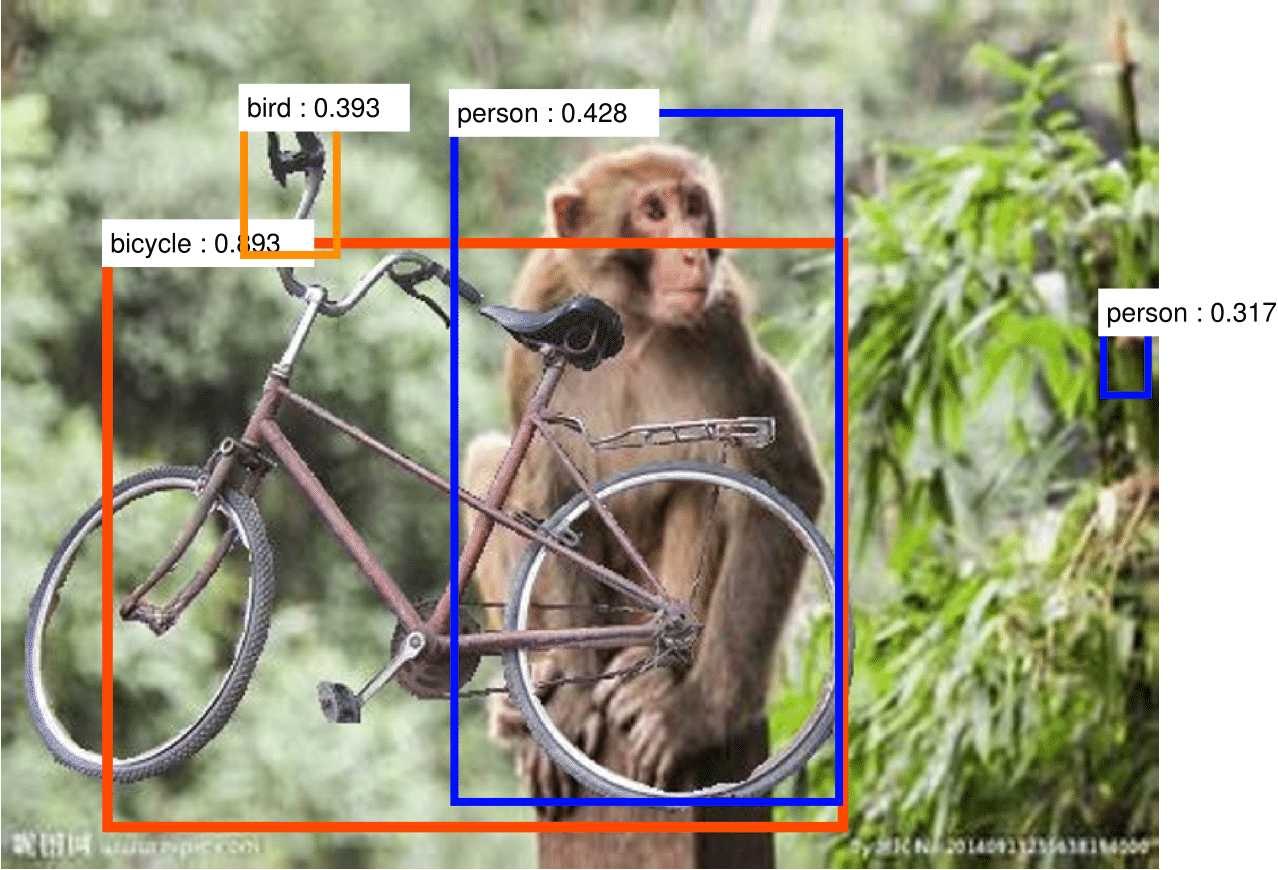}
\includegraphics[width=0.3\linewidth]{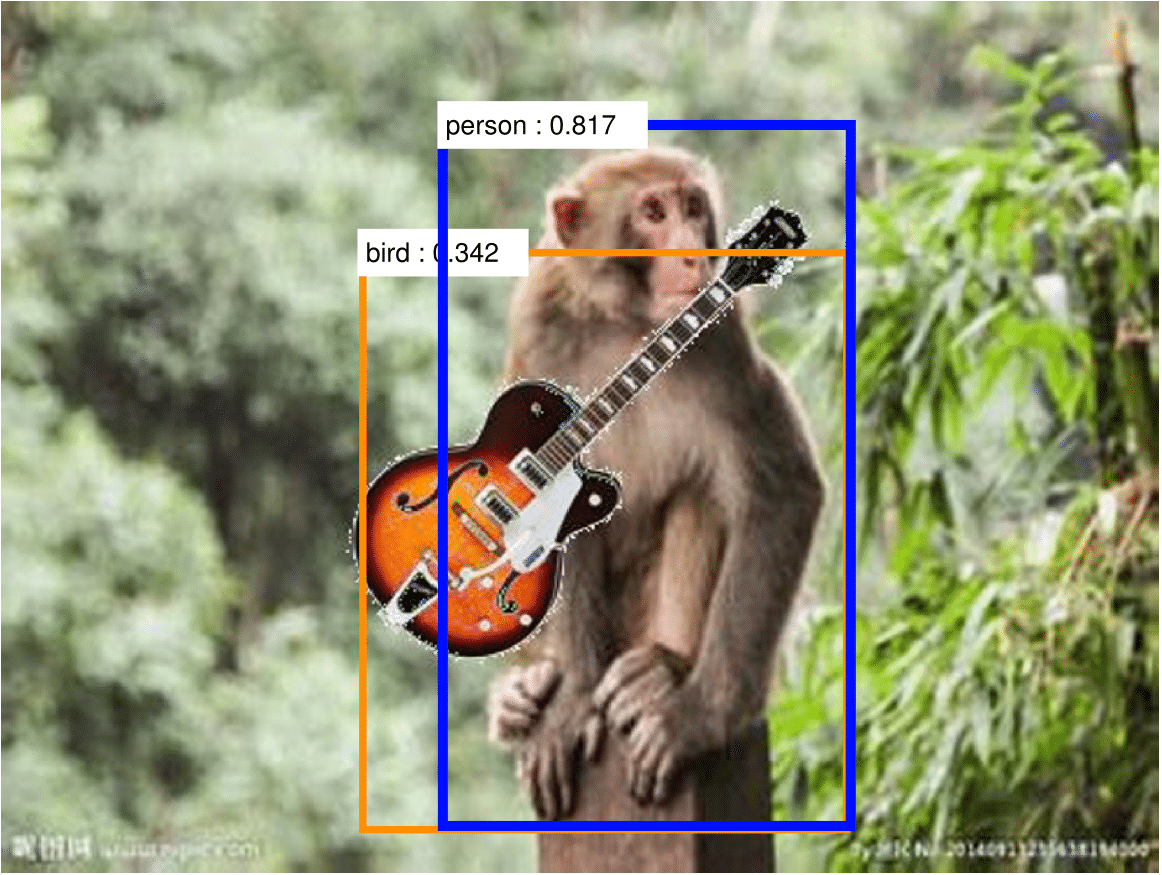}
\caption{Caption: Adding occluders causes deep network to fail. We refer such examples as adversarial context examples since the failures are caused by misusing context/occluder info. Left Panel: The occluding motorbike turns a monkey into a human. Center Panel: The occluding  bicycle turns a monkey into a human and the jungle turns the bicycle handle into a bird.
Right Panel: The occluding  guitar turns the monkey into  a human and the jungle turns the guitar into a bird. }
\label{fig:cihang_monkey}
\end{figure}

But despite the theoretical attractiveness of pattern theoretic methods they have not yet produced visual algorithms capable of competing with alternative discriminative approaches such as deep networks. One reason for this is that, due to the complexity of image patterns, it very hard to specify their underlying building blocks patterns (by contrast, it is much easier to develop grammars for natural languages where the building blocks, or terminal nodes, are words). But deep networks have their own limitations, despite their ability to learn hierarchies of image features in order to  perform an impressive range of visual tasks on challenging datasets. Deep networks are black boxes which are hard to diagnose and they have limited ability to adapt to novel data which were not included in the dataset on which they were trained. For example, figure~(\ref{fig:cihang_monkey}) shows how the performance of a deep network degrades when we superimpose a guitar on the image of a monkey. The presence of the guitar causes the network to misinterpret the monkey as being a human while also mistaking the guitar for a bird. The deep network presumably makes these mistakes because it has never seen a monkey with a guitar, or a guitar with a monkey in the jungle (but has seen a bird in the jungle or a person with a musical instrument). In short, the deep network is over-fitting to the {\it typical} context of the object. 

But over-fitting to visual context is problematic for standard machine learning approaches, such as deep networks, which assume that their training and testing data is representative of some underlying, but unknown, distribution \cite{valiant1984theory,Vapnik1995}. The problem is that images, and hence visual context, can be infinitely variable. A realistic image can be created by selecting from a huge range of objects (e.g., monkeys, guitars, dogs, cats, cars, etc.) and placing them in an enormous number of different scenes (e.g., jungle, beach, office) and in an exponential number of different three-dimensional positions. Moreover, in typical images, objects are usually occluded as shown in figure~(\ref{fig:cihang_monkey}). Occlusion is a fundamental aspect of natural images and David has recognized its importance both by proposing a  2.1 D sketch representation \cite{DBLP:conf/iccv/NitzbergM90} and also by using it as the basis for a ``dead leaves'' model for image statistics  \cite{lee2001occlusion}.

Indeed the number of ways that an object can be occluded is, in itself, exponential. For example, if an object has $P$ different parts then we can occlude any subset of them using any of $N$ different occluders yielding $O(P^N)$ possible occluding patterns. In the face of this exponential complexity the standard machine learning assumptions risk breaking down. We will never have enough data both to train, and to test, black box vision algorithms. As in well known many image datasets have confounds which mean that algorithms trained on them rarely generalize to other datasets \cite{bollier2010promise})\. Instead we should be aiming for visual algorithms that can learn objects (and other image structures) from limited amounts of data (large, but not exponentially large). In addition, we should devise strategies where the algorithms can be tested on potentially infinite amounts of training data. One promising strategy is to expand the role of {\it adversarial noise} \cite{szegedy2013intriguing}\cite{goodfellow2014explaining}\cite{xie2017adversarial}, so that we can create images which {\it stress-test} algorithms and deliberately target their weaknesses. This strategy is more reminiscent of game theory than  the standard assumptions of machine learning (e.g., decision theory). 

This motivates a research program which we initiate in this paper by addressing three issues. Firstly, we analyze the internal structure of deep networks in order both to better understand them but, more critically, to extract internal representations which can be used as building blocks for pattern theoretic models. We call these representations {\it visual concepts} and extract them by analyzing the population codes of deep networks. We quantify their ability for detecting semantic parts of objects. Secondly, we develop a simple compositional-voting model to detect semantic parts of objects by using these building blocks. These models are similar to those in ~\cite{Amit_2002_2D}.
Thirdly, we stress test these compositional models, and deep networks for comparison, on datasets where we have artificially introduced occluders. We emphasize that neither algorithm, compositional-voting or the deep network, has been trained on the occluded data and our goal is to stress test them to see how well they perform under these adversarial conditions. 

In order to perform these studies we have constructed two datasets
based on the PASCAL3D+ dataset \cite{xiang2014beyond}. Firstly, we annotate the semantic parts of the vehicles in this dataset to create the \emph{VehicleSemanticPart dataset.} Secondly, we create the {\it Vehicle Occlusion dataset} by superimposing cutouts of objects from the PASCAL segmentation dataset \cite{Everingham_2010_PASCAL} into images from the VehicleSemanticPart dataset. We use the VehicleSemanticPart dataset to extract visual concepts by clustering the features of a deep network  VGG-16~\cite{Simonyan_2015_Very}, trained on ImageNet, when the deep network is shown objects from VehicleSemanticPart. We evaluate the visual concepts, the compositional-voting mode;s, and the deep networks for detecting semantic parts image with and without occlusion, i.e. VehicleSemanticPart and Vehicle Occlusion respectively.

\section{Related Work}
\label{sec:related}

David's work on pattern theory \cite{Mumford2010pattern} was inspired by Grenander's seminal work \cite{grenander1996elements} (partly due to an ARL center grant which encouraged collaboration between Brown University and Harvard). David was bold enough to conjecture that the top-down neural connections in the visual cortex were for performing analysis by synthesis \cite{mumford1992computational}. Advantages of this approach include the theoretical ability to deal with large numbers of objects by part sharing \cite{yuille2016complexity}. But a problem for these types of models was the difficulty in specifying suitable building blocks which makes learning grammars for vision considerably harder than learning them for natural language \cite{tu2012unambiguity,tu2013unsupervised}. For example, researchers could learn some object models in an unsupervised manner, e.g., see \cite{zhu2008unsupervised}, but this only used edges as the building blocks and hence ignored most of the appearance properties of objects. Attempts have been made to obtain building blocks by studying properties of image patches \cite{papandreou2014modeling} and \cite{changbolin2017} but these have yet to be developed into models of objects. Though promising, these are pixel-based and are sensitive to spatial warps of images and do not, as yet,  capture the hierarchy of abstraction that seems necessary to deal with the complexity of real images.

Deep networks, on the other hand,  have many desirable properties. such as hierarchies of feature vectors (shared across objects) which capture increasingly abstract, or invariant, image patterns as they ascend the hierarchy from the image to the object level. This partial invariance arises  because they are trained in a discriminative way so that decisions made about whether a viewed object is a car or not must be based on hierarchical abstractions. It has been shown, see  \cite{ZhouKLOT14}\cite{LiLSH15}\cite{XiaoXYZPZ15}\cite{SimonR15}, that  the activity patterns of deep network features  do encode objects and object parts, mainly be considering the activities of individual neurons in the deep networks.

Our approach to study the activation patterns of populations of neuron \cite{wang2015unsupervised} was inspired by the well-known neuroscience debate about the neural code. On one extreme is the concept of grandmother calls  described by Barlow \cite{barlow1972single} which is contrasted to population encoding as reported by Georgopoulos {\it et al.} \cite{georgopoulos1986neuronal}. Barlow was motivated by the finding that comparatively few neurons are simultaneously active and proposed sparse coding as a general principle. In particular, this suggests a form of {\it extreme sparsity} where only one, or a few cells, respond to a specific local stimulus (similar to a matched template). As we will see in section~(\ref{sec:discussion}), visual concepts tend to have this property which is developed further in later work. There are, of course, big differences between studying population codes in artificial neural networks and studying them for real neurons. In particular, as we will show, we can modify the neural network so that the populations representing a visual concept can be encoded by  \emph{visual concept neurons}, which we call \emph{vc-neurons}. Hence our approach is consistent with both extreme neuroscience perspectives. We might speculate that the brain uses both forms of neural encoding useful for different purposes: a population, or {\it signal} representation, and a sparser {\it symbolic} representation. This might be analogous to how words can be represented as binary encodings (e.g., cat, dog, etc.) and by continuous vector space encodings which similarity between vectors captures semantic information \cite{mikolov2013distributed}. Within this picture, neurons could represent a manifold of possibilities which the vc-neurons would quantize into discrete elements. 

We illustrate the use of visual concepts to build a compositional-voting model inspired by ~\cite{Amit_2002_2D}, which is arguably the simplest pattern theoretic method. We stress that voting schemes have frequently been used in computer vision without being thought of as examples of pattern theory methods~\cite{Grimson_1990_Object}\cite{Leibe_2004_Combined}\cite{Maji_2009_Object}\cite{Okada_2009_Discriminative}.

To stress test our algorithms we train them on un-occluded images and test them on datasets with occlusion. Occlusion is almost always present in natural images, though less frequently in computer vision datasets, and is fundamental to human perception. David's work has long emphasized its significance 
\cite{DBLP:conf/iccv/NitzbergM90}, \cite{lee2001occlusion} and it is sufficiently important for computer vision researchers to have addressed it using deep networks \cite{wang2016doc}. It introduces difficulties for tasks such as object detection~\cite{Kar_2015_Amodal} or segmentation~\cite{Li_2016_Amodal}. As has already been shown, part-based models are very useful for detecting partially occluded objects~\cite{Chen_2014_Detect}
\cite{Chen_2015_Parsing}\cite{Li_2014_Integrating}.


\section{The Datasets \label{sec:Datasets}}

In this paper we use three related annotated datasets. The first is the vehicles in the {\it PASCAL3D+ dataset}  \cite{xiang2014beyond} and their keypoint interactions. The second is the {\it VehicleSemanticPart dataset} which we created by annotating semantic parts on the vehicles in the {\it PASCAL3D+ dataset} to give a richer representation of the objects than the keypoints. The third is the {\it Vehicle Occlusion dataset} where we occluded the objects in the first two datasets. 

\begin{figure}[!h]
  \centering
    \begin{minipage}[.8in]{.3\textwidth}\centering
      \includegraphics[height=1in]{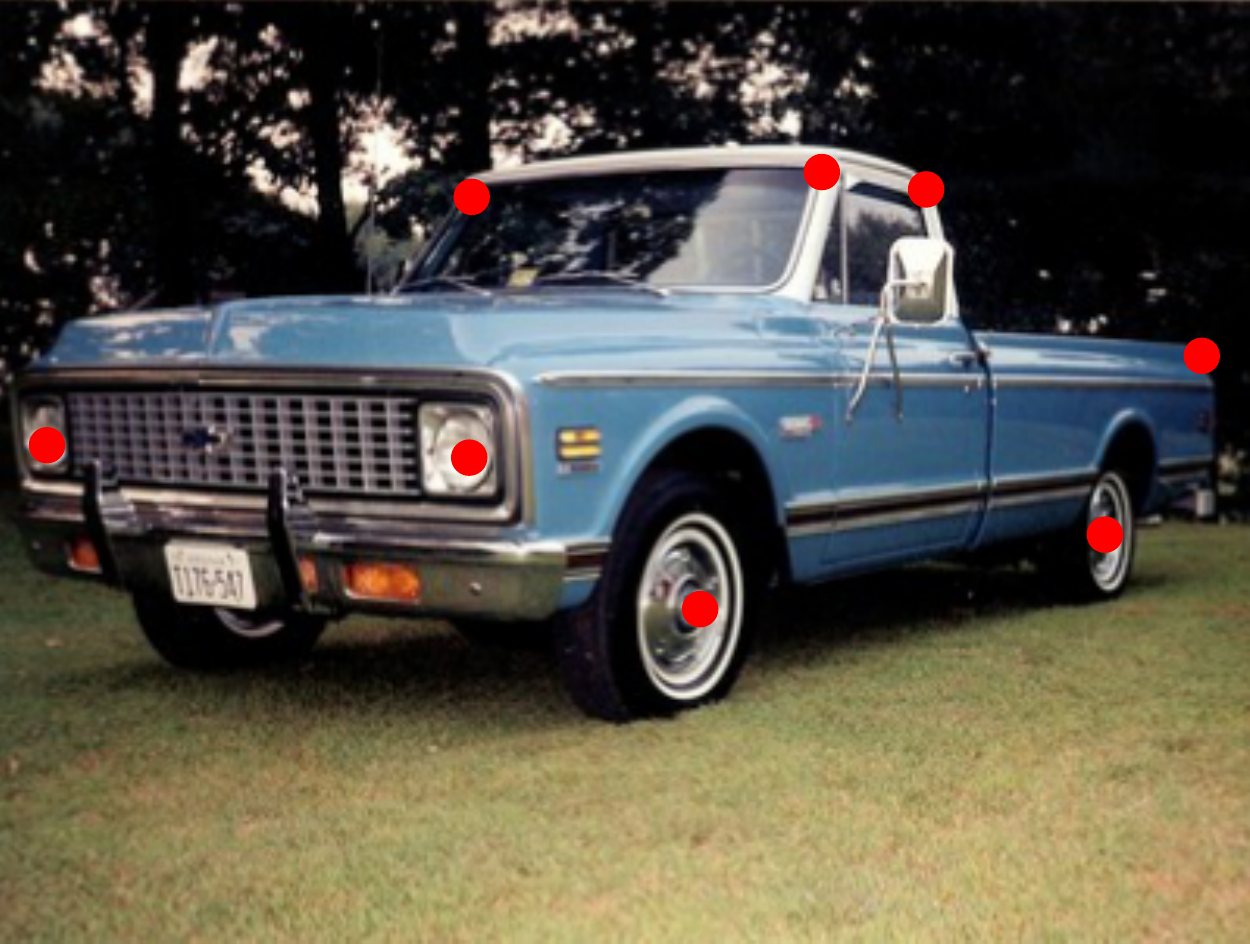}
    \end{minipage}\hfill
    \begin{minipage}[.8in]{.4\textwidth}\centering
      \includegraphics[height=1in]{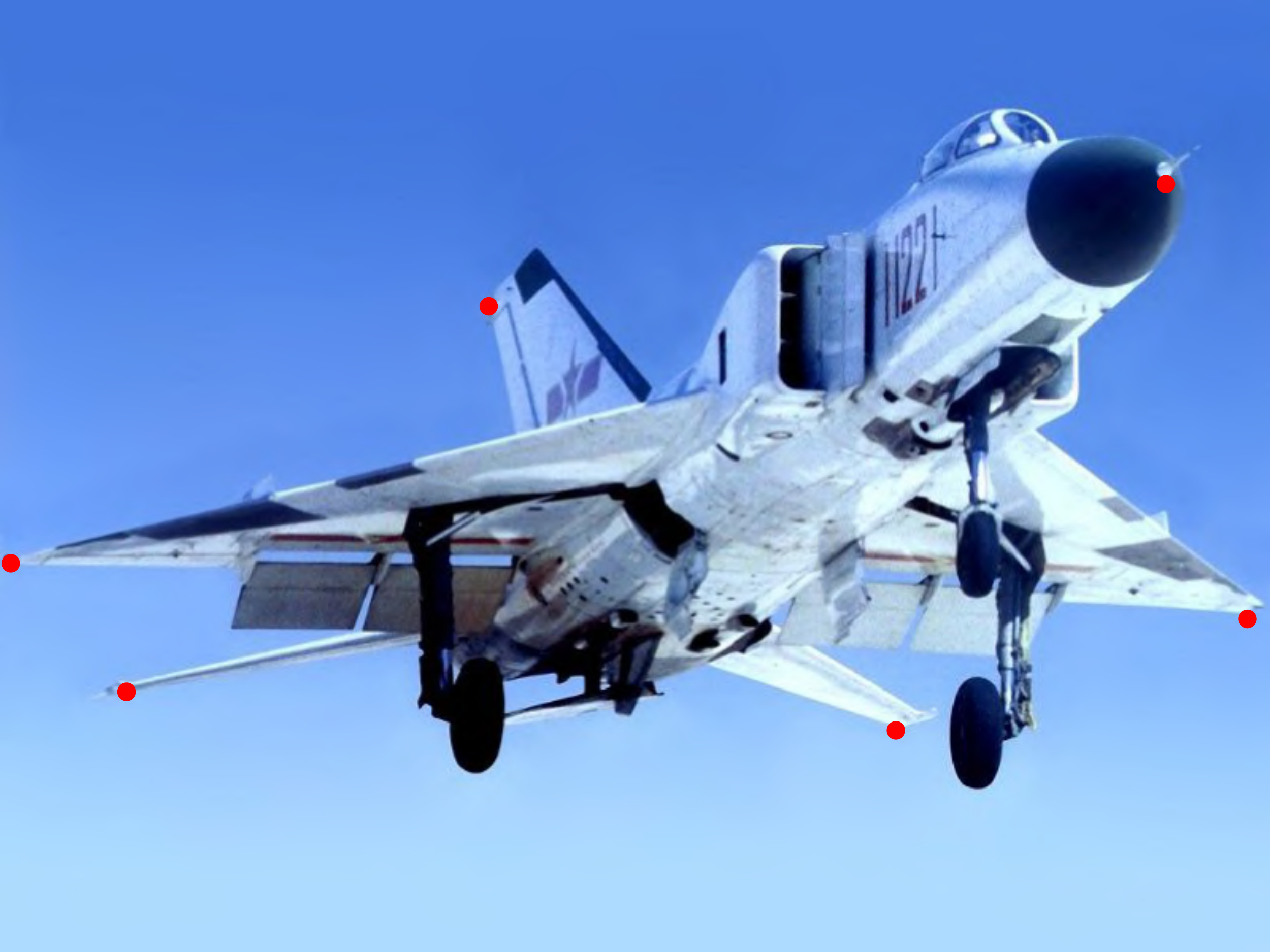}\hfill
    \end{minipage}\hfill
    \begin{minipage}[.8in]{.3\textwidth}\centering
      \includegraphics[height=1in]{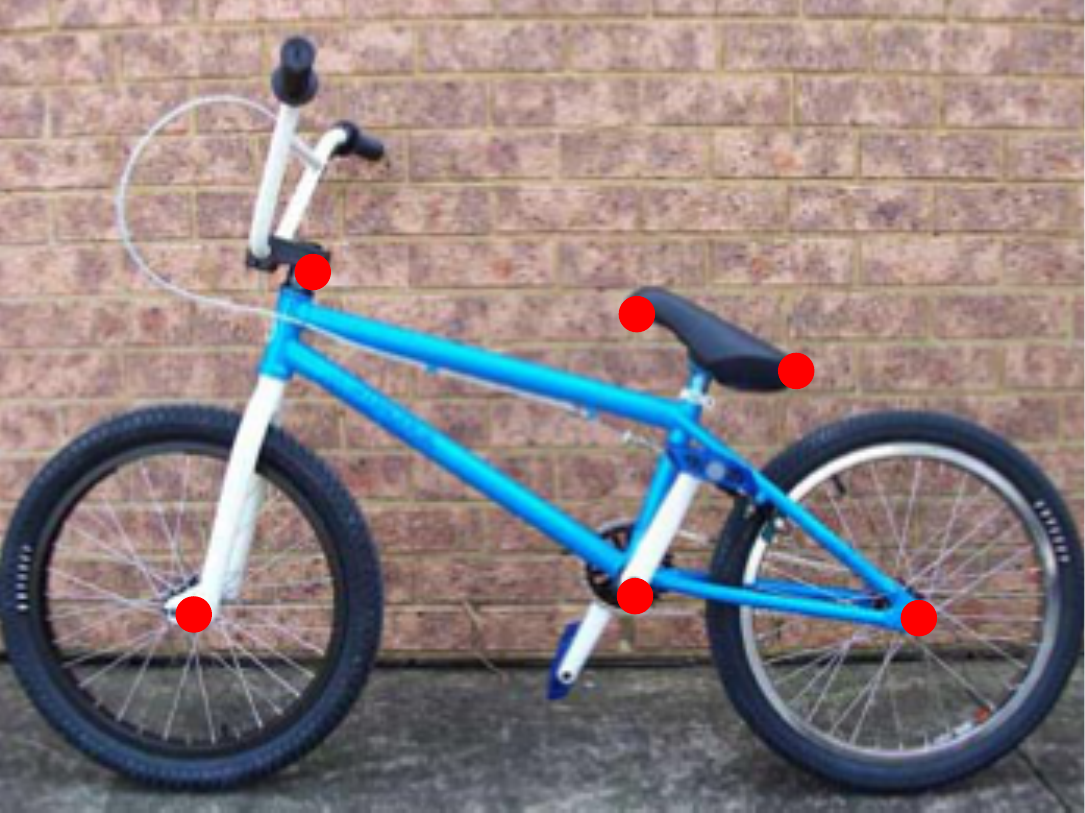}
    \end{minipage}
    \begin{minipage}[.8in]{.3\textwidth}\centering
      \includegraphics[height=1in]{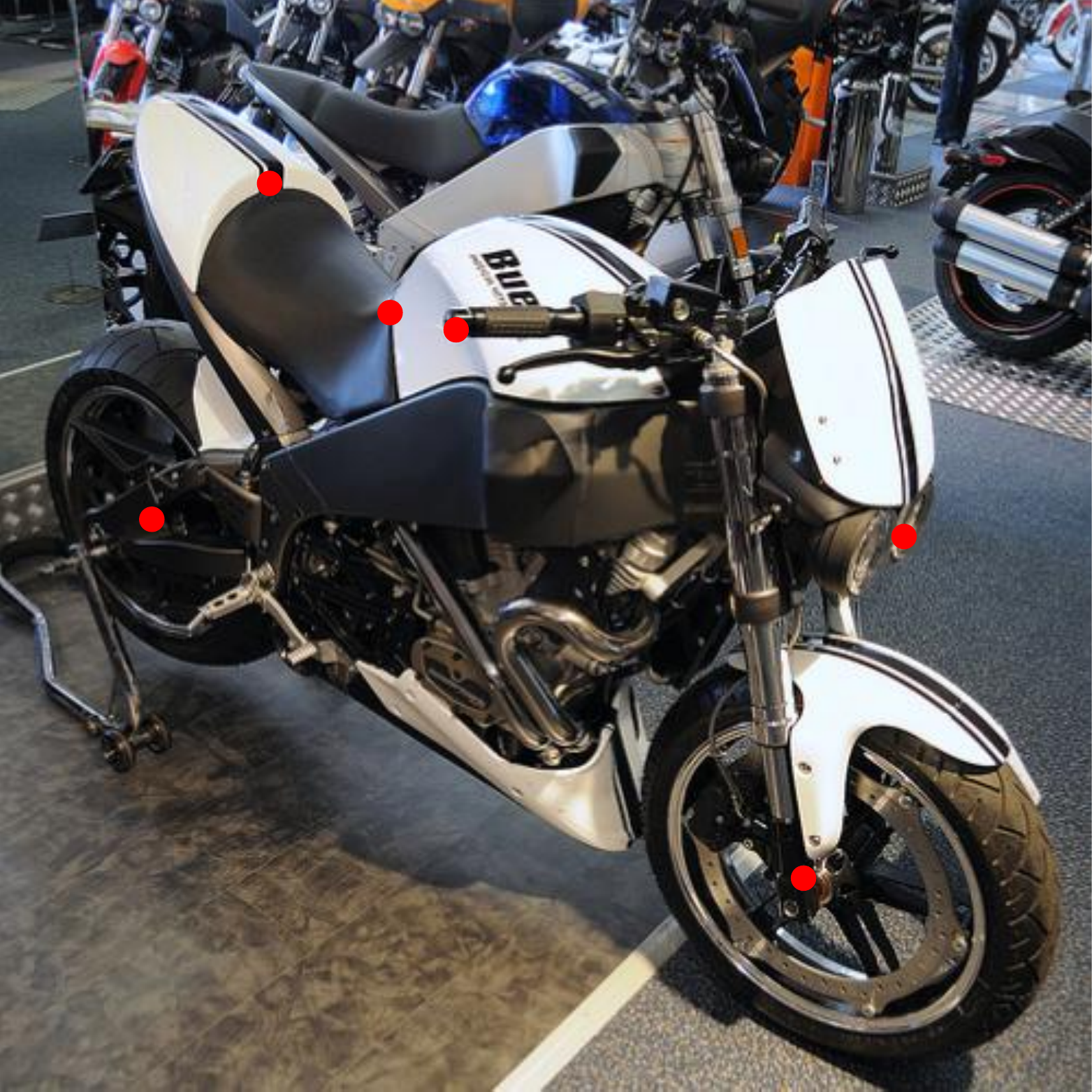}
    \end{minipage}\hfill
    \begin{minipage}[.8in]{.4\textwidth}\centering
      \includegraphics[height=1in]{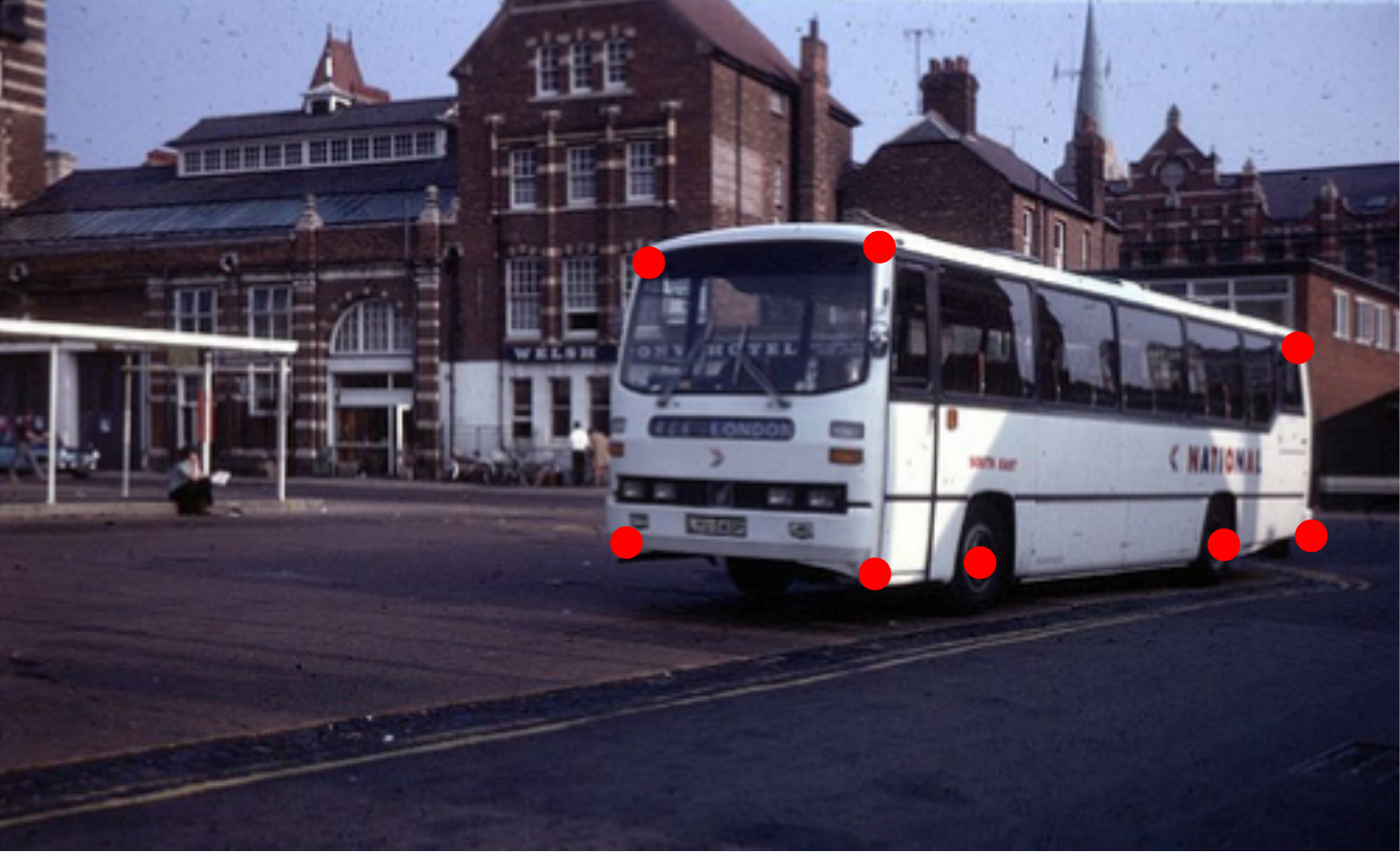}\hfill
    \end{minipage}\hfill
    \begin{minipage}[.8in]{.3\textwidth}\centering
      \includegraphics[height=1in]{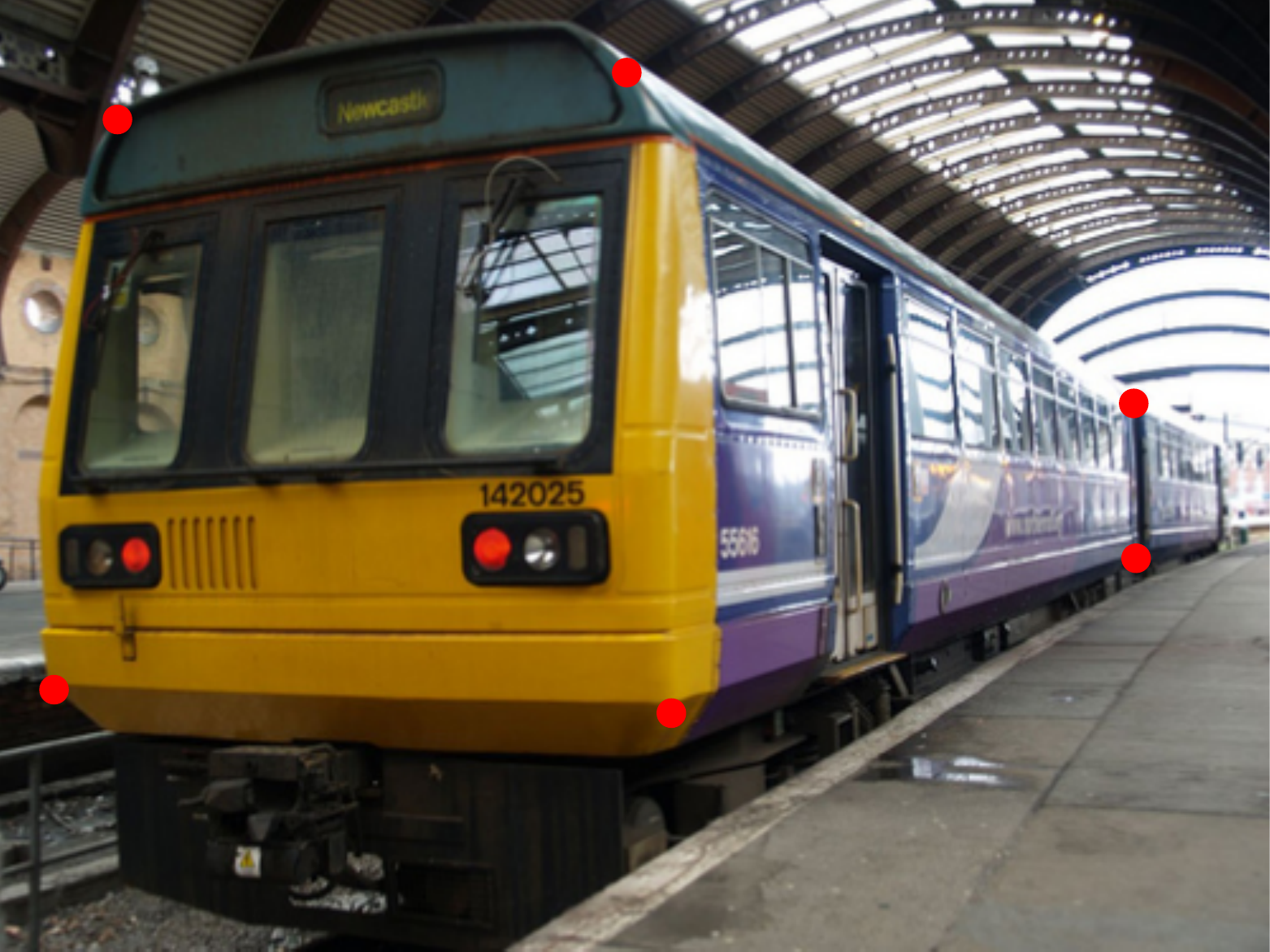}
    \end{minipage}
    \caption{The figure illustrates the keypoints on the PASCAl3D+ dataset. For cars, there are keypoints for wheels, headlights, windshield corners and trunk corners. Keypoints are specified on salient parts of other objects. Best seen in color.}
    \label{fig:annotation_keypoint}
  \end{figure}

The  vehicles in the {\it PASCAL3D+ dataset} are: (i) cars, (ii) airplanes, (iii) motorbikes, (iv) bicycles, (v) buses, and (vi) trains. These images in the PASCAL3D+ dataset are selected from the Pascal and the ImageNet datasets. In this paper we report results only for those images from ImageNet, but we obtain equivalent results on PASCAL, see arxiv paper \cite{wang2015unsupervised}. These images were supplemented with keypoint annotations (roughly ten keypoints per object) and also estimated orientations of the objects in terms of the azimuth and the elevation angles. These objects and their keypoints are illustrated in  figure~(\ref{fig:annotation_keypoint})(upper left panel).

We use this dataset to extract the visual concepts, as described in the following section. In order to do this, we first re-scale the images so that the minimum of the height and width is 224 pixels (keeping the aspect ratio fixed). We also use the keypoint annotations to evaluate the ability of the visual concepts to detect the keypoints. But there are typically only 10 keypoints for each object which means they can only give limited evaluation of the visual concepts.

\begin{figure}[!h]
    \begin{minipage}{.3\textwidth}\centering
      \includegraphics[height=.75in]{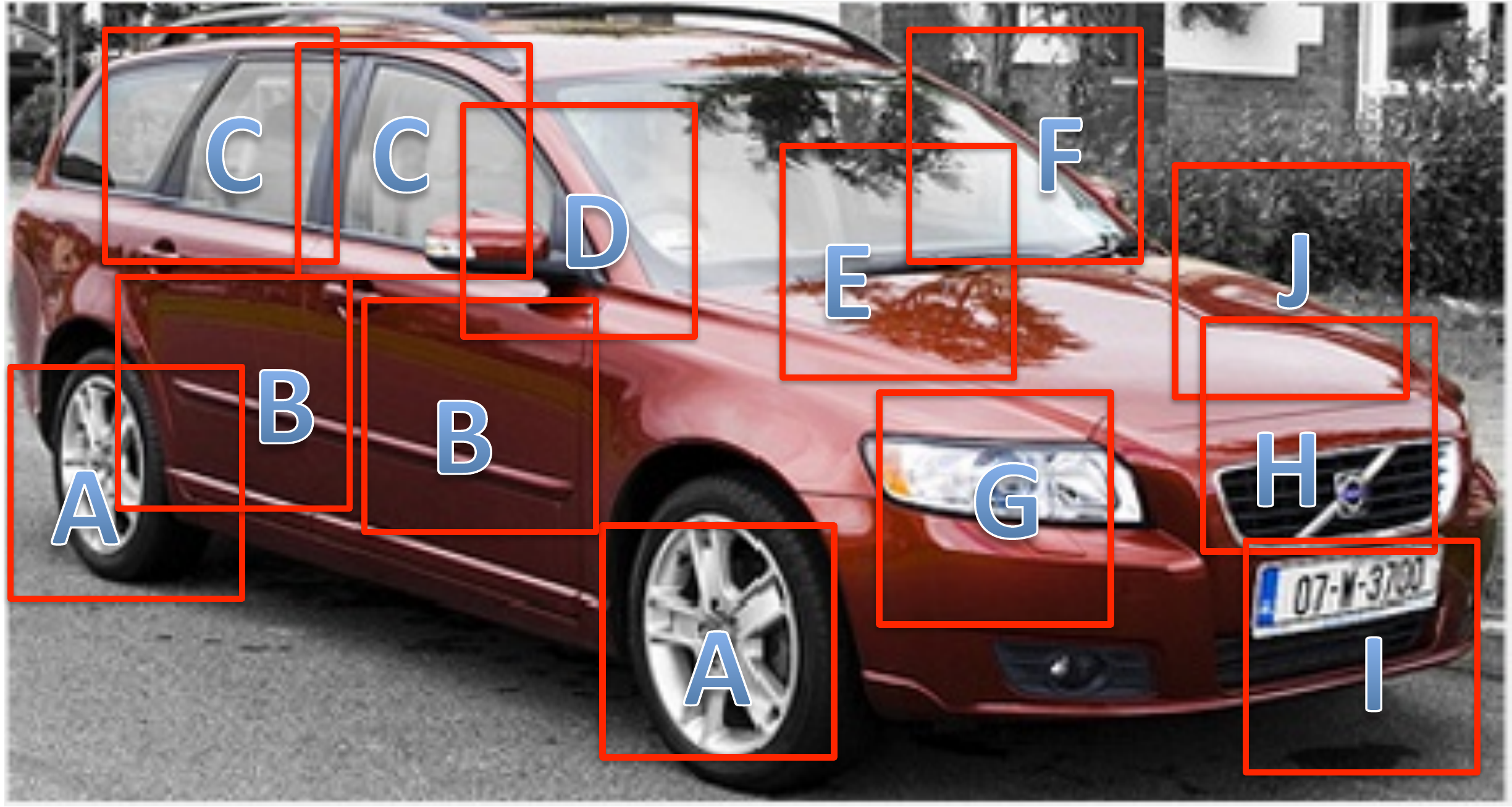}
    \end{minipage}\hfill
    \begin{minipage}{.4\textwidth}\centering
      \includegraphics[trim={0cm 0.8cm 0cm 0cm},clip,height=.75in]{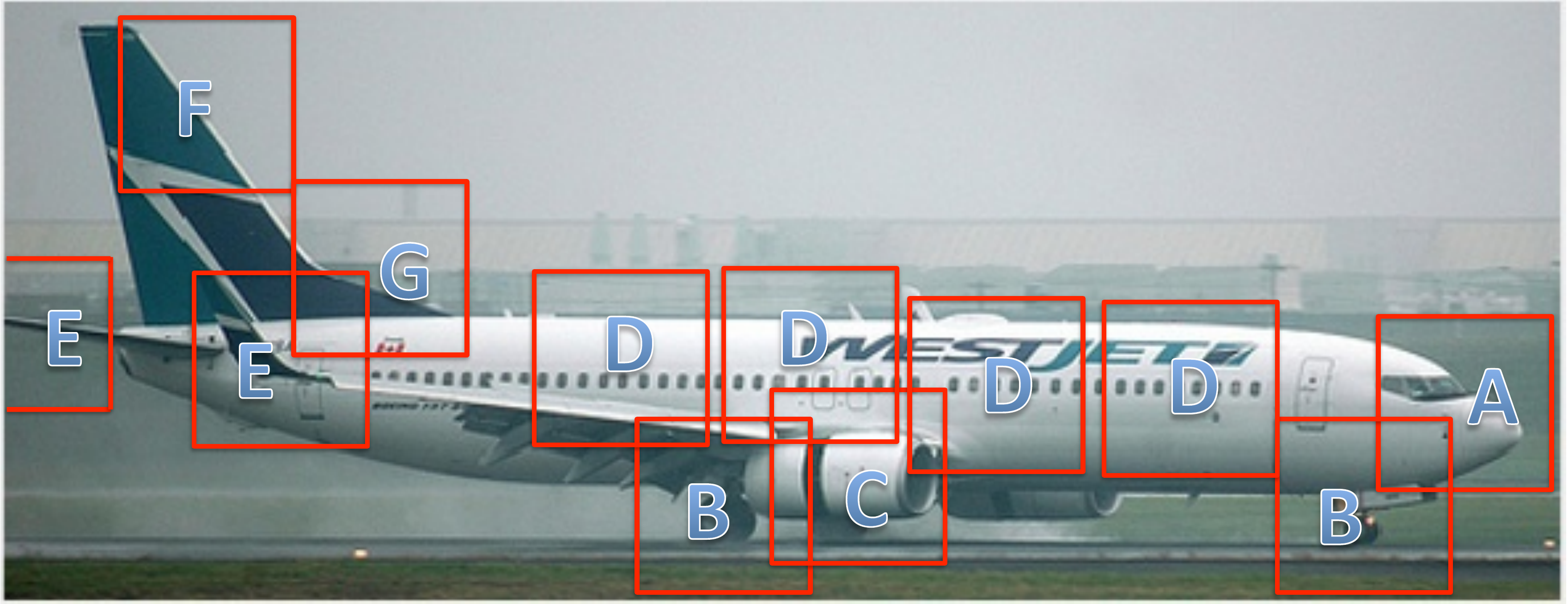}\hfill
    \end{minipage}\hfill
    \begin{minipage}{.3\textwidth}\centering
      \includegraphics[height=.75in]{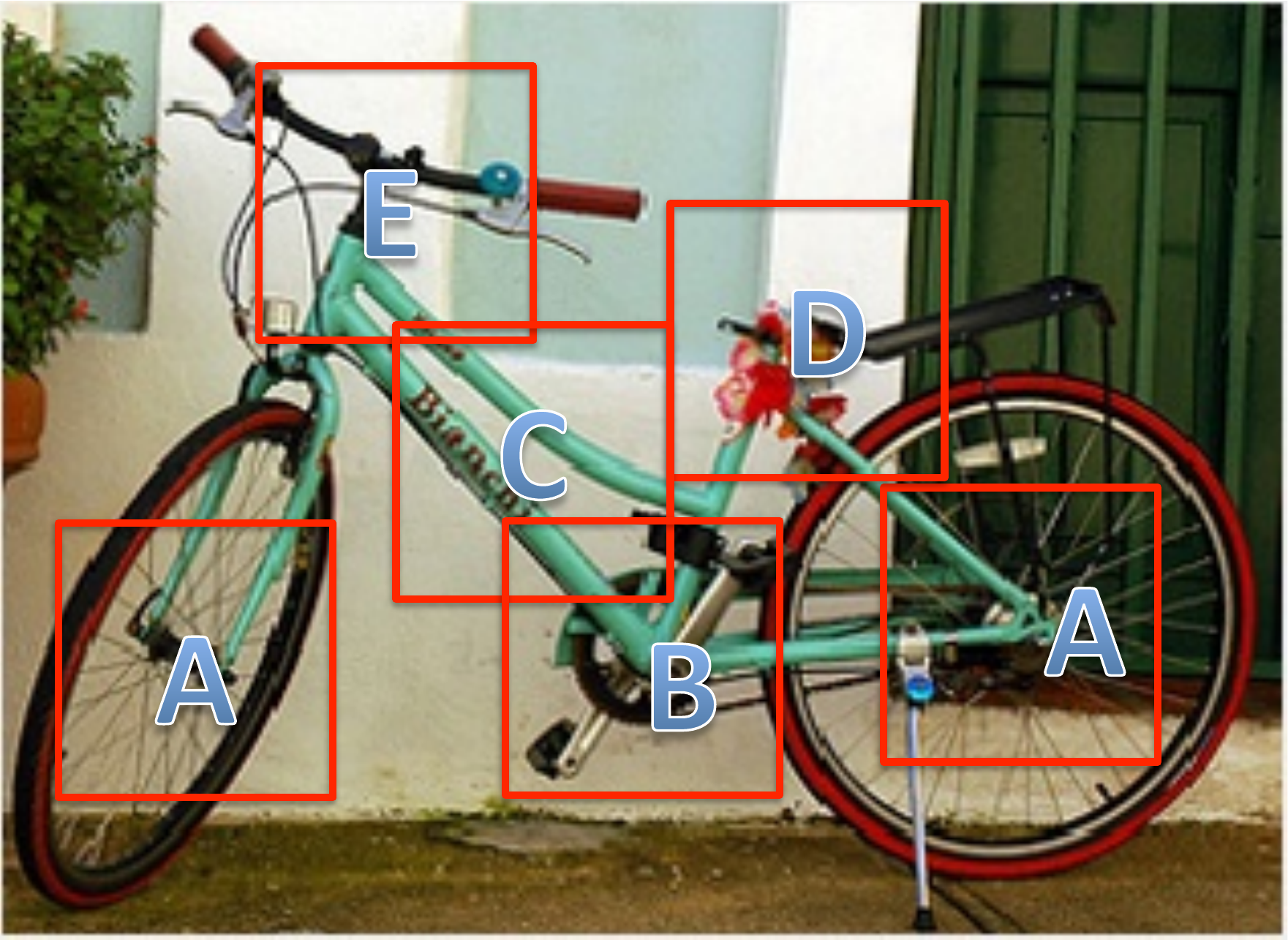}
    \end{minipage}
    \begin{minipage}{.3\textwidth}\centering
      \includegraphics[height=.75in]{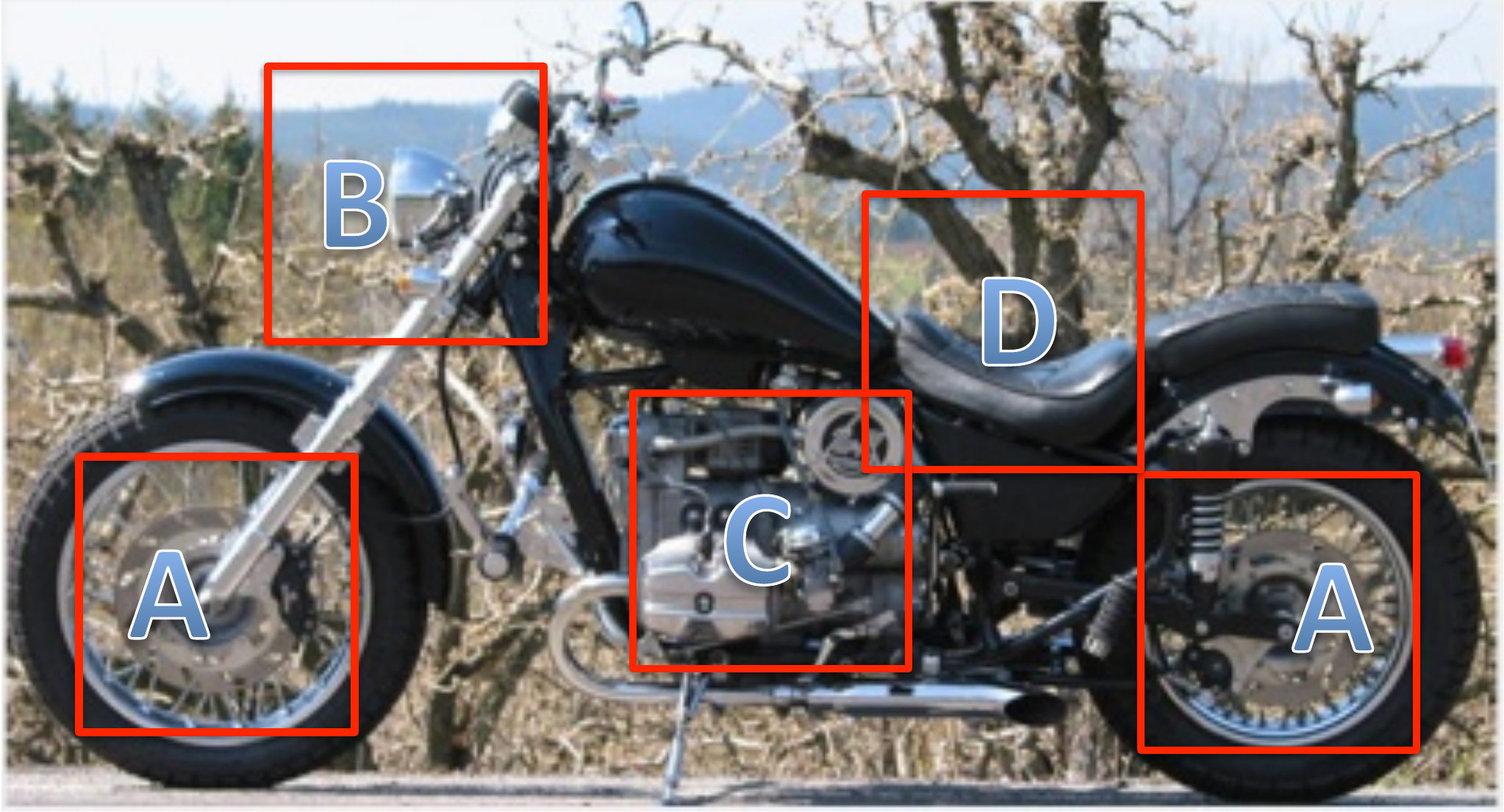}
    \end{minipage}\hfill
    \begin{minipage}{.4\textwidth}\centering
      \includegraphics[height=.75in]{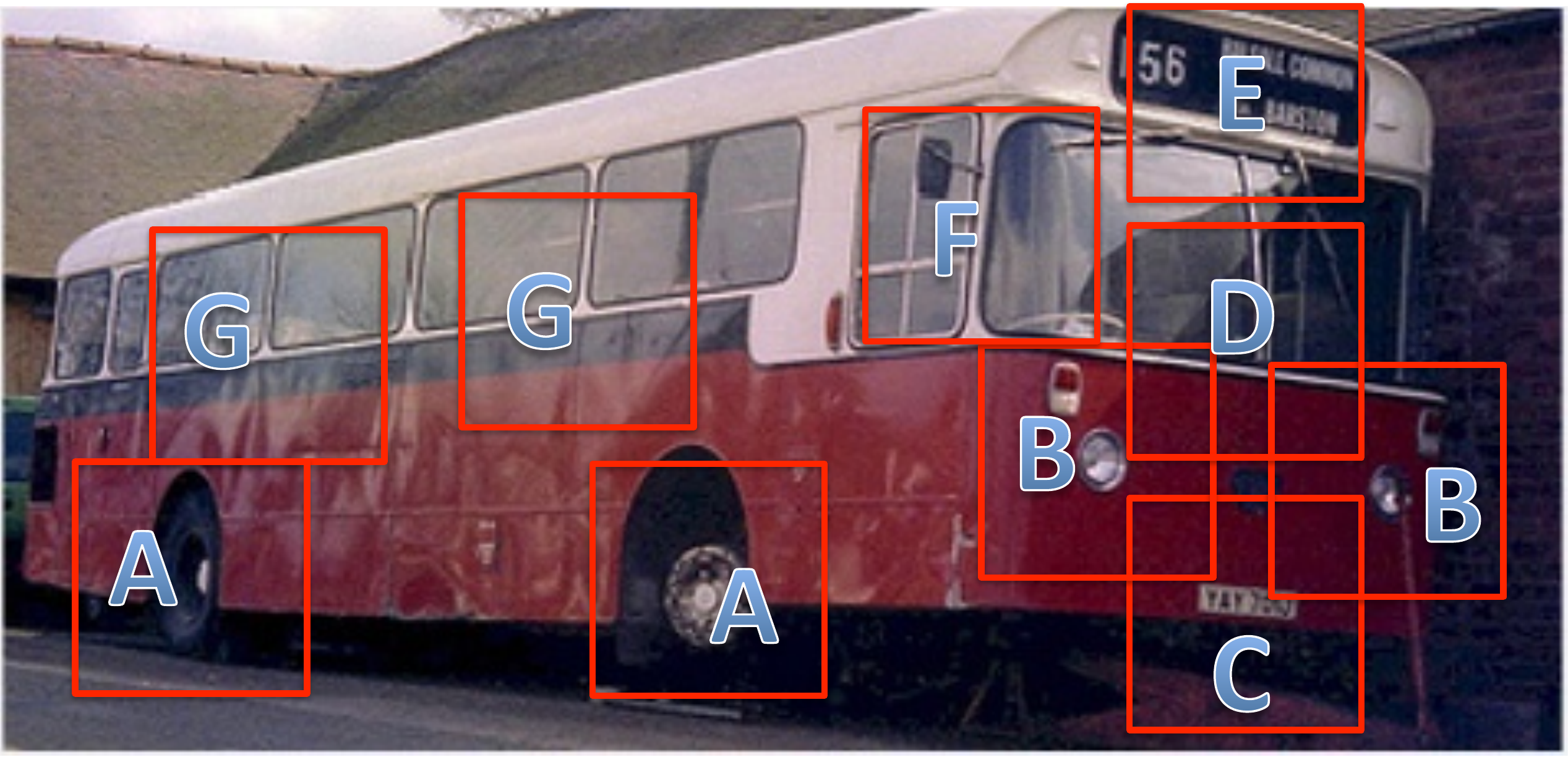}\hfill
    \end{minipage}\hfill
    \begin{minipage}{.3\textwidth}\centering
      \includegraphics[height=.75in]{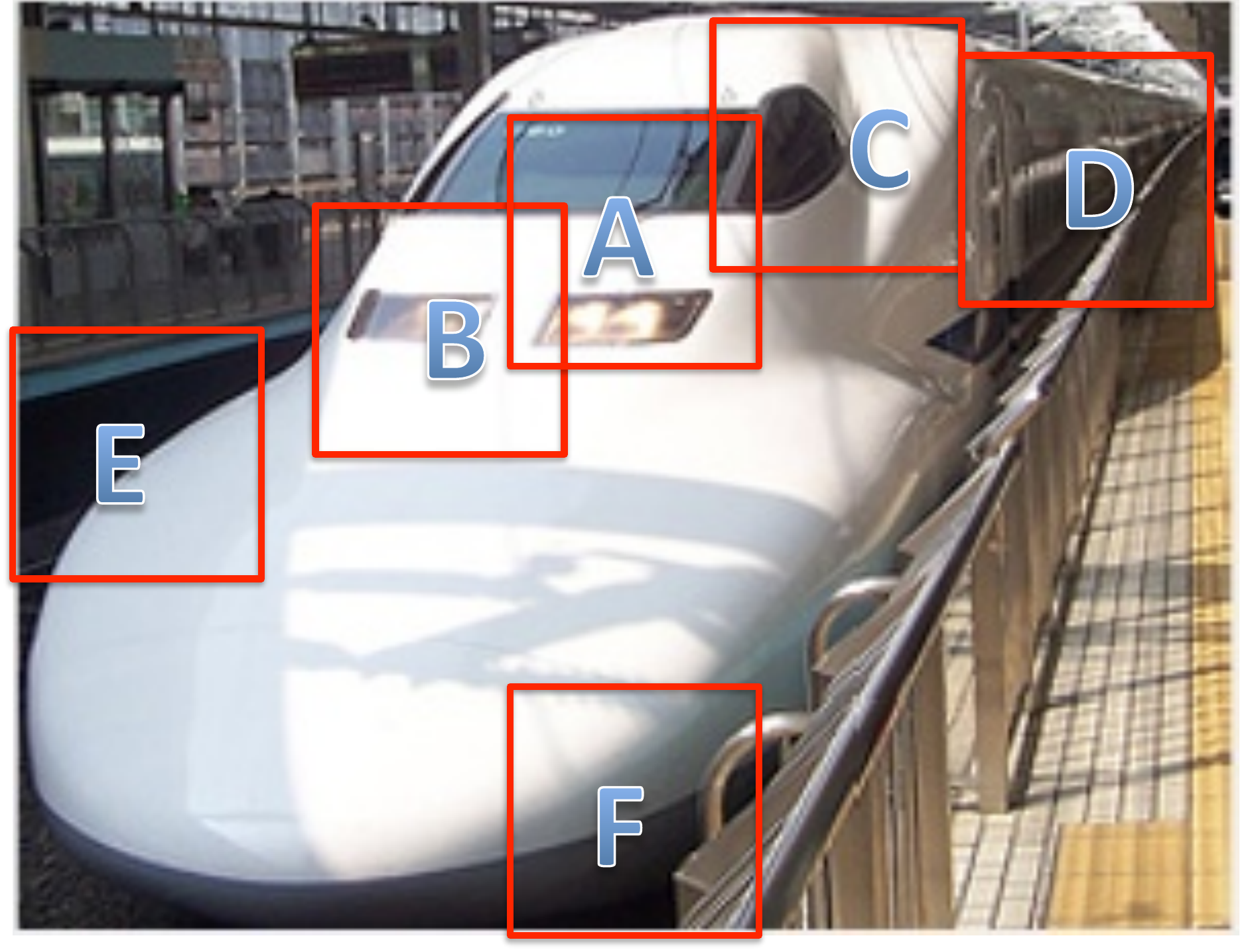}
    \end{minipage}
  \caption{This figure illustrates the semantic parts labelled on the six vehicles in the \emph{VehicleSemanticParts dataset} (remaining panels). For example, for car semantic parts. A: wheels, B: side body, C: side window, D: side mirror and side window and wind shield, E: wind shield and engine hood, F: oblique line from top left to bottom right, wind shield, G: headlight, H: engine hood and air grille, I: front bumper and ground, J: oblique line from top left to bottom right, engine hood. The parts are weakly viewpoint dependent. See text for the semantic parts of the other vehicles.}
  \label{fig:annotation}
\end{figure}

In order to test the visual concepts in more detail we  defined semantic parts on the vehicles and annotated them to create a dataset called \emph{VehicleSemanticParts.} There are thirty nine semantic parts for cars, nineteen for motorbikes, ten for buses, thirty one for aeroplanes, fourteen for bicycles, and twenty for trains. The semantic parts are regions of $100 \times 100$ pixels and provide a dense coverage of the objects, in the sense that almost every pixel on each object image is contained within a semantic part. The semantic parts have verbal descriptions, which are specified in the webpage \url{http://ccvl.jhu.edu/SP.html}. The annotators were trained by being provided by: (1) a one-sentence verbal description, and (2) typical exemplars. The semantic parts are illustrated in figure~(\ref{fig:annotation}) where we show a representative subset of the annotations indexed by ``A", ``B", etc. The car annotations are described in the figure caption and the annotations for the other objects are as follows: (I) \emph{Airplane Semantic Parts}. A: nose pointing to the right, B: undercarriage, C: jet engine, mainly tube-shape, sometimes with highly turned ellipse/circle, D: body with small windows or text, E: wing or elevator tip pointing to the left, F: vertical stabilizer, with contour from top left to bottom right, G: vertical stabilizer base and body, with contour from top left to bottom right. (II) \emph{Bicycle Semantic Parts.} A: wheels, B: pedal, C: roughly triangle structure between seat, pedal, and handle center, D: seat, E: roughly T-shape handle center. (III) \emph{Bus Semantic Parts.} A: wheels, B: headlight, C: license plate, D: window (top) and front body (bottom), E: display screen with text, F: window and side body with vertical frame in the middle, G: side windows and side body. (IV) \emph{Motorbike Semantic Parts.} A: wheels, B: headlight, C: engine, D: seat. (V) \emph{Train Semantic Parts. } A: front window and headlight, B: headlight, C: part of front window on the left and side body, D: side windows or doors and side body, E: head body on the right or bottom right and background, F: head body on the top left and background.).

In order to study occlusion we create the \textbf{VehicleOcclusion} dataset. This consists of images from the VehicleSemanticPart dataset. Then we add occlusion by randomly superimposing a few (one, two, or three) {\em occluders}, which are objects from PASCAL segmentation dataset \cite{Everingham_2010_PASCAL}, onto the images. To prevent confusion, the occluders are not allowed to be vehicles. We also vary the size of the occluders, more specifically the fraction of occluded pixels from all occluders on the target object, is constrained to lie in three ranges $0.2-0.4$, $0.4-0.6$ and $0.6-0.8$. To compute these fractions requires estimating the sizes of the vehicles in the VehicleSemanticPart dataset, which can be done using the 3D object models associated to the images in PASCAL3D+.

\begin{figure}[t]
  \centering
  \begin{subfigure}{1.0\textwidth}
    \begin{minipage}[1.2in]{.3\textwidth}\centering
      \includegraphics[height=1.2in]{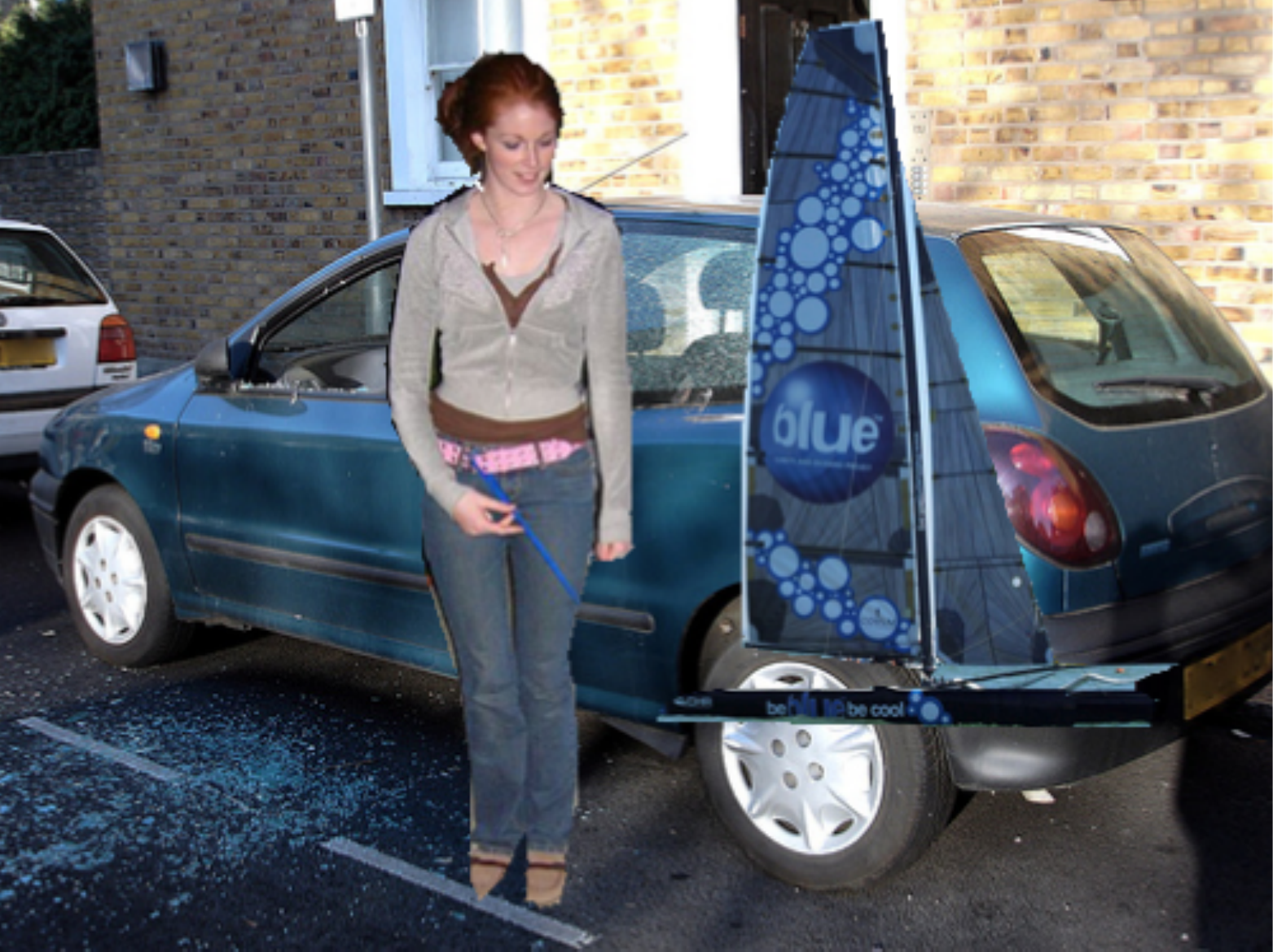}
    \end{minipage}\hfill
    \begin{minipage}[1.2in]{.4\textwidth}\centering
      \includegraphics[height=1.2in]{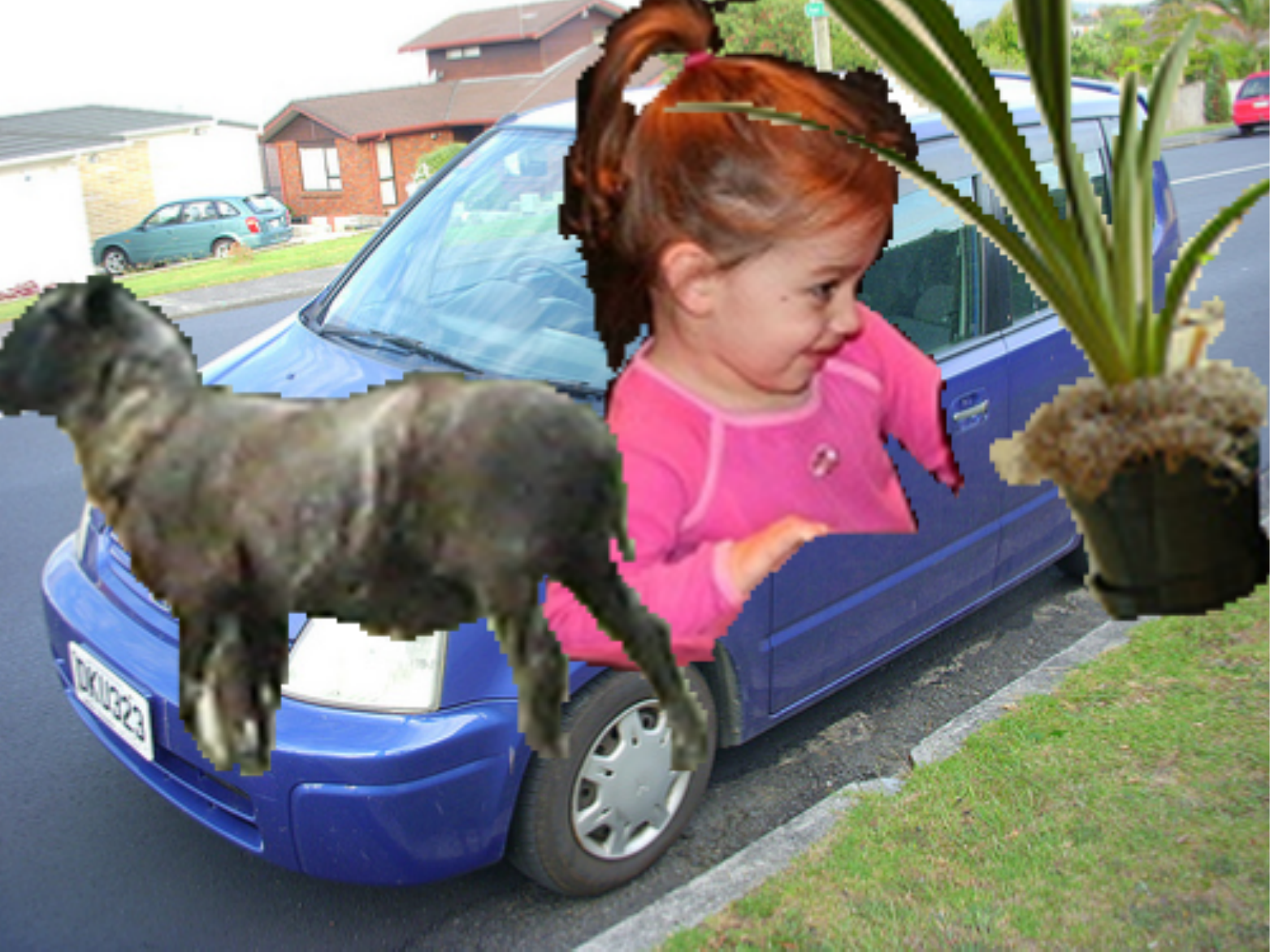}\hfill
    \end{minipage}\hfill
    \begin{minipage}[1.2in]{.3\textwidth}\centering
      \includegraphics[height=1.2in]{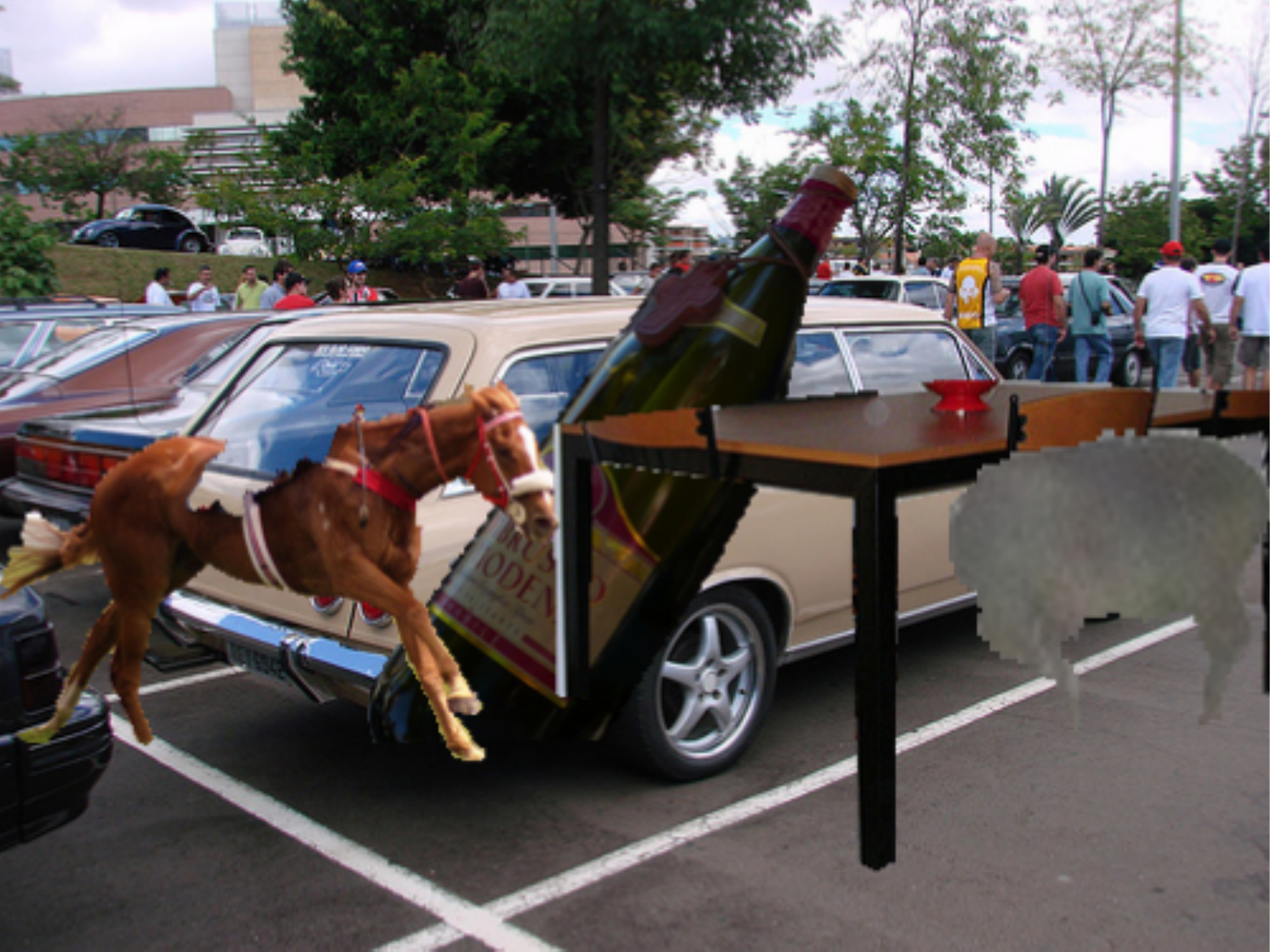}
    \end{minipage}
  \end{subfigure}
  \caption{Examples for occluded cars with occlusion level 1,5 and 9 respectively. Level 1 means 2 occluders with fraction 0.2-0.4 of the object area occluded, level 5 means 3 occluders with fraction 0.4-0.6 occluded, and level 9 means 4 occluders with fraction 0.6-0.8 occluded.}
  \label{fig:occlusion}
\end{figure}


\section{The Visual Concepts \label{sec:visualconcepts}}

We now analyze the activity patterns within deep networks and show that they give rise to visual concepts which correspond to part/subparts of objects and which can be used as building blocks for compositional models. In section~(\ref{sec:K-means}) we obtain the visual concepts by doing $K$-means clustering followed by a merging stage. Next in section~(\ref{sec:online}), we show an alternative unsupervised method which converges to similar visual concepts, but which can be implemented directly by a modified deep network with additional (unsupervised) nodes/neurons. Section~(\ref{sec:visualize}) gives visualization of the visual concepts. Section~(\ref{sec:evaluating}) evaluates them for detecting keypoints and semantic parts.

\subsection{Notation \label{sec:notation}}

We first describe the notation used in this paper. The image is defined on the image lattice and deep network feature vectors are defined on a hierarchy of lattices. The visual concepts are computed at different layers of the hierarchy, but in this paper we will concentrate on the {\it pool-4} layer, of a VGG-16~\cite{Simonyan_2015_Very} trained on ImageNet for classification, and only specify the algorithms for this layer (it is trivial to extend the algorithms to other layers). The groundtruth annotations, i.e. the positions of the semantic parts, are defined on the image lattice. Hence we specify correspondence between points of the image lattice and points on the hierarchical lattices. Visual concepts will never be activated precisely at the location of a semantic part, so we specify a neighborhood of allow for tolerance in spatial location. In the following section, we will use more sophisticated neighborhoods which depend on the visual concepts and the semantic parts.

An image $\mathbf{I}$ is defined on the image lattice $\mathcal{L}_0$ by a set of three-dimensional color vectors $\{\mathbf{I}_q : q \in \mathcal{L}_0\}$. Deep network feature vectors are computed on a hierarchical set of lattices $\mathcal{L}_l$, where $l$ indicates the layer with $\mathcal{L}_l \subset \mathcal{L}_{l-1}$ for $l=0,1,...$. This paper concentrates on the {\it pool-4} layer $\mathcal{L}_4$. We define correspondence between the image lattice and the {\it pool-4} lattice by the mappings $\pi _{0 \mapsto 4}(.)$ from $\mathcal{L}_0$ to $\mathcal{L}_4$  and $\pi _{4 \mapsto 0}(.)$  from $\mathcal{L}_4$ to $\mathcal{L}_0$ respectively. The function $\pi _{4\mapsto 0}(.)$ gives
the exact mapping from $\mathcal{L}_4$ to $\mathcal{L}_0$ (recall the lattices are defined so that $\mathcal{L}_4 \subset \mathcal{L}_0$). Conversely, $\pi _{0\mapsto 4}(q)$ denotes the closest position at the $\mathcal{L}_4$ layer grid that corresponds to $q$, {\em i.e.}, $\pi _{0\mapsto 4}(q) = \arg\min_{p}{\mathrm{Dist}(q, \mathcal{L}_0(p))}$.

The deep network feature vectors at the {\it pool-4} layer are denoted by $\{\mathbf{f}_p: \ p \in \mathcal{L}_4\}$. These feature vectors are computed by $\mathbf{f}_p = \mathbf{f}\!\left(\mathbf{I}_p\right)$, where the function $\mathbf{f}$ is specified by the deep network and $\mathbf{I}_p$ is an image patch, a subregion of the input image $\mathbf{I}$, centered on a point $\pi _{4 \mapsto 0}(p)$ on the image lattice $\mathcal{L}_0$.

The {\it visual concepts} $\{ \mathrm{VC}_v : v \in \mathcal{V}\}$ at the {\it pool-4} layer are specified  by a set  of feature vectors $\{\mathbf{f}_v: v \in \mathcal{V}\}$. They will be learnt by clustering the {\it pool-4} layer feature vectors $\{ \mathbf{f}_p\}$ computed from all the images in the dataset. The activation of a visual concept $\mathrm{VC}_v$ to a feature vector $\mathbf{f}_p$ at $p \in \mathcal{L}_4$ is a decreasing function of $||\mathbf{f}_v- \mathbf{f}_p||$ (see later for more details). The visual concepts will be learnt in an unsupervised manner from a set $\mathcal{T} = \{ \mathbf{I}^n: n=1,...,N\}$ of images, which in this paper will be vehicle images from PASCAL3D+ (with tight bounding boxes round the objects and normalized sizes).

The {\it semantic parts} are given by $\{\mathrm{SP}_s \in \mathcal{S}\}$ and are pre-defined for each vehicle class,  see section~(\ref{sec:Datasets}) and the webpage  \url{http://ccvl.jhu.edu/SP.html}. These semantic parts are annotated on the image dataset $\{ \mathbf{I}^n: n =1 ,..., N\}$ by specifying sets of pixels on the image lattice corresponding to their centers (each semantic part corresponds to an image patch of size $100\times 100$). For a semantic part $\mathrm{SP}_s$, we define $\mathcal{T}_s^+$ to be the set of points $q$ on the image lattices $\mathcal{L}_0$ where they have been labeled. From these labels, we compute a set of points $\mathcal{T}_s^-$  where the semantic part is not present (constrained so that each point in $\mathcal{T}_s^-$ is at least $\gamma$ pixels from every point in $\mathcal{T}_s^+$, where $\gamma$ is chosen to ensure no overlap).

We specify a circular neighborhood ${\mathcal{N}\!\left(q\right)}$ for all points $q$ on the image lattices $\mathcal{L}_0$, given by $\{ p \in \mathcal{L}_0 \ s.t. \ ||p-q|| \leq \gamma_{\mathrm{th}}\}$. This neighborhood is used when we evaluate if a visual concept responds to a semantic part by allowing some uncertainty in the relative location, i.e. we reward a visual concept if it responds {\it near} a semantic part, where nearness is specified by the neighborhood. Hence a visual concept at pixel $p \in \mathcal{L}_4$ responds to a semantic part $s$ at position $q$ provided $||\mathbf{f}_v - \mathbf{f}_p||$ is small and $\pi _{4 \mapsto 0}(p) \in \mathcal{N}(q)$. In this paper the neighborhood radius $\gamma_{\mathrm{th}}$ is set to be 56 pixels for the visual concept experiments in this section, but was extended to 120 for our late work on voting, see section~(\ref{sec:voting}). For voting, we start with this large neighborhood but then
learn more precise neighborhoods which localize the relative positions of visual concept responses to the positions of semantic parts.

\begin{figure}[h]
\centering
\includegraphics[width=0.32\linewidth]{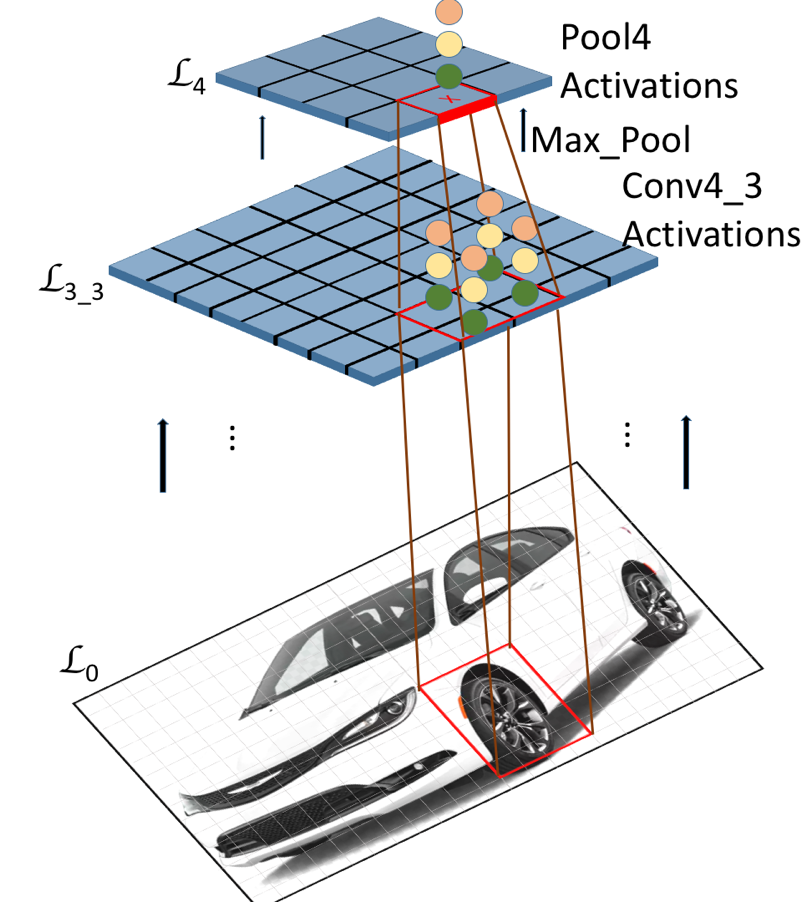}
\includegraphics[width=0.32\linewidth]{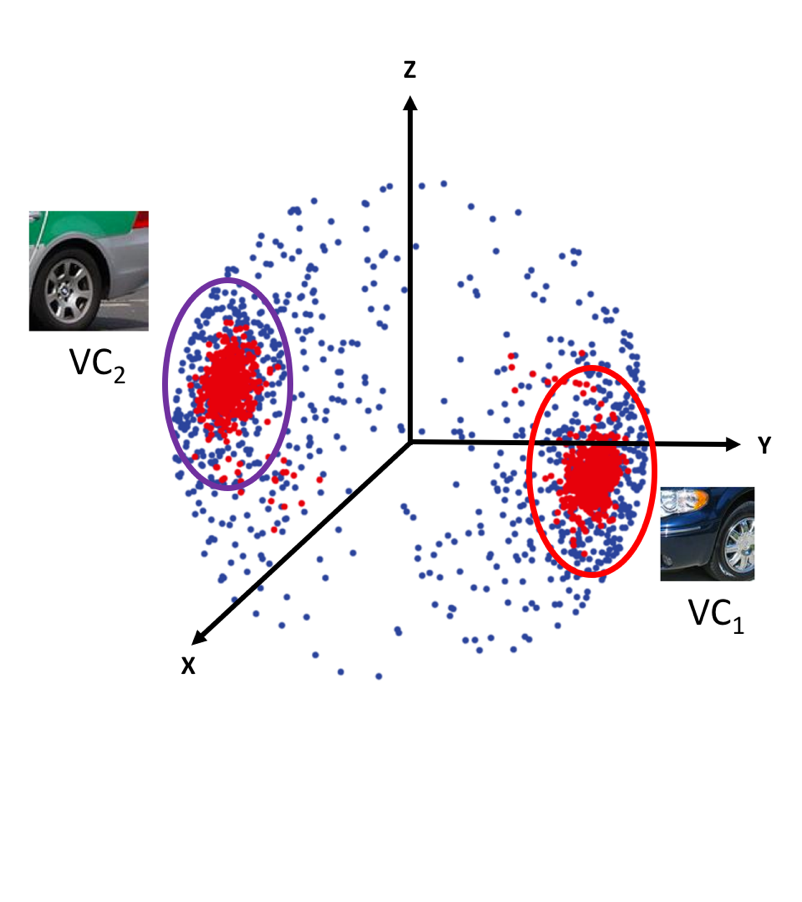}
\includegraphics[width=0.32\linewidth]{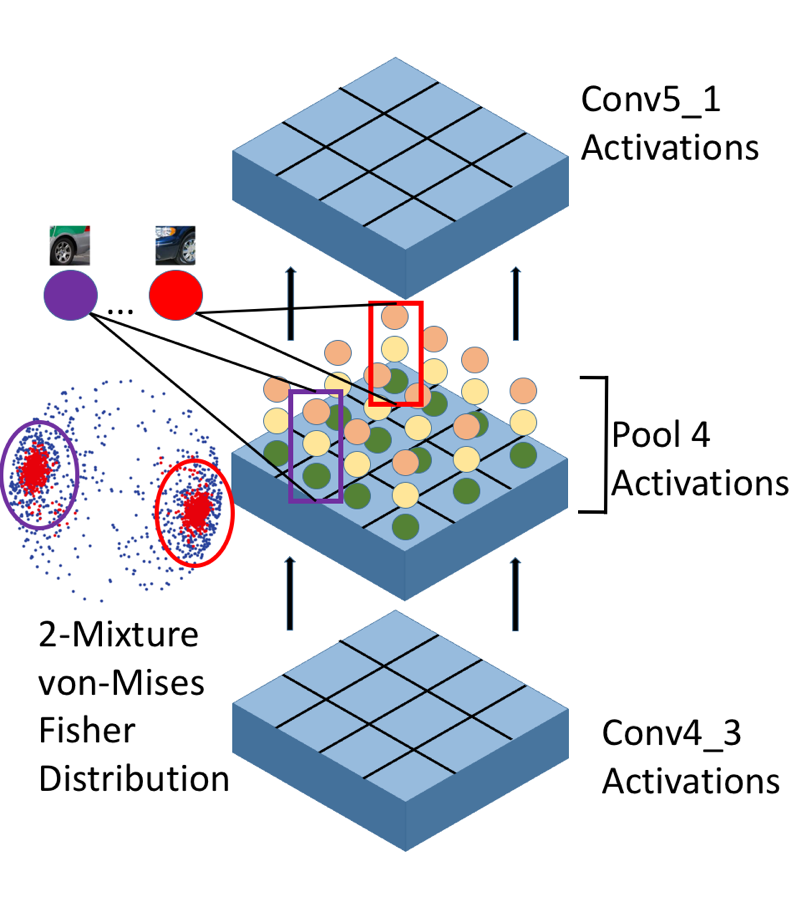}
\caption{Left Panel: The images $\mathbf{I}$ are specified on the image lattice $\mathcal{L}_0$ and a deep network extracts features on a hierarchy of lattices. We concentrate on the features $\{ \mathbf{f}_p\}$ at {\it pool-4} lattice $\mathcal{L}_4$. Projection functions $\pi _{4 \mapsto 0}()$ and $\pi _{0 \mapsto 4}()$ map pixels from $\mathcal{L}_4$ to $\mathcal{L}_0$ and vice versa. We study visual concepts $\mathrm{VC}_v$ at layer $\mathcal{L}_4$ and relate them to semantic parts $\mathrm{SP}_s$ defined on the image lattice. Center Panel: the visual concepts are obtained by clustering the normalized feature vectors $\{ \mathbf{f}_p\}$, using either $K$-means or a mixture of Von Mises-Fisher. Right Panel: extracting visual concepts by a mixture of Von Mises-Fisher is attractive since it incorporates visual concepts within the deep network by adding additional visual concept neurons (vc-neurons).}
\label{fig:vc_neurons}
\end{figure}

\subsection{Learning visual concepts by K-means \label{sec:K-means}}

We now describe our first method for extracting visual concepts which uses the $K$-means algorithm. First we  extract {\em pool-4} layer feature vectors $\{\mathbf{f}_p\}$ using the deep network VGG-16~\cite{Simonyan_2015_Very} from our set of training images $\mathcal{T}$. We normalize them to unit norm, i.e. such that $|\mathbf{f}_p|=1,$ $\forall p$.

Then we use $K$-means++~\cite{arthur2007k}, to cluster the feature vectors into a set $\mathcal{V}$ of visual concepts. Each visual concept $\mathrm{VC}_v$ has an index ${v}\in{\left\{1,2,\ldots,\left|\mathcal{V}\right|\right\}}$ and is specified by its clustering center: ${\mathbf{f}_v}\in{\mathbb{R}^{512}}$ (where $512$ is the dimension of the feature vector at {\it pool-4} of VGG-net.)

In mathematical terms, the $K$-means algorithm attempts to minimize a cost function $F(V,\{\mathbf{f}_v\}) = \sum _{v,p} V_{p,v} ||\mathbf{f}_p - \mathbf{f}_v||^2$, with respect to the assignment variable $V$ and the cluster centers $\{\mathbf{f}_v\}$. The assignment variables $V$ impose {\it hard assignment} so that each feature vector $\mathbf{f}_p$ is assigned to only one visual concept $\mathbf{f}_v$, i.e. for each $p$,  $V_{p,v*}=1$ if $v* = \argmin_v ||\mathbf{f}_p - \mathbf{f}_v||$, and $V_{p,v}= 0$ otherwise. The $K$-means algorithm minimizes the cost function $F(V,\mathbf{f}_v\})$ by minimizing with respect to $V$ and $\{ \mathbf{f}_v\}$ alternatively. The $K$-means++ algorithm initializes $K$-means by taking into account the statistics of the data $\{\mathbf{f}_p\}$. We took a random sample of 100 feature vectors per image as input.

The number of clusters $K = |\mathcal{V}|$ is important in order to get good visual concepts and we found that typically 200 visual concepts were needed for each object. We used two strategies to select a good value for $K$. The first is to specify a discrete set of values for $K$ (64, 128, 256, 512), determine how detection performance depends on $K$, and select the value of $K$ with best performance.  The second is start with an initial set of clusters followed by a merging stage. We use the Davies-Boundin index as a measure for the ``goodness" of a cluster $k$. This is given by $\textit{DB}(k) = \max_{m \neq k} {\frac{\sigma_k + \sigma_m}{||\mathbf{f}_k - \mathbf{f}_l||}}$, where $\mathbf{f}_k$ and $\mathbf{f}_m$ denote the centers of clusters $k$ and $m$ respectively. $\sigma_k$ and $\sigma_m$ denote the average distance of all data points in clusters $k$ and $m$ to their respective cluster centers, i.e., $\sigma_k = \frac{1}{n_k}\sum_{p\in \mathcal{C}_k}||\mathbf{f}_p - \mathbf{f}_k||^2$, where $\mathcal{C}_k=\{ p: V_{p,k}=1\}$ is the set of data points that are assigned to cluster $k$. The Davies-Boundin index take small values for clusters which have small variance of the feature vectors assigned to them, i.e. where $\sum _{p: V_{p,v}=1} |\mathbf{f}_p - \mathbf{f}_v|^2 /n_v$ is small (with $n_v= \sum _{p: V_{p,v}= 1}1$), but which are also well separated from other visual clusters, i.e., $|\mathbf{f}_v - \mathbf{f}_{\mu}|$ is large for all $\mu \neq v$.  We initialize the algorithm with $K$ clusters (e.g., 256, or 512) rank them by the Davies-Boundin index, and merge them in a greedy manner until the index is below a threshold, see our longer report \cite{wang2015unsupervised} for more details. In our experiments we show results with and without cluster merging, and the differences between them are fairly small.

\subsection{ Learning  Visual Concepts by a mixture of von Mises-Fisher distributions \label{sec:online}}

We can also learn visual concepts by modifying the deep network to include additional {\it vc neurons} which are connected to the neurons (or nodes) in the deep network by soft-max layers, see figure~(\ref{fig:vc_neurons}). This is an attractive alternative to $K$-means because it allows visual concepts to be integrated naturally within deep networks, enabling the construction of richer architectures which share both signal and symbolic features (the deep network features and the visual concepts). In this formulation, both the feature vectors $\{\mathrm{f}_p\}$ and the visual concepts $\{\mathbf{f}_v\}$ are normalized so lie on the unit hypersphere. Theoretical and practical advantages of using features defined on the hypersphere are described in \cite{wang2017normface}.

This can be formalized as unsupervised learning where the data, i.e. the feature vectors $\{\mathbf{f}_p\}$, are generated by a mixture of von Mises-Fisher distributions \cite{2017arXiv170604264A}. Intuitively, this is fairly similar to $K$-means (since $K$-means relates to learning a mixture of Gaussians, with fixed isotropic variance, and von-Mises-Fisher relates to an isotropic Gaussian as discussed in the next paragraph). Learning a mixture of von Mises-Fisher can be implemented by a neural network, which is similar to the classic result relating $K$-means to competitive neural networks  \cite{hertz1991introduction}.

The von-Mises Fisher distribution is of form:
\begin{equation} P(\mathbf{f}_p|\mathbf{f}_v) = {\frac{1}{Z(\eta)}} \exp \{ \eta \mathbf{f}_p \cdot \mathbf{f}_v\},\end{equation}
\noindent where $\eta$ is a constant, $Z(\eta)$ is a normalization factor, and $\mathbf{f}_p$ and $\mathbf{f}_v$ are both unit vectors. This requires us to normalize the weights of the visual concepts,  so that $|\mathbf{f}_v|=1, \forall v$, as well as the feature vectors $\{\mathbf{f}_p\}$. Arguably it is more natural to normalize both the feature vectors and the visual concepts (instead of only normalizing the feature vectors as we did in the last section). There ia a simple relationship between von-Mises Fisher and isotropic Gaussian distributions. The square distance term $||\mathbf{f}_p - \mathbf{f}_v||$ becomes equal to $2(1 - \mathbf{f}_p \cdot \mathbf{f}_v)$ if $||\mathbf{f}_p|| = ||\mathbf{f}_v||=1$. Hence an isotropic Gaussian distribution on $\mathbf{f}_p$ with mean $\mathbf{f}_v$ reduces to von Mises-Fisher if both vectors are required to lie on the unit sphere.

Learning a mixture of von Mises-Fisher can be formulated in terms of neural networks where the cluster centers, i.e. the visual concepts, are specified by \emph{visual concept neurons}, or \emph{vc-neurons}, with weights $\{\mathbf{f}_v\}$. This can be shown as follows. Recall that $V_{p,v}$ is the assignment variable between a feature vector $\mathbf{f}_p$ and each visual concept  $\mathbf{f}_v$, and denote the set of assignment by $\mathbf{V}_p = \{V_{p,v}: v \in \mathcal{V}\}$ During learning this assignment variable $\mathbf{V}_p$ is unknown but we can use the EM algorithm, which involving replacing the $V_{p,v}$ by a distribution $q_v(p)$ for the probability that $V_{p,v}=1$.

The EM algorithm can be expressed in terms of minimizing the following free energy function with respect to $\{q_v\}$ and $\{\mathbf{f}_v\}$:
\begin{equation} \mathcal{F}(\{q_v\},\{\mathbf{f}_v\}) = - \sum _p \{ \sum _{v} q_v(p) \log P(\mathbf{f_p}, \mathbf{V}_p| \{\mathbf{f}_v\}) + \sum _v q_v(p) \log q_v(p)\},\end{equation}
The update rules for the assignments and the visual concepts are respectively:
\begin{equation} q_{v}^t = {\frac{\exp \{ \eta \mathbf{f}_p \cdot \mathbf{f}_v^t\}}{\sum _{\mu} \exp \{ \eta \mathbf{f}_p \cdot \mathbf{f}_{\mu}^t\}}} \ \ \ \mathbf{f}_v ^{t+1} = \mathbf{f}_v ^t - \epsilon \eta q_v^t \mathbf{f}_v + \eta, \end{equation}
\noindent where $\eta$ is a Lagrange multiplier to enforce $|\mathbf{f}_v ^{t+1}|=1$.

\emph{Observe that the update for the assignments are precisely the standard neural network soft-max operation} where the $\{\mathbf{f}_v\}$ are interpreted as the weights of {\it vc-neurons} and the neurons compete so that their activity sums to $1$ \cite{hertz1991introduction}. After the learning algorithm has converged this softmax activation indicates soft-assignment of a feature vector $\mathbf{f}_p$ to the vc-neurons. This corresponds to the network shown in figure~(\ref{fig:vc_neurons}).

This shows that learning visual concepts can be naturally integrated into a deep network. This can either be done, as we discussed here, after the weights of a deep network have been already learnt. Or alternatively, the weights of the vc neurons can be learnt at the same times as the weights of the deep network. In the former case, the algorithm simply minimizes a cost function:
\begin{equation} - \sum _v {\frac{\exp \{ \eta \mathbf{f}_p \cdot \mathbf{f}_v\}}{\sum _{\nu} \exp \{ \eta \mathbf{f}_p \cdot \mathbf{f}_{\nu}\}}} \log \sum _{\nu} \exp \{\eta \mathbf{f}_p \cdot \mathbf{f}_{\nu}\},\label{eq:cost}\end{equation}
\noindent which can be obtained from the EM free energy by solving for the $\{q_v\}$ directly in terms of $\{\mathbf{f}_v\}$, to obtain $q_{v} = {\frac{\exp \{ \eta \mathbf{f}_p \cdot \mathbf{f}_v\}}{\sum _{\nu} \exp \{ \eta \mathbf{f}_p \cdot \mathbf{f}_{\nu}\}}}$, and substituting this back into the free energy.

Learning the visual concepts and the deep network weights simultaneously is performed by adding this cost function to the standard penalty function for the deep network. This is similar to the regularization method reported in \cite{parsimonious}. Our experiments showed that learning the visual concepts and the deep network weights simultaneously risks collapsing the features vectors and the visual concepts to a trivial solution.


\subsection{Visualizing the Visual Concepts \label{sec:visualize}}

\begin{figure}[t!]
\centering
\includegraphics[width=1\linewidth]{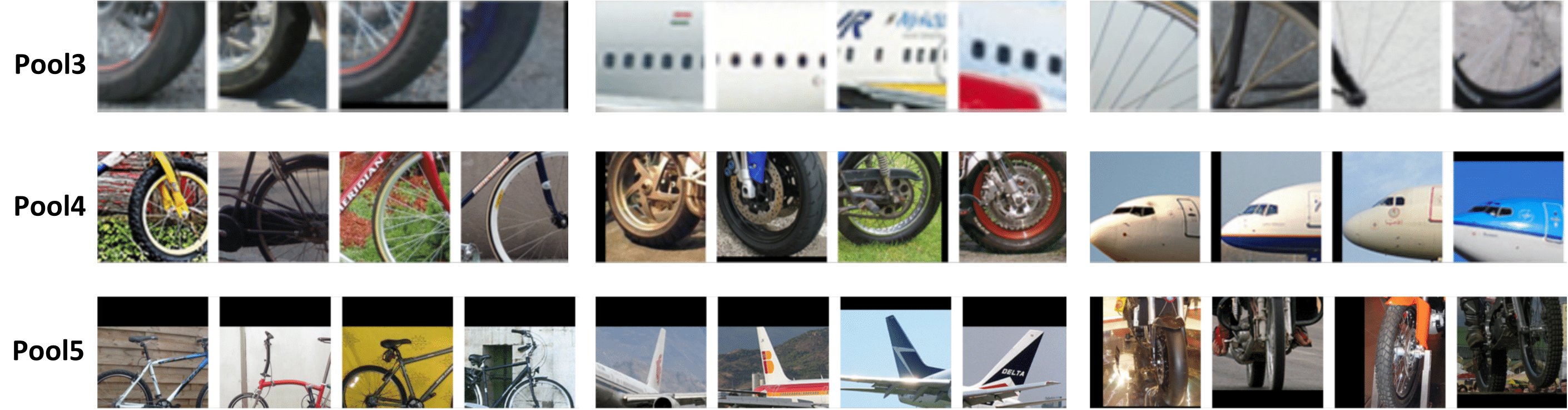}
\caption{\small This figure shows example visual concepts on different object categories from the VehicleSemanticPart dataset, for layers \texttt{pool3}, \texttt{pool4} and \texttt{pool5}. Each row visualizes three visual concepts with four example patches, which are randomly selected from a pool of the closest 100 image patches. These visual concepts are visually and semantically tight. We can easily identify the semantic meaning and parent object class.}
\label{fig:concept_viz}
\end{figure}

We visualize the visual concepts by observing the image patches which are assigned to each cluster by $K$-means. We observe that the image patches for each visual concept roughly correspond to semantic parts of the object classes with larger parts at higher pooling levels, see figure~(\ref{fig:concept_viz}). This paper concentrates on visual concepts at the {\it pool-4} layer because these are most similar in scale to the semantic parts which were annotated. The  visual concepts at {\it pool-3} layer are at a smaller scale, which makes it harder to evaluate them using the semantic parts. The visual concepts at the {\it pool-5} layer have two disadvantages. Firstly, they correspond to regions of the image which are larger than the semantic parts. Secondly, their effective receptive field sizes which were significantly smaller than their theoretical receptive field size (determined by using deconvolution networks to find which regions of the image patches affected the visual concept response) which meant that they appeared less tight when we visualized them (because only the central regions of the image patches activated them, so the outlying regions can vary a lot).

\begin{figure}[t!]
\centering
\begin{minipage}{.96\textwidth}
\centering
\includegraphics[width=0.96\textwidth]{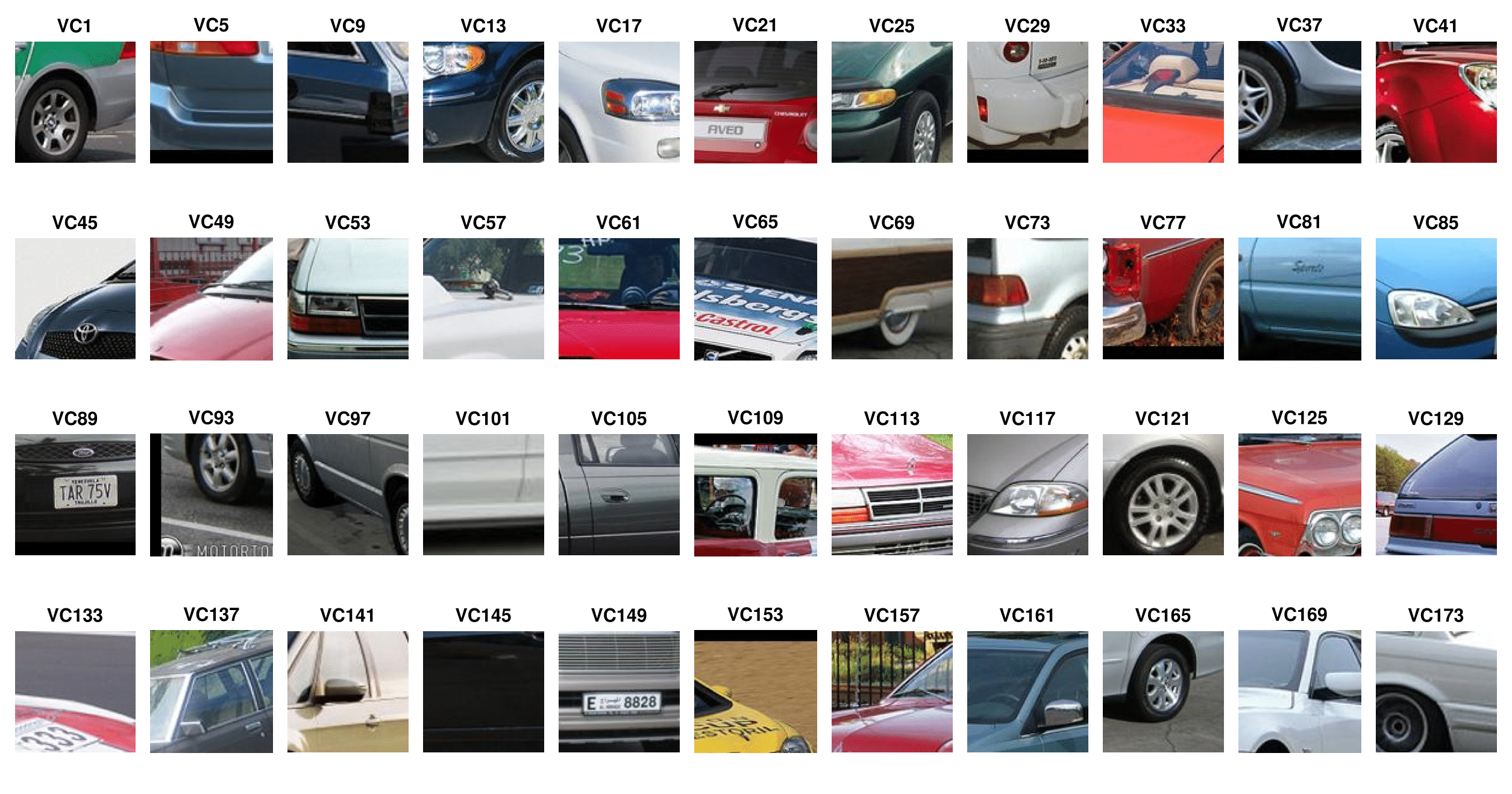}
\caption{\small This figure shows that visual concepts typically give a dense coverage of the different parts of the objects, in this case a car.}
\label{fig:VisualConceptCoverage}
\end{minipage}
\end{figure}

\begin{figure}
\begin{minipage}{.99\textwidth}
\begin{subfigure}{.49\textwidth}
  \centering
  \begin{minipage}[t]{.2\textwidth}
    \centering
    \includegraphics[width=.9999\linewidth,height=.9999\linewidth]{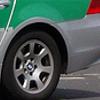}
    \includegraphics[width=.9999\linewidth,height=.9999\linewidth]{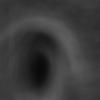}
    \includegraphics[width=.9999\linewidth,height=.9999\linewidth]{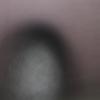}
  \end{minipage}
  \begin{minipage}[t]{.2\textwidth}
    \centering
    \includegraphics[width=.9999\linewidth,height=.9999\linewidth]{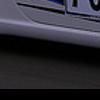}
    \includegraphics[width=.9999\linewidth,height=.9999\linewidth]{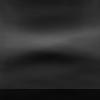}
    \includegraphics[width=.9999\linewidth,height=.9999\linewidth]{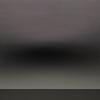}
  \end{minipage}
  \begin{minipage}[t]{.2\textwidth}
    \centering
    \includegraphics[width=.9999\linewidth,height=.9999\linewidth]{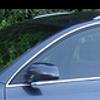}
    \includegraphics[width=.9999\linewidth,height=.9999\linewidth]{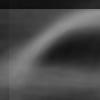}
    \includegraphics[width=.9999\linewidth,height=.9999\linewidth]{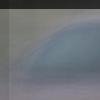}
  \end{minipage}
  \begin{minipage}[t]{.2\textwidth}
    \centering
    \includegraphics[width=.9999\linewidth,height=.9999\linewidth]{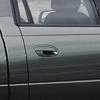}
    \includegraphics[width=.9999\linewidth,height=.9999\linewidth]{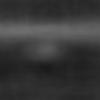}
    \includegraphics[width=.9999\linewidth,height=.9999\linewidth]{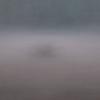}
  \end{minipage}
  \caption{\textit{pool4} visual concept}
\end{subfigure}\hfill
\begin{subfigure}{.49\textwidth}
  \centering
  \begin{minipage}[t]{.2\textwidth}
    \centering
    \includegraphics[width=.9999\linewidth,height=.9999\linewidth]{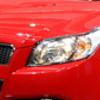}
    \includegraphics[width=.9999\linewidth,height=.9999\linewidth]{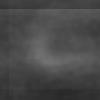}
    \includegraphics[width=.9999\linewidth,height=.9999\linewidth]{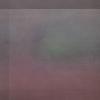}
  \end{minipage}
  \begin{minipage}[t]{.2\textwidth}
    \centering
    \includegraphics[width=.9999\linewidth,height=.9999\linewidth]{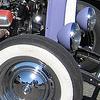}
    \includegraphics[width=.9999\linewidth,height=.9999\linewidth]{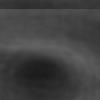}
    \includegraphics[width=.9999\linewidth,height=.9999\linewidth]{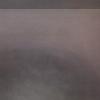}
  \end{minipage}
  \begin{minipage}[t]{.2\textwidth}
    \centering
    \includegraphics[width=.9999\linewidth,height=.9999\linewidth]{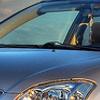}
    \includegraphics[width=.9999\linewidth,height=.9999\linewidth]{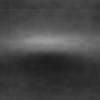}
    \includegraphics[width=.9999\linewidth,height=.9999\linewidth]{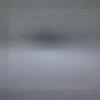}
  \end{minipage}
  \begin{minipage}[t]{.2\textwidth}
    \centering
    \includegraphics[width=.9999\linewidth,height=.9999\linewidth]{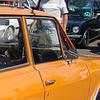}
    \includegraphics[width=.9999\linewidth,height=.9999\linewidth]{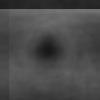}
    \includegraphics[width=.9999\linewidth,height=.9999\linewidth]{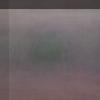}
  \end{minipage}
  \caption{\textit{pool4} single filter}
\end{subfigure}
  \vspace{-0.3cm}
  \caption{\small This figure shows four visual concept examples (left four) and four single filter examples (right four). The top row contains example image patches, and the bottom two rows contain average edge maps and intensity maps respectively obtained using top 500 patches.Observe that the means of the visual concepts are sharper than the means of the single filters for both average edge maps and intensity maps, showing visual concepts capture patterns more tightly than the single filters.
  }
  \label{fig:avg}
\end{minipage}
\end{figure}

Our main findings are that the {\it pool-4} layer visual concepts are: (i) very tight visually, in the sense that image patches assigned to the same visual concept tend to look very similar which is illustrated in figure~(\ref{fig:avg})
and (ii) give a dense coverage of the objects as illustrated in figure~(\ref{fig:VisualConceptCoverage}). Demonstrating tightness and coverage of the visual concepts is hard to show in a paper,  so we have created a webpage \url{http://ccvl.jhu.edu/cluster.html} where readers can see this for themselves by looking at many examples. To illustrate visual tightness we show the average of the images and edge maps of the best 500 patches for a few visual concepts, in figure~(\ref{fig:avg}), and compare them to the averages for single filters. These averages show that the clusters are tight.

\subsection{Evaluating visual concepts for detecting keypoints and semantic parts \label{sec:evaluating}}


We now evaluate the visual concepts as detectors of keypoints and semantic parts (these are used for evaluation only and not for training). Intuitively, a visual concept $\mathrm{VC}_v$ is a good detector for a semantic part $\mathrm{SP}_s$ if the visual concept ``fires" near most occurrences of the semantic part, but rarely fires in regions of the image where the semantic part is not present. More formally, a  visual concept $\mathrm{VC}_v$ fires at a position $p \in \mathcal{L}_4$ provided $|\mathbf{f}_p -\mathbf{f}_v| < T$, where $T$ is a threshold (which will be varied to get a precision-recall curve). Recall that for each semantic part $s \in \mathcal{S}$ we have a set of points $\mathcal{T}^+_s$ where the semantic part occurs, and a set $\mathcal{T}^-_s$ where it does not. A visual concept $v$ has a true positive detection for $q \in \mathcal{T}^+_s$ if $\min _{p \in \mathcal{N}(q)} |\mathbf{f}_p -\mathbf{f}_v| < T$. Similarly it has a false positive if the same condition holds for $s \in \mathcal{T}^-_s$.  As the threshold $T$ varies we obtain a precision-recall curve and report the average precision.

We also consider variants of this approach by seeing if a visual concept is able to detect several semantic parts. For example, a visual concept which fires on semantic part ``side-window" is likely to also respond to the semantic parts ``front-window" and ``back-window", and hence should be evaluated by its ability to detect all three types of windows. Note that if a visual concept is activated for several semantic parts then it will tend to have fairly bad average precision for each semantic part by itself (because the detection for the other semantic parts will count as false positives).

We show results for two versions of visual concept detectors, with and without merging indicated by {\bf S-VC$_{M}$} and {\bf S-VC$_K$} respectively. We also implemented two baseline methods for comparison methods. This first baseline uses the magnitude of a single filter response, and is denoted by {\bf S-F}. The second baseline is strongly supervised and trains a support vector machines (SVM) taking the feature vectors at the {\it pool-4} layer as inputs,  notated by {\bf SVM-F}. The results we report are based on obtaining visual concepts by $K$-means clustering (with or without merging), but we obtain similar results if we use the von Mises-Fisher online learning approach,

\subsubsection{Strategies for Evaluating Visual Concepts}

There are two possible strategies for evaluating visual concepts for detection. The first strategy is to crop the semantic parts to $100 \times 100$ image patches, $\{P_1,P_2,\cdots,P_n\}$, and provide some negative patches $\{P_{n+1},P_{n+2},\cdots,P_{n+m}\}$ (where the semantic parts are not present). Then we can calculate the response of $VC_v$ by calculating $Res_i=||\mathbf{f}_i-\mathbf{f}_v||$. By sorting those $m+n$ candidates by the VC responses, we can get a list $P_{k_1},P_{k_2},\cdots,P_{k_{m+n}}$ in which $k_1,k_2,\cdots,k_{m+n}$ is a permutation of $1,2,\cdots,m+n$. By varying the threshold from $Res_{k_1}$ to $Res_{k_{m+n}}$, we can calculate the precision and recall curve. The average precision can be obtained by averaging the precisions at different recalls.

This first evaluation strategy is commonly used for detection tasks in computer vision, for example to evaluate edge detection, face detection, or more generally object detection. But it is not suitable for our purpose because different semantic parts of object can be visually similar (e.g., the front, side, and back windows of cars). Hence the false positives of a semantic part detector will often be other parts of the same object, or a closely related object (e.g., car and bus semantic parts can be easily confused). In practice, we will want to detect semantic parts of an object when a large region of the object is present. The first evaluation strategy does not take this into account, unless it is modified so that the set of negative patches are carefully balanced to include large numbers of semantic parts from the same object class.

Hence we use a second evaluation strategy, which is to crop the object bounding boxes from a set of images belonging to the same object class to make sure they only have limited backgrounds, and then select image patches which densely sample these bounding boxes. This ensures that the negative image patches are strongly biased towards being other semantic parts of the object (and corresponds to the typical situation where we will want to detect semantic parts in practice). It will, however, also introduce image patches which partly overlap with the semantic parts and hence a visual concept will typically respond multiple times to the same semantic part, which we address by using non-maximum suppression. More precisely, at each position a visual concept $\mathrm{VC}_v$ will have a response $||\mathbf{f}_p-\mathbf{f}_v||$. Next we apply non-maximum suppression to eliminate overlapping detection results, i.e. if two activated visual concepts have sufficiently similar responses, then the visual concept with weaker response (i.e. larger $||\mathbf{f}_p-\mathbf{f}_v||$) will be suppressed by the stronger one. Then we proceed as for the first strategy, i.e. threshold the responses, calculate the false positives and false negatives, and obtain the precision recall curve by varying the threshold. (A technical issue for the second strategy is that some semantic parts may be close together, so after non-maximum suppression, some semantic parts may be missed since the activated visual concepts could be suppressed by nearby stronger visual concept responses. Then a low AP will be obtained, even though this visual concept might be very for detecting that semantic part).

\subsubsection{Evaluating single Visual Concepts for Keypoint detection}

To evaluate how well visual concepts detect keypoints we tested them for vehicles in the PASCAL3D+ dataset and compared them to the two baseline methods, see table~(\ref{tab:avg_pre}). Visual concepts and single filters are unsupervised so we evaluate them for each keypoint, by by selecting the visual concept, or single filter, which have best detection performance, as measured by average precision. Not surprisingly, the visual concepts outperformed the single filters but were less successful than the supervised SVM approach. This is not surprising, since visual concepts and single filters are unsupervised. The results are fairly promising for visual concepts since  for almost all semantic parts we either found a visual concept that was fairly successful at detecting them. We note that typically several visual concepts were good at detecting each keypoint, recall that there are roughly 200 visual concepts but only approximately 10 keypoints. This suggests that we would get better detection results by combining visual concepts, as we will do later in this paper.

\begin{table}[t!]
\centering
\tabcolsep=0.04cm
\begin{tabular}{ |c||c|c|c|c|c|c|c|c||c|c|c|c|c|c||c|c|c|c|c| }
\hline
        & \multicolumn{8}{|c||}{\textbf{Car}} & \multicolumn{6}{|c||}{\textbf{Bicycle}} & \multicolumn{5}{|c|}{\textbf{Motorbike}} \\
\hline
        & 1 & 2 & 3 & 4 & 5 & 6 & 7 & {\scriptsize mAP} & 1 & 2 & 3 & 4 & 5 & {\scriptsize mAP} & 1 & 2 & 3 & 4 & {\scriptsize mAP} \\
\hline
{\bf S-F}      & .86  &  .44  &  .30  &  .38  &  .19  &  .33   &  .13  & .38     &  .23  &  .67  &  .25  &  .43  &  .43  &  .40      &  .26  &  .60   &  .30  &  .22  & .35  \\
{\bf S-VC$_{M}$}   & .92  &  .51  &  .27  &  .41  &  .36  &  .46   &  .18  & .45     &  .32  &  .78  &  .30  &  .55  &  .57  &  .50      &  .35  &  .75   &  .43  &  .25  & .45  \\
{\bf S-VC$_K$}    & .94  &  .51  &  .32  &  .53  &  .36  &  .48   &  .22  & .48     &  .35  &  .80  &  .34  &  .53  &  .63  &  .53      &  .40  &  .76   &  .44  &  .43  & .51  \\
\hline
{\bf SVM-F}     & .97  &  .65  &  .37  &  .76  &  .45  &  .57   &  .30  & .58     &  .37  &  .80  &  .34  &  .71  &  .64  &  .57      &  .37  &  .77   &  .50  &  .60  & .56  \\
\hline
\end{tabular}

\begin{tabular}{ |c||c|c|c|c|c|c|c||c|c|c|c|c|c||c|c|c|c|c|c| }
\hline
       & \multicolumn{7}{|c||}{\textbf{Bus}} & \multicolumn{6}{|c||}{\textbf{Train}} & \multicolumn{6}{|c|}{\textbf{Aeroplane}} \\
\hline
       & 1 & 2 & 3 & 4 & 5 & 6 & {\scriptsize mAP} & 1 & 2 & 3 & 4 & 5 & {\scriptsize mAP} & 1 & 2 & 3 & 4 & 5 & {\scriptsize mAP} \\
\hline
{\bf S-F}      & .45  &  .42   &  .23  &  .38  &  .80   &  .22  & .42     & .39  &  .33   &  .24  &  .16  &  .15  & .25      & .41  &  .25  &  .22  &  .13  &  .31  &  .26  \\
{\bf S-VC$_{M}$}   & .41  &  .59   &  .26  &  .29  &  .86   &  .51  & .49     & .41  &  .30   &  .30  &  .28  &  .24  & .30      & .21  &  .47  &  .31  &  .16  &  .34  &  .30  \\
{\bf S-VC$_K$}    & .41  &  .51   &  .26  &  .33  &  .86   &  .52  & .48     & .42  &  .32   &  .30  &  .28  &  .25  & .32      & .31  &  .47  &  .31  &  .20  &  .35  &  .33  \\
\hline
{\bf SVM-F}     & .74  &  .70   &  .52  &  .63  &  .90   &  .61  & .68     & .71  &  .49   &  .50  &  .36  &  .39  & .49      & .72  &  .60  &  .50  &  .32  &  .49  &  .53  \\
\hline
\end{tabular}
\caption{\small AP values for keypoint detection on PASCAL3D+ dataset for six object categories. Visual concepts, with {\bf S-VC}$_m$ or without {\bf S-VC}$_K$ merging, achieves much higher results than the single filter baseline {\bf S-F}. The SVM method {\bf SVM-F} does significantly better, but this is not surprising since it is supervised. The keypoint number-name mapping is provided below. Cars -- 1: wheel 2: wind shield 3: rear window 4: headlight 5: rear light 6: front 7: side; Bicycle --1: head center 2: wheel 3: handle 4: pedal 5: seat; Motorbike -- 1: head center 2: wheel 3: handle 4: seat; Bus -- 1: front upper corner 2: front lower corner 3: rear upper corner 4: rear lower corner 5: wheel 6: front center; Train -- 1: front upper corner 2: front lower corner 3: front center 4: upper side 5: lower side; Aeroplane -- 1: nose 2: upper rudder 3: lower rudder 4: tail 5: elevator and wing tip.}
\label{tab:avg_pre}
\end{table}

\subsubsection{Evaluating single Visual Concepts for Semantic Part detection}

The richer annotations on VehicleSemanticPart allows us to get better understanding of visual concepts. We use the same evaluation strategy as for keypoints but report results here only for visual concepts with merging {\bf S-VC} and for single filters {\bf S-F}. Our longer report \cite{wang2015unsupervised} gives results for different variants of visual concepts (e.g., without merging, with different values of $K$, etc.) but there is no significant difference. Our main findings are: (i) that visual concepts do significantly better than single filters, and (ii) for every semantic part there is a visual concept that detects it reasonably well. These results, see Table~(\ref{tab:compart_PR}), support our claim that the visual concepts give a dense coverage of each object (since the semantic parts label almost every part of the object). Later in this paper, see table~(\ref{Tab:NoOcclusion}) (known scale), we compare the performance of SVMs to visual concepts for detecting semantic parts. This comparison uses a tougher evaluation criterion, based on the interSection over union (IOU) \cite{Everingham_2010_PASCAL}, but the relative performance of the two methods is roughly the same as for the keypoints.

\begin{table*}[t!]
\centering
\begin{subtable}{\linewidth}
\centering
\tabcolsep=0.04cm
\begin{tabular}{ |c||c|c|c|c|c|c|c|c|c|c|c|c|c|c|c|c|c|c|c| }
\hline
\textbf{} & 1 & 2 & 3 & 4 & 5 & 6 & 7 & 8 & 9 & 10 & 11 & 12 & 13 & 14 & 15 & 16 & 17 & 18 \\
\hline
{\bf S-F}      & .86  &  .87  &  .84  &  .68  &  .70  &  .83  &  .83   & .82  & .72  &  .18  &  .22  &  .49   &  .33  &  .25  &  .22  &  .18   &  .37  &  .18  \\
{\bf S-VC$_{M}$}       & .94  &  .97  &  .94  &  .93  &  .94  &  .95  &  .94   & .92  & .94  &  .42  &  .48  &  .58  &  .45  &  .48  &  .46  &  .54   &  .60  &  .34  \\
\hline \hline
\textbf{} & 19 & 20 & 21 & 22 & 23 & 24 & 25 & 26 & 27 & 28 & 29 & 30 & 31 & 32 & 33 & 34 & 35 & {\scriptsize mAP} \\
\hline
{\bf S-F}  & .28  &  .13  &  .16  &  .14  &  .30  &  .20    &  .16   & .17  & .15  &  .27  &  .19  &  .25  &  .18  &  .19  &  .16  &  .07   &  .07  &  .36 \\
{\bf S-VC$_{M}$}   & .38  &  .19  &  .26  &  .37  &  .42  &  .32    &  .26   & .40  & .29  &  .40  &  .18  &  .41  &  .53  &  .31  &  .57  &  .36   &  .26  &  .53 \\
\hline
\end{tabular}
\caption{Car}
\end{subtable}

\begin{subtable}{\linewidth}
\centering
\tabcolsep=0.04cm
\begin{tabular}{ |c||c|c|c|c|c|c|c|c|c|c|c|c|c|c|c|c|c|c|c| }
\hline
\textbf{} & 1 & 2 & 3 & 4 & 5 & 6 & 7 & 8 & 9 & 10 & 11 & 12 & 13 & 14 & 15 & 16 & 17 & {\scriptsize mAP} \\
\hline
{\bf S-F}      & .32  &  .42  &  .08  &  .61  &  .58  &  .42  &  .49   & .31   & .23   &  .09  &  .07  &  .20   &  .12  &  .29  &  .29  &  .07   &  .09  &  .28    \\
{\bf S-VC$_{M}$}       & .26  &  .21  &  .07  &  .84  &  .61  &  .44   & .63   & .42   & .34   &  .15  &  .11  &  .44   &  .23  &  .59  &  .65  &  .08   &  .15  &  .37 \\
\hline
\end{tabular}
\caption{Aeroplane}
\end{subtable}

\begin{subtable}{\linewidth}
\centering
\tabcolsep=0.04cm
\begin{tabular}{ |c||c|c|c|c|c|c|c|c|c|c|c|c|c|c|c|c|c|c|c| }
\hline
\textbf{} & 1 & 2 & 3 & 4 & 5 & 6 & 7 & 8 & 9 & 10 & 11 & 12 & 13 & {\scriptsize mAP} \\
\hline
\small
{\bf S-F}      & {.77}  &  {.84}   & {.89}  & {.91}  &  {.94}  &  {.92}  &  {.94}  &  {.91}  &  {.91}  &  {.56}  &  {.53}   &  {.15}  &  {.40}  &  {.75}  \\
{\bf S-VC$_{M}$}   &  {.91}  &  {.95}   & {.98}  & {.96}  &  {.96}  &  {.96}  &  {.97}  &  {.96}  &  {.97}  &  {.73}  &  {.69}   &  {.19}  &  {.50}  &  {.83}  \\
\hline
\end{tabular}
\caption{Bicycle}
\end{subtable}

\begin{subtable}{\linewidth}
\centering
\tabcolsep=0.04cm
\begin{tabular}{ |c||c|c|c|c|c|c|c|c|c|c|c|c|c|c|c|c|c|c|c| }
\hline
\textbf{} & 1 & 2 & 3 & 4 & 5 & 6 & 7 & 8 & 9 & 10 & 11 & 12 & {\scriptsize mAP} \\
\hline
{\bf S-F}      &  .69  &  .46  &  .76   & .67  & .66  &  .57  &  .54  &  .70  &  .68  &  .25  &  .17  &  .22   &  .53   \\
{\bf S-VC$_{M}$}   &  .89  &  .64  &  .89   & .77  & .82  &  .63  &  .73  &  .75  &  .88  &  .39  &  .33  &  .29   &  .67   \\
\hline
\end{tabular}
\caption{Motorbike}
\end{subtable}

\begin{subtable}{\linewidth}
\centering
\tabcolsep=0.04cm
\begin{tabular}{ |c||c|c|c|c|c|c|c|c|c|c|c|c|c|c|c|c|c|c|c| }
\hline
\textbf{} & 1 & 2 & 3 & 4 & 5 & 6 & 7 & 8 & 9 & {\scriptsize mAP} \\
\hline
{\bf S-F}      & .90  & .40  &  .49  &  .46  &  .31  &  .28  &  .36  &  .38   &  .31  &  .43  \\
{\bf S-VC$_{M}$}   & .93  & .64  &  .69  &  .59  &  .42  &  .48  &  .39  &  .32   &  .27  &  .53  \\
\hline
\end{tabular}
\caption{Bus}
\end{subtable}

\begin{subtable}{\linewidth}
\centering
\tabcolsep=0.04cm
\begin{tabular}{ |c||c|c|c|c|c|c|c|c|c|c|c|c|c|c|c|c|c|c|c| }
\hline
\textbf{} & 1 & 2 & 3 & 4 & 5 & 6 & 7 & 8  & 9 & 10 & 11 & 12 & 13 & 14 & {\scriptsize mAP} \\
\hline
{\bf S-F}      &  .58  &  .07  &  .20  &  .15   & .21  & .15  &  .27  & .43  &  .17  &  .27  &  .16  &  .25  &  .17   &  .10  &  .23 \\
{\bf S-VC$_{M}$}   &  .66  &  .50  &  .32  &  .28   & .24  & .15  &  .33  & .72   &  .36  &  .41  &  .27  &  .45  &  .27   &  .47  &  .39   \\
\hline
\end{tabular}
\caption{Train}
\end{subtable}
\caption{\small The AP values for semantic part detection on VehicleSemanticPart dataset for cars (top), aeroplanes (middle) and bicycle (bottom). We see that visual concepts {\bf S-VC}$_m$  achieves much higher AP than single filter method {\bf S-F}. The semantic part names are provided in the supplementary material.}
\label{tab:compart_PR}
\end{table*}

\subsubsection{What do the other visual concepts do?}

The previous experiments have followed the ``best single visual concept evaluation". In other words, for each semantic part (or keypoint), we find the single visual concept which best detects that semantic part. But this ignores three issues: (I)
A visual concept may respond to more than one semantic parts (which means that its AP for one semantic part is small because the others are treated as false positives). (II) There are far more visual concepts than semantic parts, so our previous evaluations have not told us what all the visual concepts are doing (e.g., for cars there are roughly 200 visual concepts but only 39 semantic parts, so we are only reporting results for twenty percent of the visual concepts). (III) Several visual concepts may be good for detecting the same semantic part, so combining them may lead to better semantic part detection.

In this section we address issues (I) and (II), which are closely related, and leave the issue of combining visual concepts to the next section. We find that some visual concepts do respond better to more than one semantic part, in the sense that their AP is higher when we evaluate them for detecting a small subset of semantic parts. More generally, we find that almost all the visual concepts respond to a small number of semantic parts (one, two, three, or four) while most of the limited remaining visual concepts appear to respond to frequently occuring ``backgrounds", such as the sky in the airplane images.

\begin{figure}[h!]
	\centering
	\begin{subfigure}{.30\linewidth}
{\includegraphics[width=\linewidth]{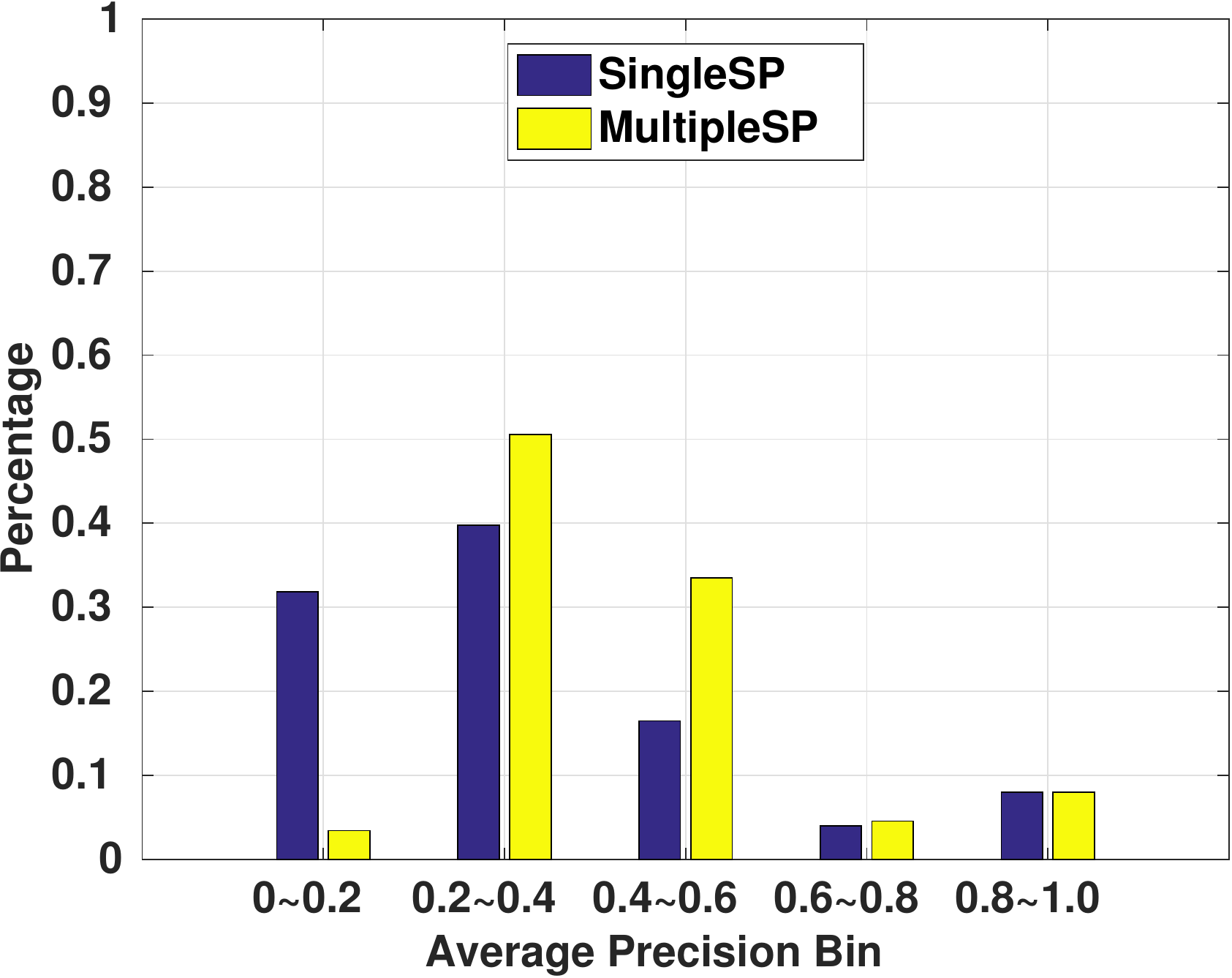}}
\caption{Car}
\end{subfigure}
	\begin{subfigure}{.30\linewidth}
{\includegraphics[width=\linewidth]{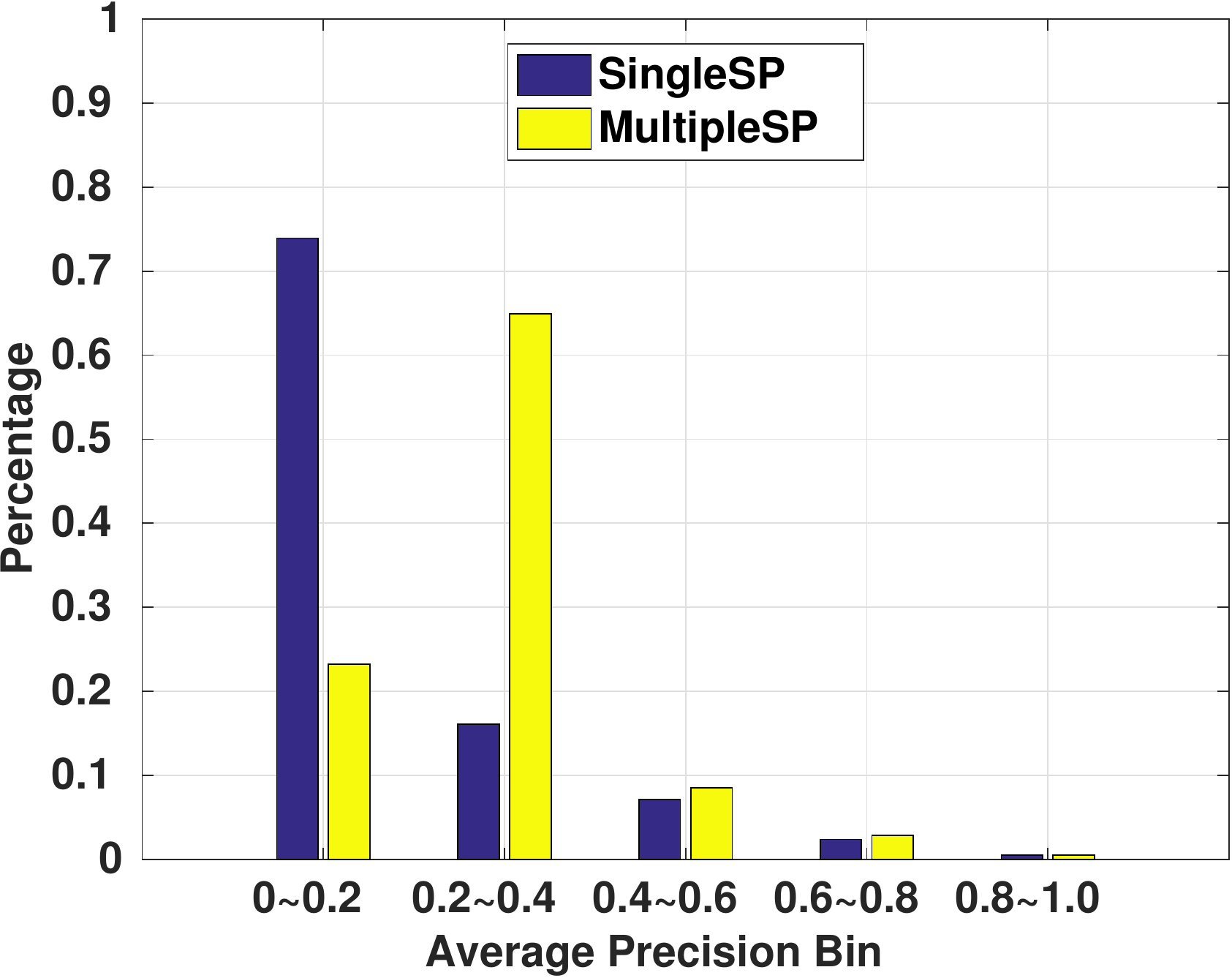}}
\caption{Aeroplane}
\end{subfigure}
\begin{subfigure}{.30\linewidth}
{\includegraphics[width=\linewidth]{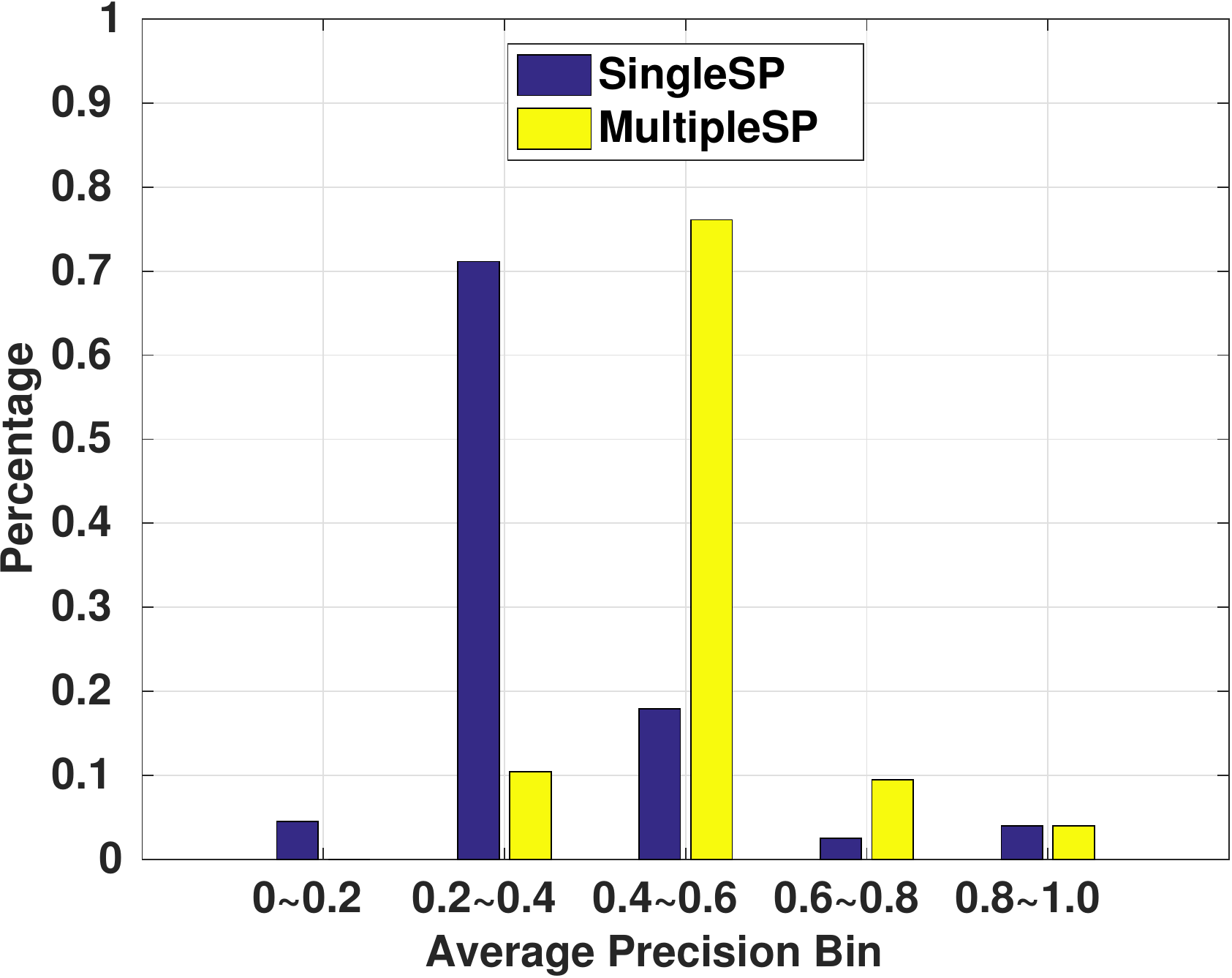}}
\caption{Bus}
\end{subfigure}
	\begin{subfigure}{.30\linewidth}
{\includegraphics[width=\linewidth]{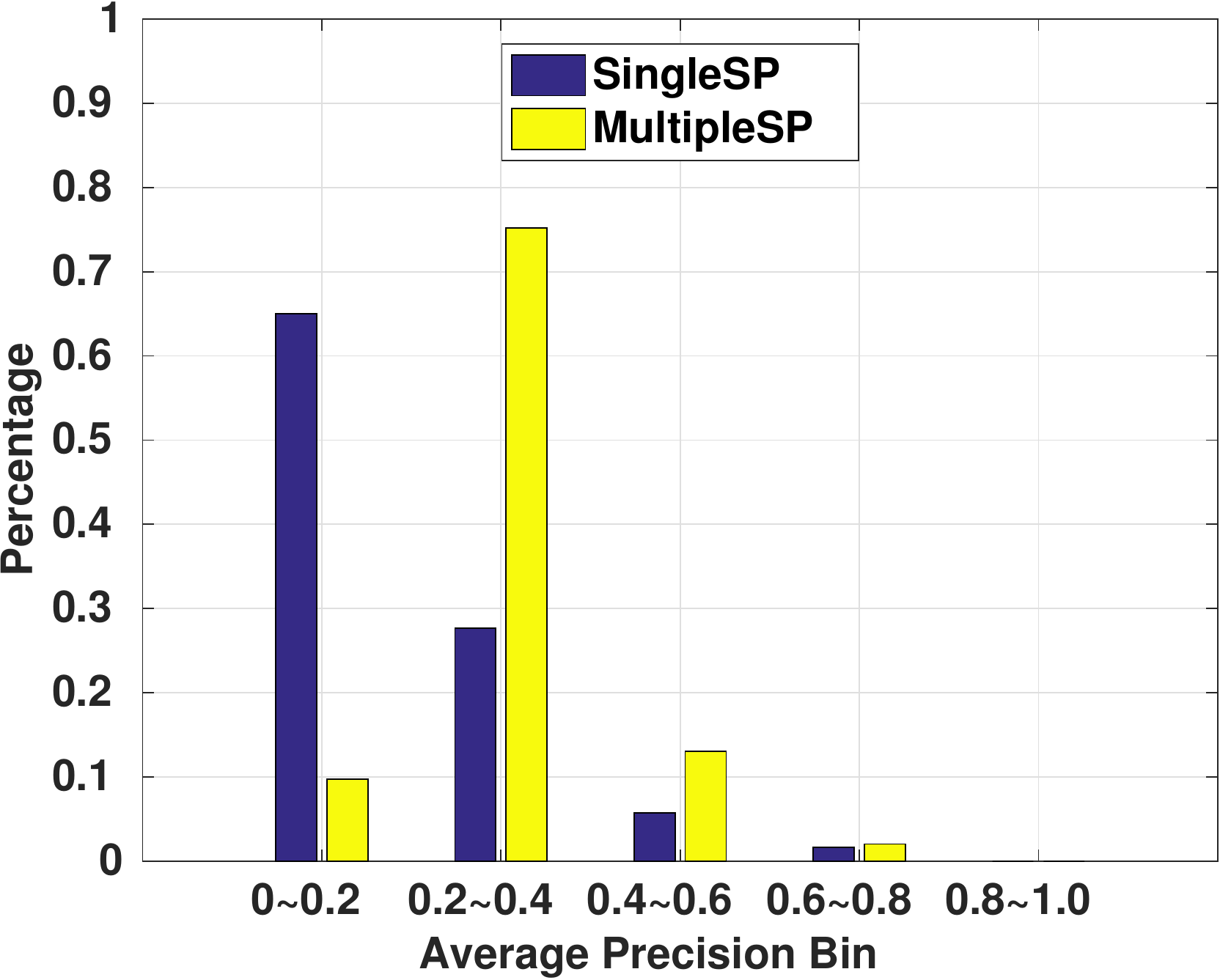}}
\caption{train}
\end{subfigure}
\begin{subfigure}{.30\linewidth}
{\includegraphics[width=\linewidth]{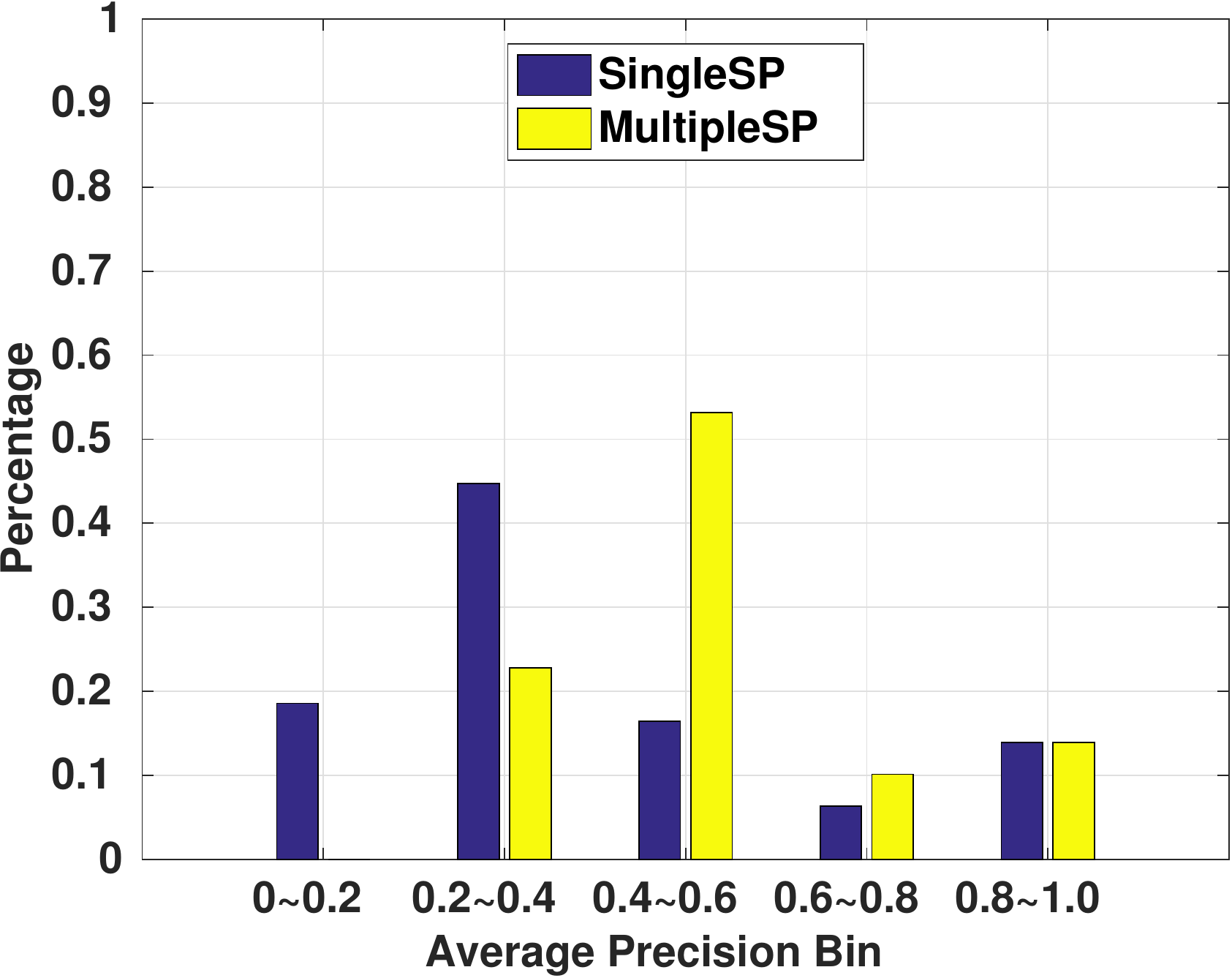}}
\caption{Bicycle}
\end{subfigure}
\begin{subfigure}{.30\linewidth}
{\includegraphics[width=\linewidth]{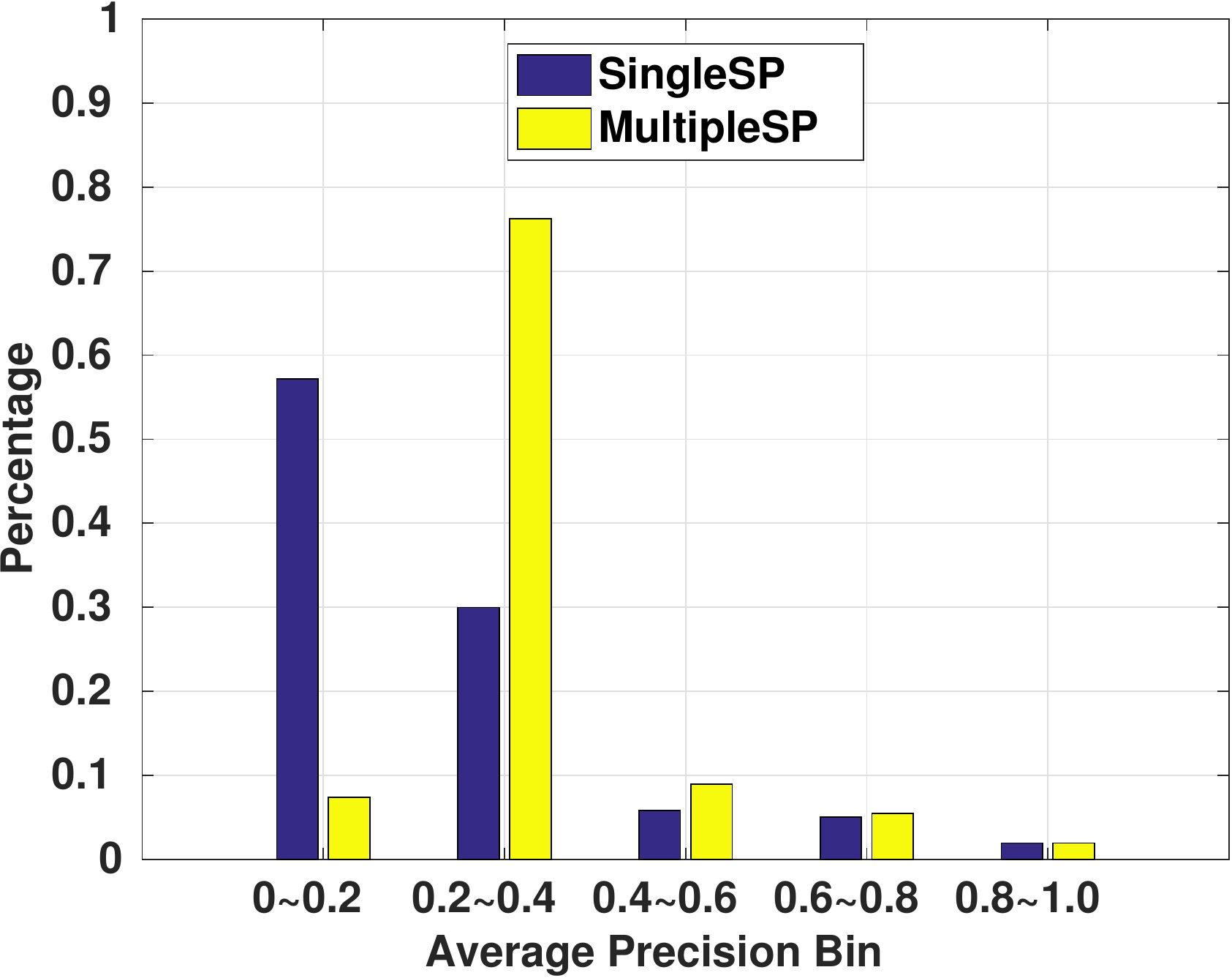}}
\caption{Motorbike}
\end{subfigure}
	\caption{\small The histograms of the AP responses for all visual concepts. For ``SingleSP", we evaluate each visual concept by reporting its AP for its best semantic part (the one it best detects). Some visual concepts have very low APs when evaluated in this manner. For ``MultipleSP", we allow each visual concept to detect a small subset of semantic parts (two, three, or four) and report the AP for the best subset (note, the visual concept is penalized if it does not detect all of them). The APs rise considerably using the MultipleSP evaluation, suggesting that some of the visual concepts detect a subset of semantic parts. The remaining visual concepts with very low APs correspond to the background.}
	\label{fig:many}
\end{figure}

\begin{figure}[t!]
	\centering
	\includegraphics[width=.8\linewidth]{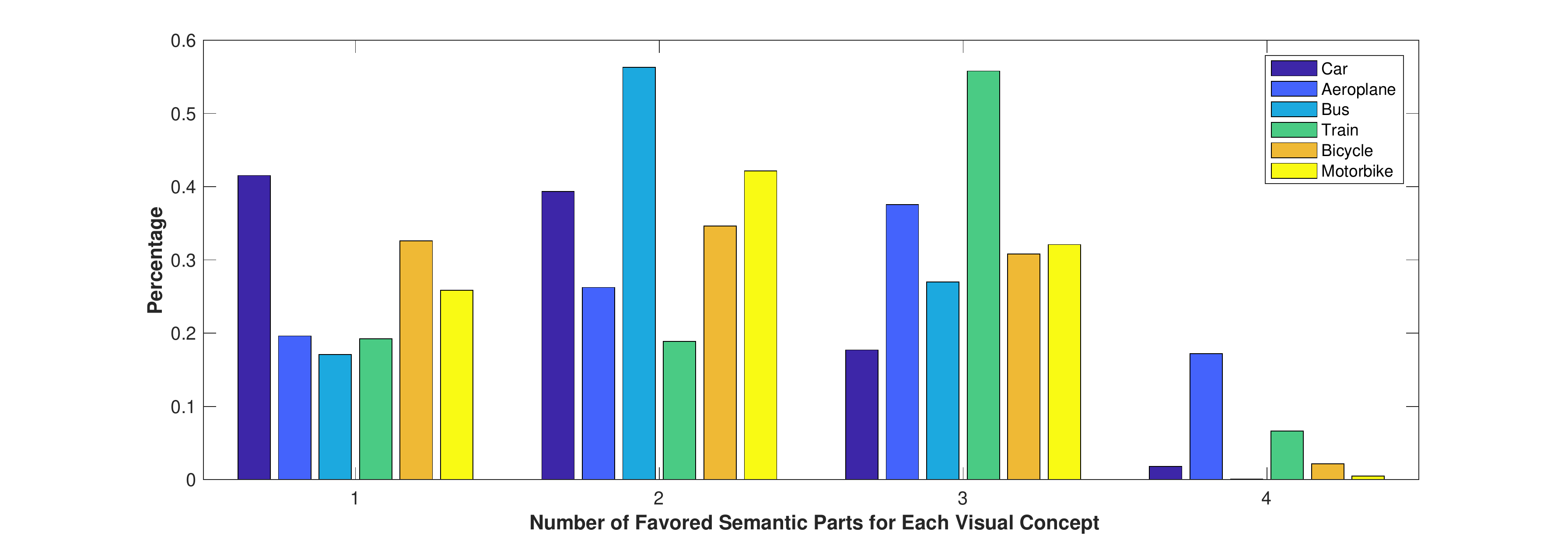}
	\caption{\small Distribution of number of semantic parts favored by each visual concept for 6 objects. Most visual concepts like one, two, or three semantic parts.}
	\label{fig:sp_num}
	\vspace{-0.5cm}
\end{figure}

\begin{figure}[t!]
  \centering
\begin{minipage}{.45\linewidth}
  \includegraphics[width=0.85cm,height=0.85cm]{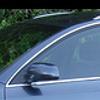}
  \includegraphics[width=0.85cm,height=0.85cm]{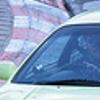}
  \includegraphics[width=0.85cm,height=0.85cm]{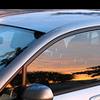}
  \includegraphics[width=0.85cm,height=0.85cm]{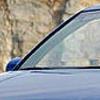}
  \includegraphics[width=0.85cm,height=0.85cm]{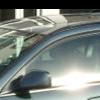}
  \includegraphics[width=0.85cm,height=0.85cm]{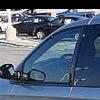}
\end{minipage}
\hspace{0.5cm}
\begin{minipage}{.055\linewidth}
  \includegraphics[width=0.85cm,height=0.85cm]{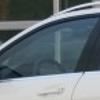}
\end{minipage}
\hspace{0.5cm}
\begin{minipage}{.055\linewidth}
  \includegraphics[width=0.85cm,height=0.85cm]{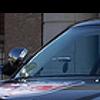}
\end{minipage} \\ \vspace{.2cm}
\begin{minipage}{.45\textwidth}
  \includegraphics[width=0.85cm,height=0.85cm]{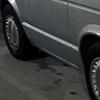}
  \includegraphics[width=0.85cm,height=0.85cm]{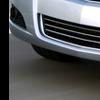}
  \includegraphics[width=0.85cm,height=0.85cm]{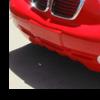}
  \includegraphics[width=0.85cm,height=0.85cm]{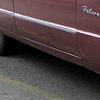}
  \includegraphics[width=0.85cm,height=0.85cm]{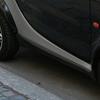}
  \includegraphics[width=0.85cm,height=0.85cm]{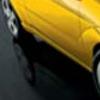}
\end{minipage}
\hspace{0.5cm}
\begin{minipage}{.055\linewidth}
  \includegraphics[width=0.85cm,height=0.85cm]{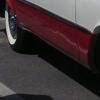}
\end{minipage}
\hspace{0.5cm}
\begin{minipage}{.055\linewidth}
  \includegraphics[width=0.85cm,height=0.85cm]{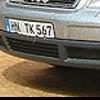}
\end{minipage}
\caption{\small This figure shows two examples of visual concepts that correspond to multiple semantic parts. The first six columns show example patches from that visual concept, and the last two columns show prototypes from two different corresponding semantic parts. We see that the visual concept in the first row corresponds to ``side window" and ``front window", and the visual concept in the second row corresponds to ``side body and ground" and ``front bumper and ground". Note that these semantic parts look fairly similar.}
\label{fig:dual_SP}
\vspace{-0.1cm}
\end{figure}

\begin{figure}[t!]
  \centering
\begin{minipage}{.9\textwidth}
\centering
  \includegraphics[width=0.96cm,height=0.96cm]{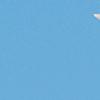}
  \includegraphics[width=0.96cm,height=0.96cm]{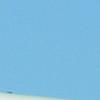}
  \includegraphics[width=0.96cm,height=0.96cm]{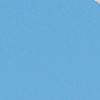}
  \includegraphics[width=0.96cm,height=0.96cm]{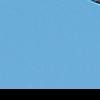}
  \includegraphics[width=0.96cm,height=0.96cm]{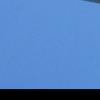}
  \includegraphics[width=0.96cm,height=0.96cm]{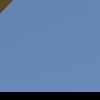}
  \includegraphics[width=0.96cm,height=0.96cm]{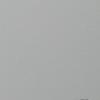}
  \includegraphics[width=0.96cm,height=0.96cm]{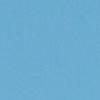}
\end{minipage} \\ \vspace{.2cm}
\begin{minipage}{.9\textwidth}
\centering
  \includegraphics[width=0.96cm,height=0.96cm]{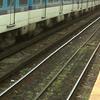}
  \includegraphics[width=0.96cm,height=0.96cm]{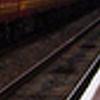}
  \includegraphics[width=0.96cm,height=0.96cm]{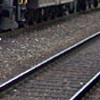}
  \includegraphics[width=0.96cm,height=0.96cm]{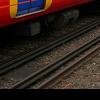}
  \includegraphics[width=0.96cm,height=0.96cm]{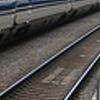}
  \includegraphics[width=0.96cm,height=0.96cm]{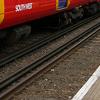}
  \includegraphics[width=0.96cm,height=0.96cm]{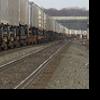}
  \includegraphics[width=0.96cm,height=0.96cm]{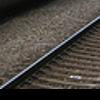}
\end{minipage} \\ \vspace{.2cm}
\begin{minipage}{.9\textwidth}
\centering
  \includegraphics[width=0.96cm,height=0.96cm]{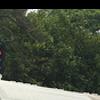}
  \includegraphics[width=0.96cm,height=0.96cm]{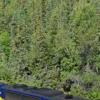}
  \includegraphics[width=0.96cm,height=0.96cm]{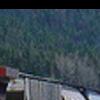}
  \includegraphics[width=0.96cm,height=0.96cm]{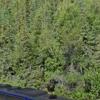}
  \includegraphics[width=0.96cm,height=0.96cm]{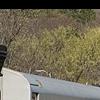}
  \includegraphics[width=0.96cm,height=0.96cm]{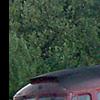}
  \includegraphics[width=0.96cm,height=0.96cm]{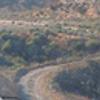}
  \includegraphics[width=0.96cm,height=0.96cm]{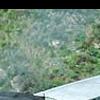}
\end{minipage}
\caption{\small The first row shows a visual concept corresponding to sky from the aeroplane class, and the second and third rows are visual concepts corresponding to railway track and tree respectively from the train class.}
\label{fig:background}
\end{figure}

We proceed as follows. Firstly, for each visual concept we determine which semantic part it best detects, calculate the AP, and plot the histogram as shown in Figure \ref{fig:many} (SingleSP). Secondly, we allow each visual concept to select a small subset -- two, three, or four -- of semantic parts that it can detect (penalizing it if it fails to detect all of them). We measure how well the visual concept detects this subset using AP and plot the same histogram as before. This generally shows much better performance than before. From Figure \ref{fig:many} (MultipleSP) we see that the histogram of APs gets shifted greatly to the right, when taking into account the fact that one visual concept may correspond to one or more semantic parts. Figure \ref{fig:sp_num} shows the percentage of how many semantic parts are favored by each visual concept. If a visual concept responds well to two, or more, semantic parts, this is often because those parts are visually similar, see Figure \ref{fig:dual_SP}. For example, the semantic parts for car windows are visually fairly similar. For some object classes, particularly aeroplanes and trains, there remain some visual concepts with low APs even after allowing multiple semantic parts. Our analysis shows that many of these remaining visual concepts are detecting background (e.g., the sky for the aeroplane class, and railway tracks or coaches for the train class), see Figure \ref{fig:background}. A few other visual concepts have no obvious interpretation and are probably due to limitations of the clustering algorithm and the CNN features.

Our results also imply that there are several visual concepts for each semantic part. In other work, in preparation, we show that these visual concepts typically correspond to different subregions of the semantic parts (with some overlap) and that combining them yields better detectors for all semantic parts with a mean gain of 0.25 AP.

\subsection{Summary of Visual Concepts \label{sec:summary_vc}}

This section showed that visual concepts were represented as internal representations of deep networks and could be found either by $K$-means or by an alternative method where the visual concepts could be treated as hidden variables attached to the deep network. The visual concepts  were visually tight, in the sense that image patches assigned to the same visual concept looked similar. Almost all the visual concepts corresponded to semantic parts of the objects (one, tow, three, or four), several corresponded to background regions (e.f., sky), but a few visual concepts could not be interpreted. More results and analysis can be found in our longer report \cite{wang2015unsupervised} and the webpage \url{http://ccvl.jhu.edu/cluster.html}.

In particular, we observed that visual concepts significantly outperformed single filters for detecting semantic parts, but performed worse than supervised methods, such as support vector machines (SVMs), which took the same {\it pool-4} layer features as input. This worse performance was due to several factors.  Firstly, supervised approaches have more information so they generally tend to do better on discriminative tasks. Secondly, several visual concepts tended to fire on the same semantic part, though in different locations, suggesting that the visual concepts correspond to subparts of the semantic part. Thirdly, some visual concepts tended to fire on several semantic parts, or in neighboring regions.

These findings suggest that visual concepts are suitable as building blocks for pattern theoretic models. We will evaluate this in the next section where we construct a simple compositional-voting model which combines several visual concepts to detect semantic parts. 

\section{Combining Visual Concepts for Detecting Semantic Parts with Occlusion\label{sec:voting}}

The previous section suggests that visual concepts roughly correspond to object parts/subparts and hence could be used as building blocks for compositional models. To investigate this further we study the precise spatial relationships between the  visual concepts and the semantic parts. Previously we only studied whether the visual concepts were active within a fixed size circular window centered on the semantic parts. Our more detailed studies show that visual concepts tend to fire in much more localized positions relative to the centers of the semantic parts.

This study enables us to develop a compositional voting algorithm where visual concepts combine together to detect semantic parts. Intuitively the visual concepts provide evidence for subparts of the semantic parts, taking into account their spatial relationships, and we can detect the semantic parts by combining this evidence. The evidence comes both from visual concepts which overlap with the semantic parts, but also from visual concepts which lie outside the semantic part but provide contextual evidence for it. For example, a visual concept which responds to a wheel gives contextual evidence for the semantic part ``front window" even though the window is some distance away from the wheel. We must be careful, however, about allowing contextual evidence because this evidence can be unreliable if there is occlusion.  For example, if we fail to detect the wheel of the car (i.e. the evidence for the wheel is very low) then this  should not prevent us from detecting a car window because the car wheel might be occluded. But conversely, if can detect the wheel then this should give contextual evidence which helps detect the window. 

\begin{figure}[h]
  \centering
\includegraphics[width=\linewidth]{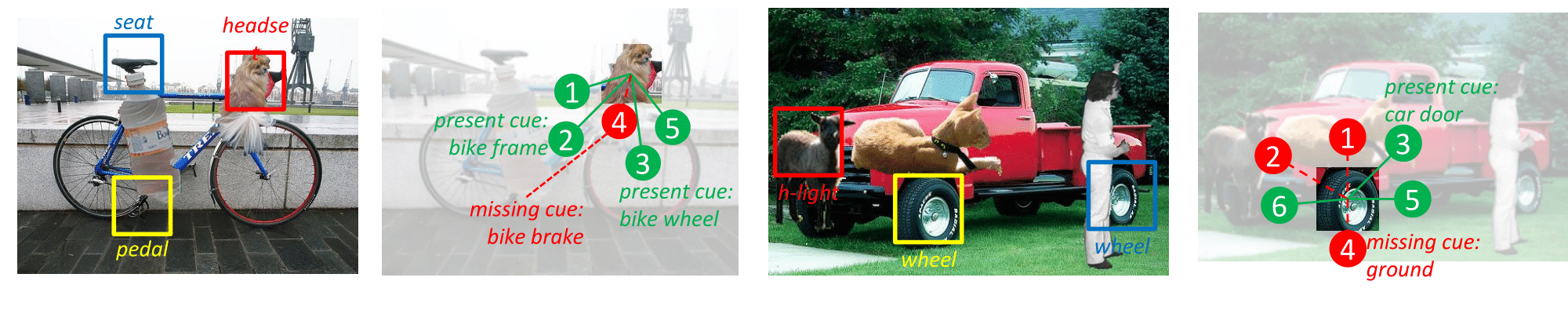}
\caption{The goal is to detect semantic parts of the object despite occlusion, but without being trained on the occlusion. Center panel: yellow indicates unoccluded part, blue means partially occluded part, while red means fully occluded part. Right panel: Visual concepts (the circles) give evidence for semantic parts. They can be switched off automatically in presence of occlusion (the red circles) so that only the visual concepts that fire (green circles) give positive evidence for the presence of the parts. This robustness to occlusion if characteristic of our compositional model and enables it to outperform fully supervised deep networks.}
\label{fig:generative_mixture}
\end{figure}

A big advantage of our compositional voting approach is that it gives a flexible and adaptive way to deal with context which is particularly helpful when there is occlusion, see figure~(\ref{fig:generative_mixture}). If the evidence for a visual concept is below a threshold, then the algorithm automatically switches off the visual concept so that it does not supply strong negative evidence. Intuitively ``absence of evidence is not evidence of absence". This allows a flexible and adaptive way to include the evidence from visual concepts if it is supportive, but to automatically switch it off if it is not. We note that humans have this ability to adapt flexibly to novel stimuli, but computer vision and machine learning system rarely do since the current paradigm is to train and test algorithms on data which are presume to be drawn from the same underlying distribution. Our ability to use information flexibly means that our system is in some cases able to detect a semantic part from context even if the part itself is completely occluded and, by analyzing which visual concepts have been switched off, to determine if the semantic part itself is occluded. In later work, this switching off can be extended to deal with the ``explaining away" phenomena, highlighted by work in cognitive science \cite{kanizsa1976convexity}.

\subsection{Compositional Voting: detecting Semantic Parts with and without Occlusion}

\begin{figure}[h]
\includegraphics[width=\linewidth]{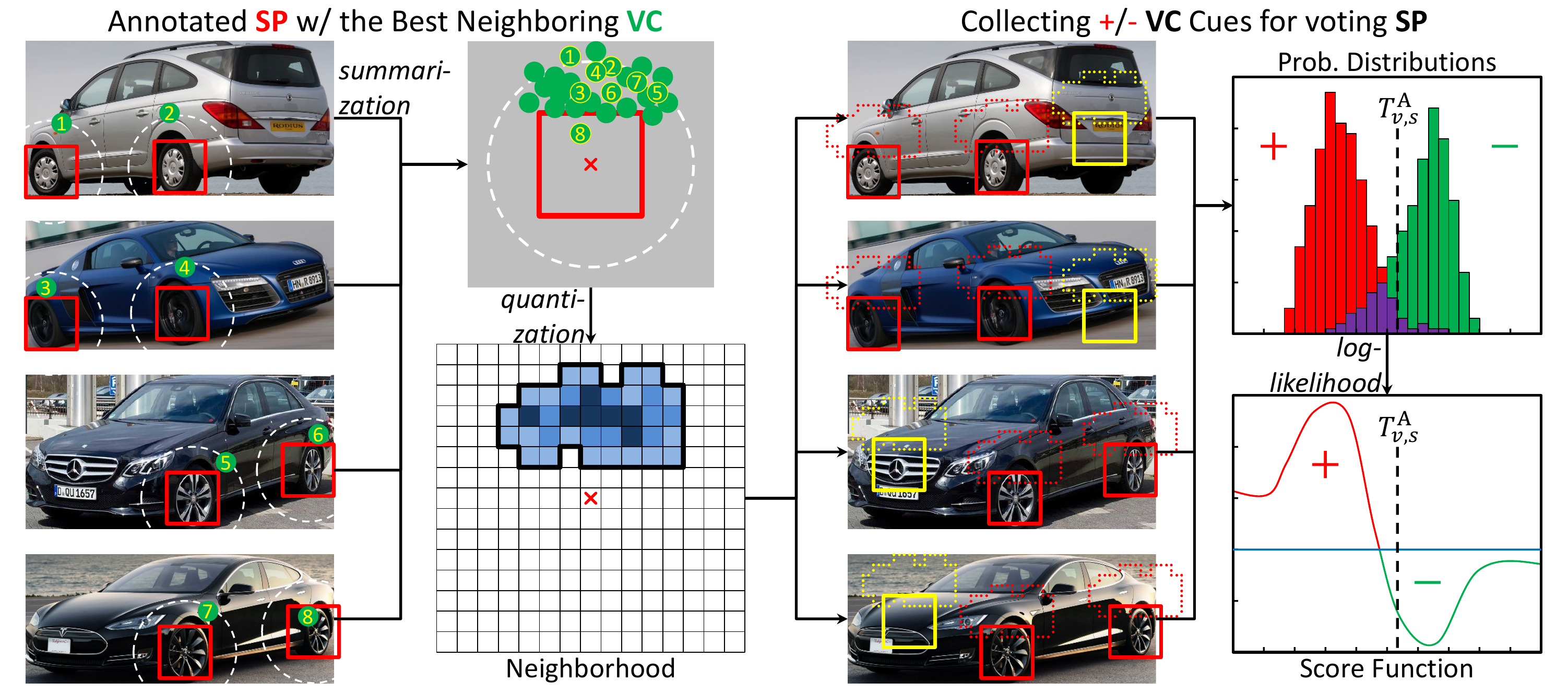}
\caption{Illustration of the training phase of one $\left(\mathrm{VC}_v,\mathrm{SP}_s\right)$ pair. Only one negative point is shown in each image (marked in a yellow frame in the third column), though the negative set is several times larger than the positive set, see \protect\cite{wangXZZXY2017} for details.}
\label{fig:voting_illustration}
\end{figure}

Our compositional model for detecting semantic parts is illustrated in figure~(\ref{fig:voting_illustration}). We develop it in the following stages. Firstly, we study the spatial relations between the visual concepts and the semantic parts in greater detail. In the previous section, we only use spatial relations (e.g., is the visual concept activated within a window centered on the semantic part, where the window size and shape was independent of the visual concept or the semantic part). Secondly, we compute the probability distributions of the visual concept responses depending on whether the semantic part is present or nor. This enables us to use the log-likelihood ratio of these two distributions as evidence for the presence of the semantic part. Thirdly, we describe how the evidence from the visual concepts can be combined to vote for a semantic part, taking into account the spatial positions of the parts and allowing visual concepts to switch off their evidence if it falls below threshold (to deal with occlusion).

\subsubsection{Modeling the Spatial Relationship between Visual Concepts and Semantic Parts}
\label{Algorithm:Training:SpatialRelationship}

We now examine the relationship between visual concepts and semantic parts by studying the  spatial relationship of each $\left(\mathrm{VC}_v,\mathrm{SP}_s\right)$ pair in more detail. Previously we rewarded a visual concept if it responded in
a neighborhood $\mathcal{N}$ of a semantic part, but this neighborhood was fixed and did not depend on the visual concept or the semantic part. But our observations showed that if a visual concept $\mathrm{VC}_v$ provides good evidence  to locate a semantic part $\mathrm{SP}_s$, then the $\mathrm{SP}_s$ may only appear at a restricted set of  positions relative to $\mathrm{VC}_v$. For example, if $\mathrm{VC}_v$ represents the upper part of a {\em wheel}, we expect the semantic part ({\em wheel}) to appear slightly below the position of $\mathrm{VC}_v$.

Motivated by this, for each $\left(\mathrm{VC}_v,\mathrm{SP}_s\right)$ pair we learn a neighborhood function $\mathcal{N}_{v,s}$ indicating the most likely spatial relationship between $\mathrm{VC}_v$ and $\mathrm{SP}_s$. This is illustrated in figure~(\ref{fig:voting_illustration}). To learn the neighborhood we measured the spatial statistics for each $\left(\mathrm{VC}_v,\mathrm{SP}_s\right)$ pair. For each annotated ground-truth position $q \in \mathcal{T}^+_s$ we computed the position $p^\star$ in its $\mathcal{L}_4$ neighborhood which best activates $\mathrm{VC}_v$, {\em i.e.} ${p^\star}={{\arg\min_{p\in\mathcal{N}\!\left(q\right)}}\left\|\mathbf{f}\!\left(\mathbf{I}_p\right)-\mathbf{f}_v\right\|}$, yielding a spatial offset $\Delta p = p ^\star - q$. We normalize these spatial offsets to yield a spatial frequency map ${\mathrm{Fr}_{v,s}\!\left(\Delta p\right)}\in{\left[0,1\right]}$. The neighborhood $\mathcal{N}_{v,s}$ is obtained by thresholding the spatial frequency map. For more details see \cite{wangXZZXY2017}.

\subsubsection{Probability Distributions, Supporting Visual Concepts and Log-likelihood Scores}
\label{Algorithm:Training:Evidences}

We use loglikelihood ratio tests to quantify the evidence a visual concept ${\mathrm{VC}_v}\in{\mathcal{V}}$  can give for a semantic part ${\mathrm{SP}_s}\in{\mathcal{S}}$. For each $\mathrm{SP}_s$, we use the positive and negative samples $\mathcal{T}_s^+$ and  $\mathcal{T}_s^-$
to compute the conditional distributions:
\begin{equation}
\label{Eqn:ProbPositive}
{\mathrm{F}_{v,s}^+\!\left(r\right)}=
{\displaystyle{\frac{\mathrm{d}}{\mathrm{d}r}}\mathrm{Pr}\!\left[\displaystyle{{\min_{p\in\mathcal{N}_{v,s}\!\left(q\right)}}}
    \left\|\mathbf{f}\!\left(\mathbf{I}_p\right)-\mathbf{f}_v\right\|\leqslant r\mid q\in\mathcal{T}_s^+\right]},
\end{equation}
\begin{equation}
\label{Eqn:ProbNegative}
{\mathrm{F}_{v,s}^-\!\left(r\right)}=
{\displaystyle{\frac{\mathrm{d}}{\mathrm{d}r}}\mathrm{Pr}\!\left[\displaystyle{{\min_{p\in\mathcal{N}_{v,s}\!\left(q\right)}}}
    \left\|\mathbf{f}\!\left(\mathbf{I}_p\right)-\mathbf{f}_v\right\|\leqslant r\mid q\in\mathcal{T}_s^-\right]}.
\end{equation}
We call $\mathrm{F}_{v,s}^+\!\left(r\right)$ the {\em target} distribution, which specifies the probably activity pattern for
$\mathrm{VC}_v$ if there is a semantic part $\mathrm{SP}_s$ nearby. If  $\left(\mathrm{VC}_v,\mathrm{SP}_s\right)$ is a good pair,
then the probability $\mathrm{F}_{v,s}^+\!\left(r\right)$ will be peaked close to ${r}={0}$,
({\em i.e.}, there will be some feature vectors within $\mathcal{N}_{v,s}\!\left(q\right)$ that cause $\mathrm{VC}_v$ to activate).
The second distribution, $\mathrm{F}_{v,s}^-\!\left(r\right)$ is the {\em reference} distribution
which specifies the response of the feature vector if the semantic part is not present.

The evidence is provided by the log-likelihood ratio~\cite{Amit_2002_2D}:
\begin{equation}
\label{Eqn:ScoreFunction}
\Lambda_{v,s}\!\left(r\right)=
    {\log\frac{\mathrm{F}_{v,s}^+\!\left(r\right)+\varepsilon}{\mathrm{F}_{v,s}^-\!\left(r\right)+\varepsilon}},
\end{equation}
\noindent where $\varepsilon$ is a small constant chosen to prevent numerical instability.

In practice, we use a simpler method to first decide which visual concepts are best suited to detect each semantic part. We prefer $(\mathrm{VC}_v, \mathrm{SP}_s)$ pairs for which ${\mathrm{F}_{v,s}^+\!\left(r\right)}$ is peaked at small $r$. This corresponds to visual concepts which are fairly good detectors for the semantic part and, in particular, those which have high recall (few false negatives). This can be found from our studies of the visual concepts described in section~(\ref{sec:visualconcepts}). For more details see \cite{wangXZZXY2017}. For each semantic part we select $K$ visual concepts to vote for it. We report results for $K=45$, but good results can be found with fewer. For $K=20$ the performance drops by only $3 \%$, and for $K=10$ it drops by $7.8 \%$.


\subsection{Combining the Evidence by Voting}
\label{Algorithm:Testing}

The voting process starts by extracting CNN features on the {\em pool-4} layer, i.e.  $\{\mathbf{f}_p\})$ for $p \in \mathcal{L}_4$.
We compute the log-likelihood evidence $\Lambda _{v,s}$ for each $\left(\mathrm{VC}_v,\mathrm{SP}_s\right)$ at each position $p$. We threshold this evidence and only keep those positions which are unlikely to be false negatives. Then the vote that a visual concepts gives for a semantic part is obtained by a weighted sum of the loglikelihood ratio (the evidence) and the spatial frequency.

The final score which is added to the position $p+\Delta p$ is computed as:
\begin{equation}
\label{Eqn:Voting}
\mathrm{Vote}_{v,s}\!\left(p\right)=
    {\left(1-\beta\right)\Lambda_{v,s}\!\left(\mathbf{I}_p\right)+\beta\log\frac{\mathrm{Fr}\!\left(\Delta p\right)}{U}}.
\end{equation}
 The first term ensures that there is high evidence of $\mathrm{VC}_v$ firing,
and the second term acts as the spatial penalty ensuring that this $\mathrm{VC}_v$ fires in the right relative position
Here we set ${\beta}={0.7}$, and define ${\log\frac{\mathrm{Fr}\!\left(\Delta p\right)}{U}}={-\infty}$
when ${\frac{\mathrm{Fr}\!\left(\Delta p\right)}{U}}={0}$, and $U$ is a constant.

In order to allow the voting to be robust to occlusion we need this ability to automatically ``switch off" some votes if they are inconsistent with other votes. Suppose a semantic part is partially occluded. Some of its visual concepts may still be visible (i.e. unoccluded) and so they will vote correctly for the semantic part. But some of the visual concepts will be responding to the occluders and so their votes will be contaminated. So we switch off votes which fall below a threshold (chosen to be zero) so that failure of a visual concept to give evidence for a semantic part does not prevent the semantic part from being detected -- i.e. absence of evidence is not evidence of absence. Hence a visual concept is allowed to support the detection of a semantic part, but not allowed to inhibit it.

The final score for detecting a semantic part $\mathrm{SP}_s$ is the sum of the thresholded votes of its visual concepts:
\begin{equation}
\label{Eqn:FinalScore}
{\mathrm{Score}_s\!\left(p\right)}=
    {{\sum_{\mathrm{VC}_v\in\mathcal{V}_s}}\max\left\{0,\mathrm{Vote}_{v,s}\!\left(p\right)\right\}}.
\end{equation}

In practical situations, we will not know the size of the semantic parts in the image and will have to search over different scales. For details of this see \cite{wangXZZXY2017}. Below we report results if the scale is known or unknown.

\renewcommand{\colwidth}{.5cm}
\begin{table*}
\centering
\tabcolsep=0.06cm
\begin{tabular}
{|c|c|c|c|c||c|c|c|}
\hline
\multirow{2}{*}{} & \multicolumn{4}{c||}{Unknown Scale}
                  & \multicolumn{3}{c|}{Known Scale}                              \\
\cline{2-8}
{Object}          & \multicolumn{1}{c|}{\bf S-VC$_{M}$}
                  & \multicolumn{1}{c|}{\bf SVM-VC}
                  & \multicolumn{1}{c|}{\bf FR}         & \multicolumn{1}{c||}{\bf VT}
                  & \multicolumn{1}{c|}{\bf S-VC$_{M}$}
                  & \multicolumn{1}{c|}{\bf SVM-VC}        & \multicolumn{1}{c|}{\bf VT}         \\
\hline\hline
{\em airplane}    & $10.1 $          & $18.2 $          & $\mathbf{45.3 }$ & $30.6 $
                  & $18.5 $          & $25.9 $          & $\mathbf{41.1 }$                    \\
\hline
{\em bicycle}     & $48.0 $          & $58.1 $          & $75.9 $          & $\mathbf{77.8 }$
                  & $61.8 $          & $73.8 $          & $\mathbf{81.6 }$                    \\
\hline
{\em bus}         & $ 6.8 $          & $26.0 $          & $\mathbf{58.9} $  & $58.1 $
                  & $27.8 $          & $39.6 $          & $\mathbf{60.3 }$                    \\
\hline
{\em car}         & $18.4 $          & $27.4 $          & $\mathbf{66.4 }$ & $63.4 $
                  & $28.1 $          & $37.9 $          & $\mathbf{65.8 }$                    \\
\hline
{\em motorbike}   & $10.0 $          & $18.6 $          & $45.6 $          & $\mathbf{53.4 }$
                  & $34.0 $          & $43.8 $          & $\mathbf{58.7 }$                    \\
\hline
{\em train}       & $ 1.7 $          & $ 7.2 $          & $\mathbf{40.7 }$ & $35.5 $
                  & $13.6 $          & $21.1 $          & $\mathbf{51.4 }$                    \\
\hline
{\bf mean}        & $15.8 $          & $25.9 $          & $\mathbf{55.5} $ & $53.1 $
                  & $30.6 $          & $40.4 $          & $\mathbf{59.8 }$                    \\
\hline
\end{tabular}
\caption{
    Detection accuracy (mean AP, $\%$) and scale prediction loss without occlusion. Performance for the deep network {\bf FR} is not altered by knowing the scale.}
\label{Tab:NoOcclusion}
\end{table*}

\subsection{Experiments for Semantic Part Detection with and without Occlusion}
\label{Experiments}

We evaluate our compositional voting model using the VehicleSemanticParts and the VehicleOcclusion datasets. For these experiments we use a tougher, and more commonly used, evaluation criterion~\cite{Everingham_2010_PASCAL}, where a detected semantic part is a true-positive if itsIntersection-over-Union (IoU) ratio between the groundtruth region (i.e. the $100 \times 100$ box centered on the semantic part) and the predicted region is greater or equal to $0.5$, and duplicate detection is counted as false-positive.

In all experiments our algorithm, denoted as {\bf VT}, is compared to three other approaches. The first is  single visual concept detection {\bf S-VC}$_m$ (described in the previous section). The others two are baselines: (I)
Faster-RCNN, denoted by {\bf FR}, which trains a Faster-RCNN~\cite{Ren_2015_Faster} deep network for each of the six vehicles, where each semantic part of that vehicle is considered to be an ``object category''. (II) the Support Vector Machine, denoted by {\bf SVM-VC}, used in the previous section. Note that the Faster-RCNN is much more complex than the alternatives, since it trains an entire deep network, while the others only use the visual concepts plus limited additional training.


We first assume that the target object is not occluded by any irrelevant objects.
Results of our algorithm and its competitors are summarized in Table~\ref{Tab:NoOcclusion}.
Our voting algorithm achieves comparable detection accuracy to Faster-RCNN, the state-of-the-art object detector, and outperforms it if the scale is known. We observe that the best single visual concept {\bf S-VC}$_M$ performs worse than the support vector machine {\bf SVM-VC} (consistent with our results on keypoints), but our voting method {\bf VT} performs much better than the support vector machine.

\renewcommand{\colwidth}{1.7cm}
\begin{table*}
\centering
\tabcolsep=0.1cm
\begin{tabular}
{|c||c|c||c|c||c|c|}
\hline
\multirow{3}{*}{} & \multicolumn{2}{c||}{$2$ Occluders}
                  & \multicolumn{2}{c||}{$3$ Occluders}
                  & \multicolumn{2}{c| }{$4$ Occluders}           \\

\cline{2-7}{}     & \multicolumn{2}{c||}{${0.2}\leqslant{r}<{0.4}$}
                  & \multicolumn{2}{c||}{${0.4}\leqslant{r}<{0.6}$}
                  & \multicolumn{2}{c|}{${0.6}\leqslant{r}<{0.8}$}     \\

\cline{2-7}
{Object}          & \multicolumn{1}{c|}{\bf FR}         & \multicolumn{1}{c||}{\bf VT}
                  & \multicolumn{1}{c|}{\bf FR}         & \multicolumn{1}{c||}{\bf VT}
                  & \multicolumn{1}{c|}{\bf FR}         & \multicolumn{1}{c|}{\bf VT}         \\
\hline\hline
{\em airplane}    & $\mathbf{26.3 }$ & $23.2 $
                  & $\mathbf{20.2 }$ & $19.3 $
                  & $\mathbf{15.2 }$ & $15.1 $          \\
\hline
{\em bicycle}     & $63.8 $          & $\mathbf{71.7 }$
                  & $53.8 $          & $\mathbf{66.3 }$
                  & $37.4 $          & $\mathbf{54.3 }$ \\
\hline
{\em bus}         & $\mathbf{36.0 }$ & $31.3 $
                  & $\mathbf{27.5 }$ & $19.3 $
                  & $\mathbf{18.2 }$ & $ 9.5 $          \\
\hline
{\em car}         & $32.9 $          & $\mathbf{35.9 }$
                  & $19.2 $          & $\mathbf{23.6 }$
                  & $11.9 $          & $\mathbf{13.8 }$ \\
\hline
{\em motorbike}   & $33.1 $          & $\mathbf{44.1 }$
                  & $26.5 $          & $\mathbf{34.7 }$
                  & $17.8 $          & $\mathbf{24.1 }$ \\
\hline
{\em train}       & $17.9 $          & $\mathbf{21.7 }$
                  & $\mathbf{10.0 }$ & $ 8.4 $
                  & $\mathbf{ 7.7 }$ & $ 3.7 $          \\
\hline
{\bf mean}        & $35.0 $          & $\mathbf{38.0 }$
                  & $26.2 $          & $\mathbf{28.6 }$
                  & $18.0 $          & $\mathbf{20.1 }$ \\
\hline
\end{tabular}
\caption{
    Detection accuracy (mean AP, $\%$) when the object is partially occluded and the scale is unknown. Three levels of occlusion are considered.}
\label{Tab:Occlusion}
\end{table*}


\begin{figure}[h]
  \centering
\includegraphics[width=0.9\linewidth]{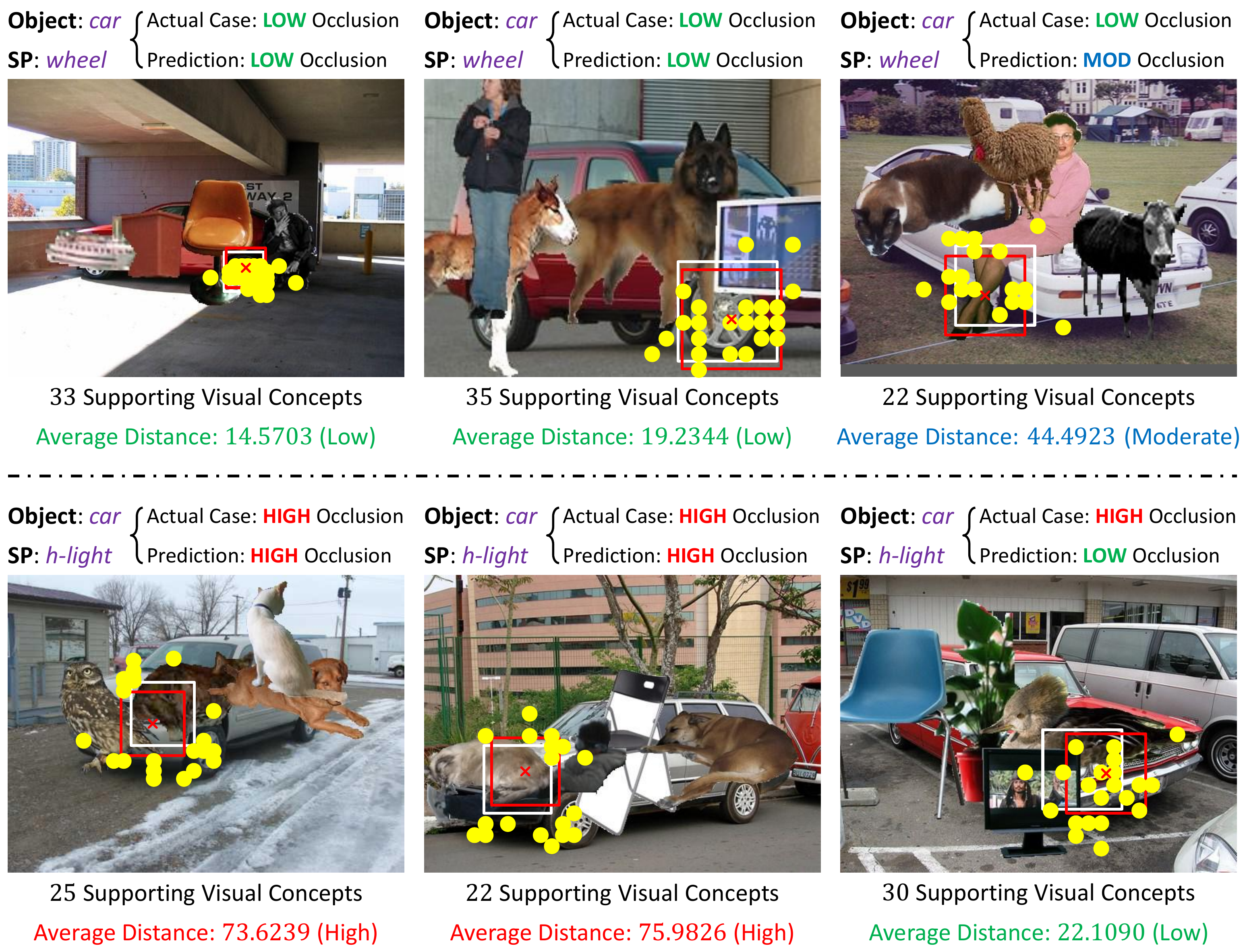}
\caption{This figure shows detection from context: when there are occlusions, the cuesmay come from context which is far away from the part}
\label{fig:detection_context}
\end{figure}

Next, we investigate the case that the target object is partially occluded using the VehicleOcclusion dataset.
The results are shown in Table~\ref{Tab:Occlusion}. Observe that accuracy drops significantly as the amount of occlusion increases, but our compositional voting method {\bf VT} outperforms the deep network {\bf FR} for all levels of occlusion.

Finally we illustrate, in figure~(\ref{fig:detection_context}), that our voting method is able to detect semantic parts using context, even when the semantic part itself is occluded. This is a benefit of our algorithm's ability to use evidence flexibly and adaptively.

\section{Discussion: Visual Concepts and Sparse Encoding \label{sec:discussion}}

Our compositional voting model shows that we can use visual concepts to develop a pattern theoretic models which can outperform deep networks when tested on occluded images. But this is only the starting point. We now briefly describes ongoing work where visual concepts can be used to provide a sparse encoding of objects. 

\begin{figure*}[h]
\centering
 \includegraphics[width=0.3\linewidth]{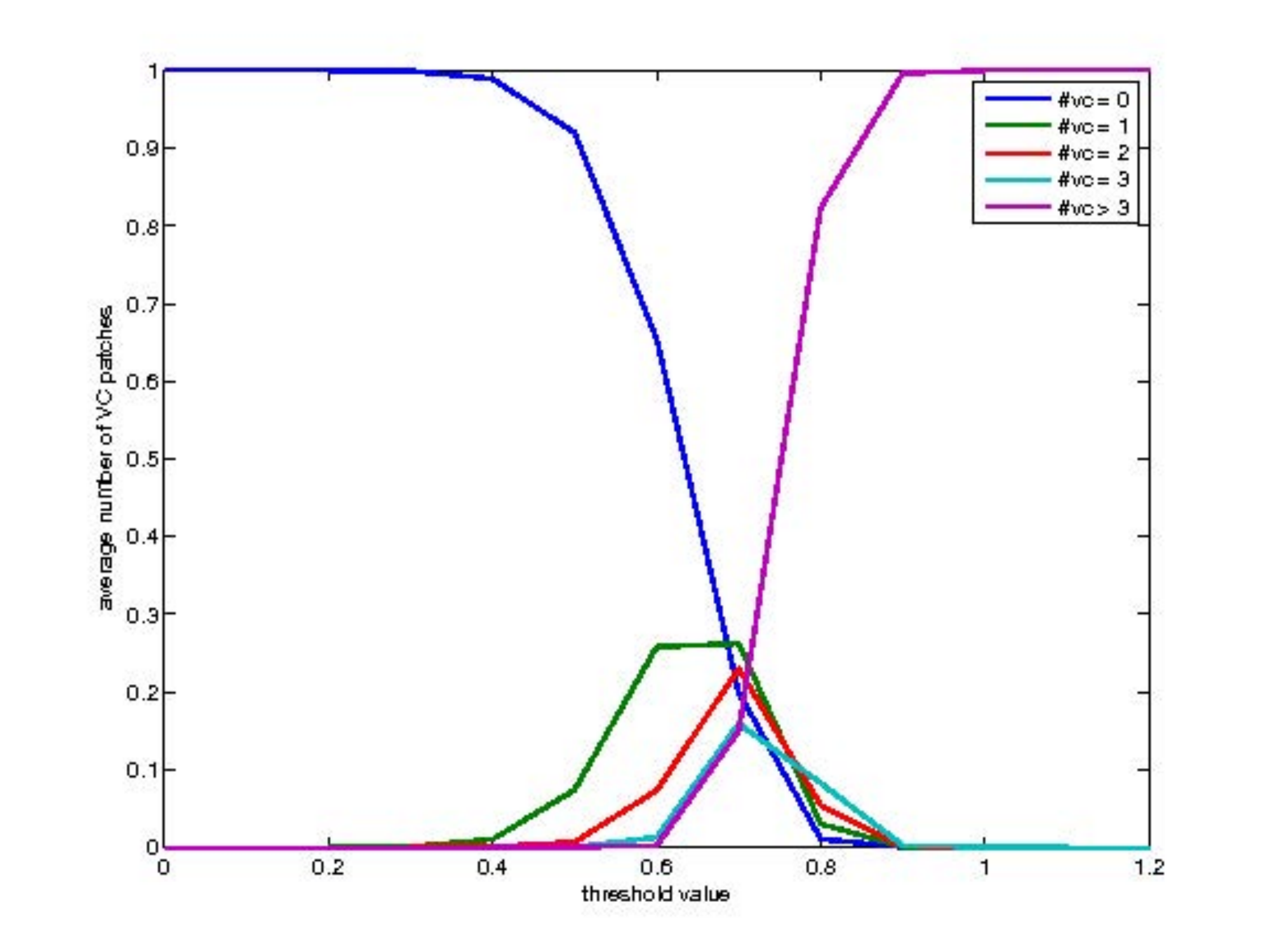}
\includegraphics[width=0.3\linewidth]{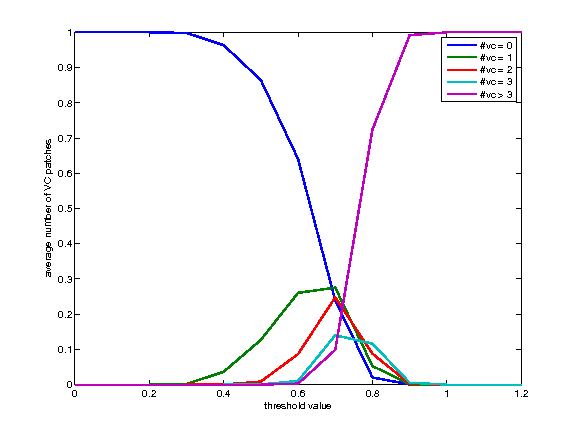} \includegraphics[width=0.3\linewidth]{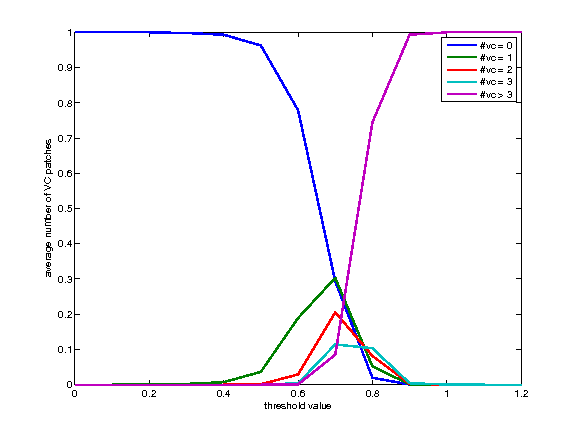}
\caption{This figure shows there is a critical threshold $T_m$, at roughly 0.7, where most pixels are encoded (i.e. activate) a small number of visual concepts. Below this threshold most pixels have no visual concepts activated. By contrast, above the threshold three of more visual concepts are activated. These results are shown for level four features on the cars (left panel), bikes (center panel) and trains (right panel) in VehicleSemanticPart, and we obtain almost identical plots for the other objects in VehicleSemanticPart.}
\label{fig:magic_threshold}
\end{figure*}

The basic idea is to encode an object in terms of those visual concepts which have significant response at each pixel. This gives a much more efficient representation than using the high-dimensional feature vectors  (typically 256 or 512). To obtain this encoding we threshold the distance between the feature vectors (at each pixel) and the centers of the visual concepts. As shown in figure~(\ref{fig:magic_threshold}) we found that a threshold of $T_m=0.7$ ensured that most pixels of the objects were encoded on average by a single visual concept.

 This means that we can approximately encode a feature vector $\mathbf{f}$ by specifying the visual concept that is closest to it, i.e. by ${\hat v} = \arg \min _{v} |\mathbf{f}- \mathbf{f}_v|$ -- provided this distance is less than a threshold $T_m$. More precisely, we can encode a pixel $p$ with feature vectors $\mathbf{f}_p$ by a binary feature vector $\mathbf{b}_p= (b^p_1,..., b^p_v, ...)$ where $v$ indexes the visual concepts and $b^p_v =1$ if $|\mathbf{f}_p - \mathbf{f}_v| < T_m$ and $b^p_v=0$ otherwise. In our neural network implementation, the $b^p_v$ correspond to the activity of the vc-neurons (if soft-max is replaced by hard-max).

There are several advantages to being able to represent spatial patterns of activity in terms of sparse encoding by visual concepts. Instead of representing patterns as {\it signals} $\mathbf{f}$ in the high-dimensional feature space we can give them a more concise {\it symbolic} encoding in terms of visual concepts, or by thresholding the activity of the vc-neurons. More practically, this leads to a natural similarity measure between patterns, so that two patterns with similar encodings are treated as being the same underlying pattern. 

This encoding perspective follows the spirit of Barlow's sparse coding ideas 
\cite{barlow1972single} and is also helpful for developing pattern theory models, which are pursuing in current research. it makes it more practical to learning generative models of the spatial structure of image patterns of an object or a semantic part $\mathrm{S}_s$. This is because generating the binary vectors $\{\mathbf{b}\}$ is much easier than generating high-dimensional feature vectors $\{\mathbf{f}\}$. It can be down, for example, by a simply factorized model $P(\{\mathbf{b}_p\}|\mathrm{SP}_s) = \prod _p P(\mathbf{f}_p|\mathrm{SP}_s)$, where $P(\mathbf{f}_p|\mathrm{SP}_s)= \prod _v P(b_{v,p}|\mathrm{SP}_s)$, and $b_{v,p}=1$ if the $v^{th}$ vc-neuron (or visual concept) is above threshold at position $p$. It also yields simple distance measures between different image patches which enables one-shot/few-shot learning, and also to cluster image patches into different classes, such as into different objects and different viewpoints. 

\section{Conclusion}

This paper is a first attempt to develop novel visual architectures, inspired by pattern theory, which are more flexible and adaptive than deep networks but which build on their successes. We proceeded by studying the activation patterns of deep networks and extracting visual concepts, which both throws light on the internal representations of deep networks but which can also be used as building blocks for compositional models. We also argued that the complexity of vision, particularly due to occlusions, meant that vision algorithms should be tested on more challenging datasets, such as those where occlusion is randomly added, in order to ensure true generalization. Our studies showed that compositional models outperformed deep networks when both were trained on unoccluded datasets  but tested when occlusion was present. 

We conclude that converting deep networks into compositional models, with more explicit representations of parts, will lead to more flexible and adaptive visual architectures than can overcome the limitations of current deep networks and start approaching the abilities of the human visual system. We hope that this research program will help flesh out the details of David's pattern theory \cite{Mumford2010pattern} in order to both improve its practical performance but also to relate it more closely to biological visual systems \cite{mumford1992computational}.

\section*{Acknowledgements}
We gratefully acknowledge support from the National Science Foundations
with NSF STC award CCF-1231216 (Center for Brains, Minds, and Machines)
and NSF Expedition in Computing award  CCF-1317376. We would also like
to acknowledge support from the Office of Naval Research (ONR) with N00014-15-1-2356.


\bibliographystyle{named}
\bibliography{egbib}

\end{document}